\documentclass[10pt,twocolumn,letterpaper]{article}

\usepackage{cvpr}
\usepackage{times}
\usepackage{epsfig}
\usepackage{graphicx}
\usepackage{amsmath}
\usepackage{amssymb}
\usepackage{subfigure}

\graphicspath{{./graphics/}}
\usepackage{tabularx}
\usepackage{booktabs}
\usepackage{caption}
\usepackage{tikz}
\usepackage{pgfplots}
\usepackage{pgfplotstable}
\usepackage{standalone}
\usepackage{bm}
\usepackage[normalem]{ulem}
\newlength{\itemwidth}

\newcolumntype{Y}{>{\centering\arraybackslash}X}

\usetikzlibrary{calc}
\usetikzlibrary{tikzmark}

\pgfplotsset{compat=newest}
\definecolor{999933}{RGB}{153, 153, 51}
\definecolor{BEBADA}{RGB}{190,186,218}
\definecolor{FB8072}{RGB}{251,128,114}
\definecolor{80B1D3}{RGB}{128,177,211}
\definecolor{FDB463}{RGB}{253,180,98}
\definecolor{8DD3C7}{RGB}{141,211,199}
\pgfplotscreateplotcyclelist{custompalette}{
    {999933!50!black,fill=999933},
    {BEBADA!50!black,fill=BEBADA},
    {FB8072!50!black,fill=FB8072},
    {80B1D3!50!black,fill=80B1D3},
    {FDB463!50!black,fill=FDB463},
    {8DD3C7!50!black,fill=8DD3C7}
}

\usepackage[pagebackref=true,breaklinks=true,letterpaper=true,colorlinks,bookmarks=false]{hyperref}

\cvprfinalcopy

\begin{document}

\title{Context-aware Synthesis for Video Frame Interpolation}

\author{Simon Niklaus\\
Portland State University\\
{\tt\small sniklaus@pdx.edu}
\and
Feng Liu\\
Portland State University\\
{\tt\small fliu@cs.pdx.edu}
}

\maketitle

\begin{abstract}

    Video frame interpolation algorithms typically estimate optical flow or its variations and then use it to guide the synthesis of an intermediate frame between two consecutive original frames. To handle challenges like occlusion, bidirectional flow between the two input frames is often estimated and used to warp and blend the input frames. However, how to effectively blend the two warped frames still remains a challenging problem. This paper presents a context-aware synthesis approach that warps not only the input frames but also their pixel-wise contextual information and uses them to interpolate a high-quality intermediate frame. Specifically, we first use a pre-trained neural network to extract per-pixel contextual information for input frames. We then employ a state-of-the-art optical flow algorithm to estimate bidirectional flow between them and pre-warp both input frames and their context maps. Finally, unlike common approaches that blend the pre-warped frames, our method feeds them and their context maps to a video frame synthesis neural network to produce the interpolated frame in a context-aware fashion. Our neural network is fully convolutional and is trained end to end. Our experiments show that our method can handle challenging scenarios such as occlusion and large motion and outperforms representative state-of-the-art approaches.

\end{abstract}

\vspace{-0.2in}
\section{Introduction}
\label{sec:intro}
\begin{figure}\centering
    \setlength{\tabcolsep}{0.0cm}
    \setlength{\itemwidth}{4.1cm}

    \begin{tabularx}{\textwidth}{c @{\hspace{0.1cm}} c @{\hspace{0.05cm}} c}
            \includegraphics[width=\itemwidth]{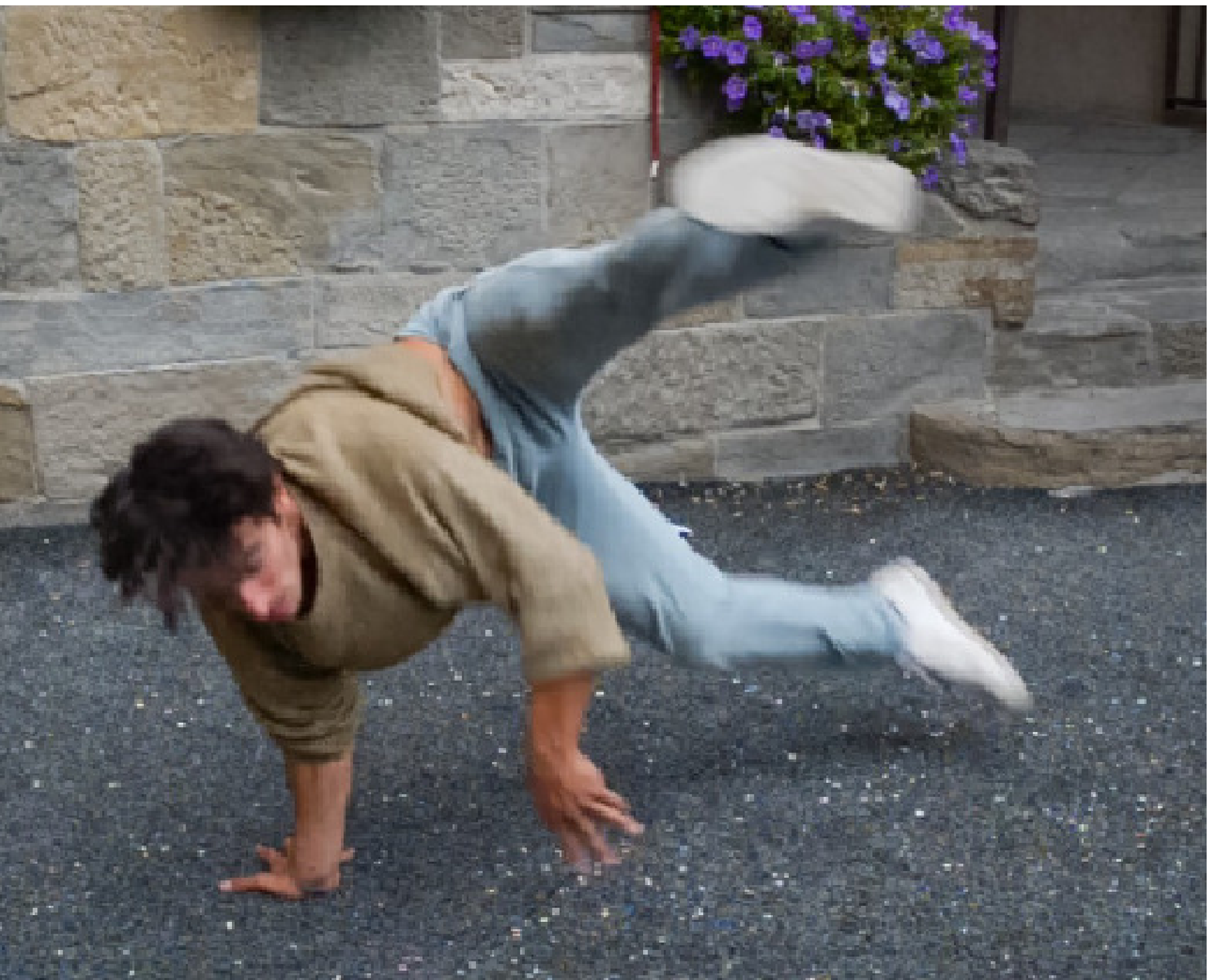}
        &
            \includegraphics[width=\itemwidth]{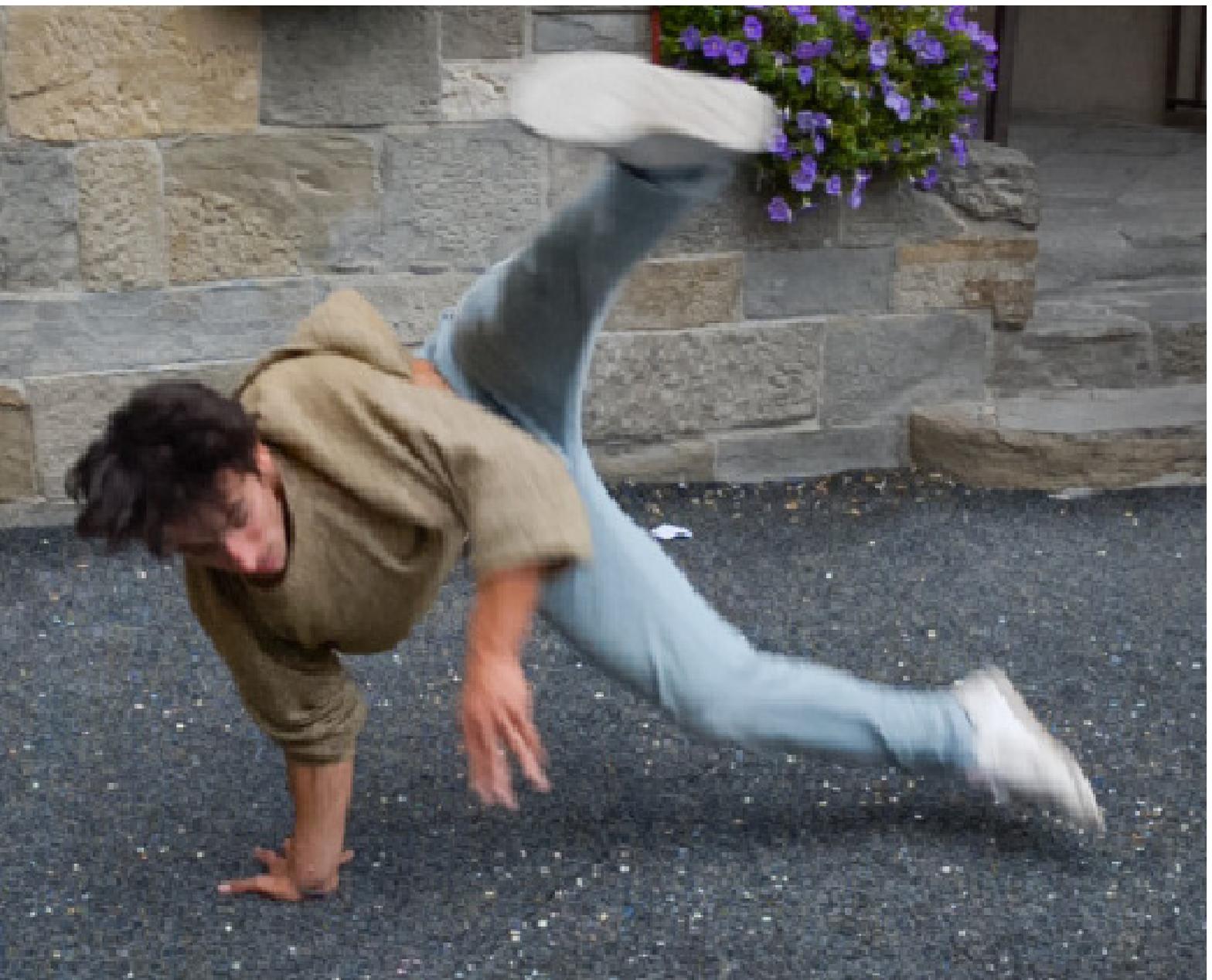}
        \vspace{-0.1cm} \\
            \footnotesize First frame, $I_1$
        &
            \footnotesize Second frame, $I_2$
        \vspace{0.1cm} \\
            \includegraphics[width=\itemwidth]{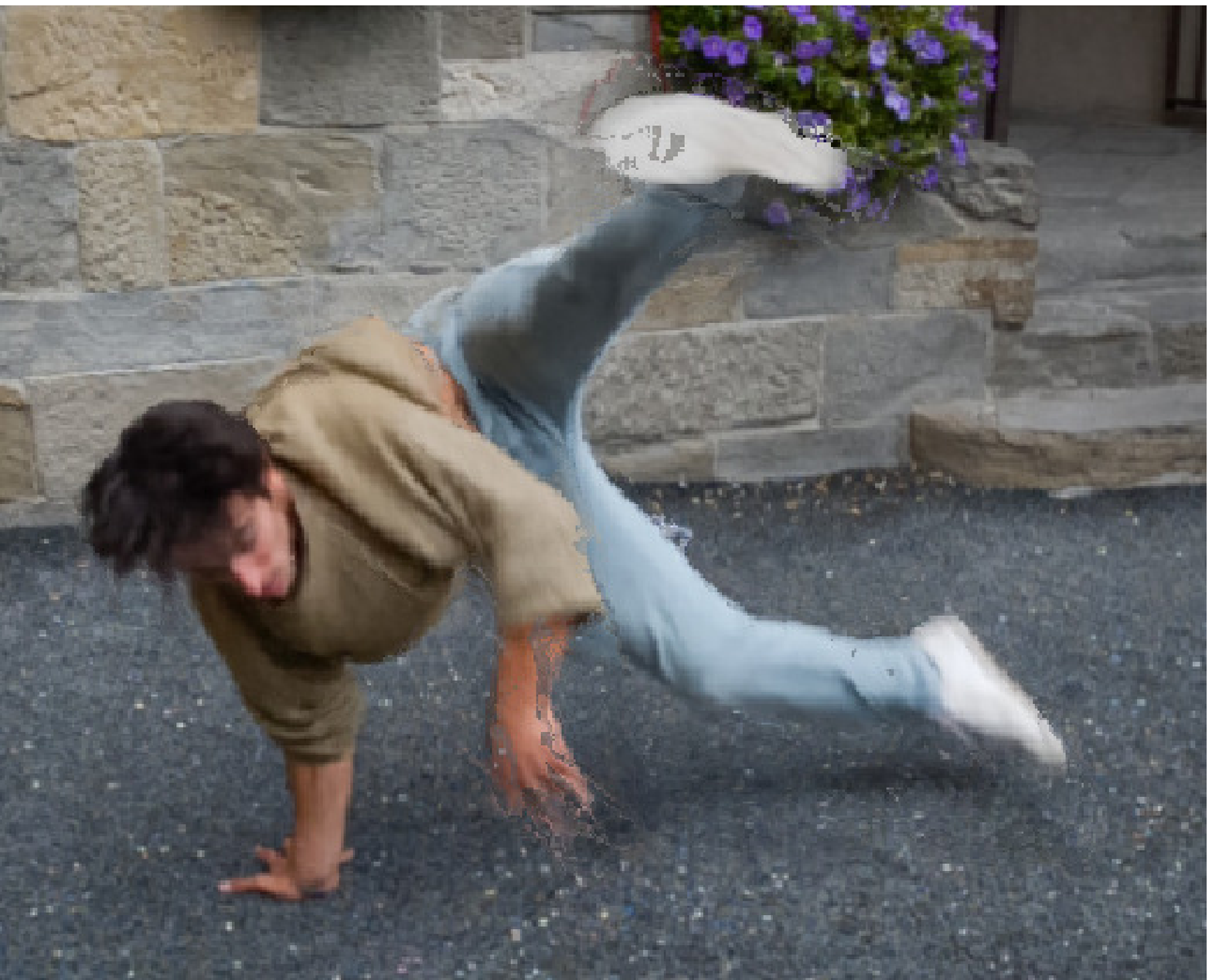}
        &
            \includegraphics[width=\itemwidth]{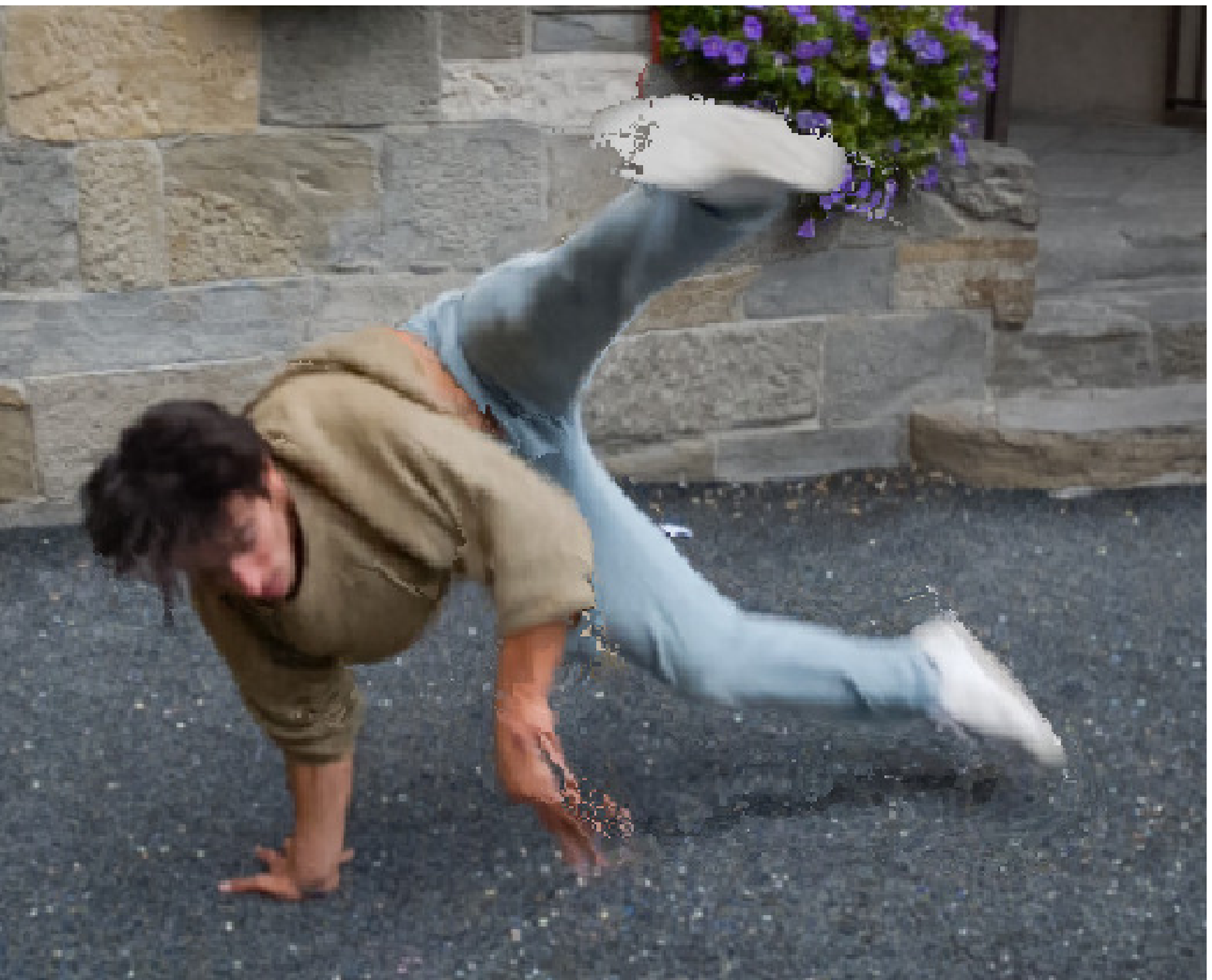}
        \vspace{-0.1cm} \\
            \footnotesize Bidirectional flow-guided blending
        &
            \footnotesize MDP-Flow2~\cite{Xu_PAMI_2012}
        \vspace{0.1cm} \\
            \includegraphics[width=\itemwidth]{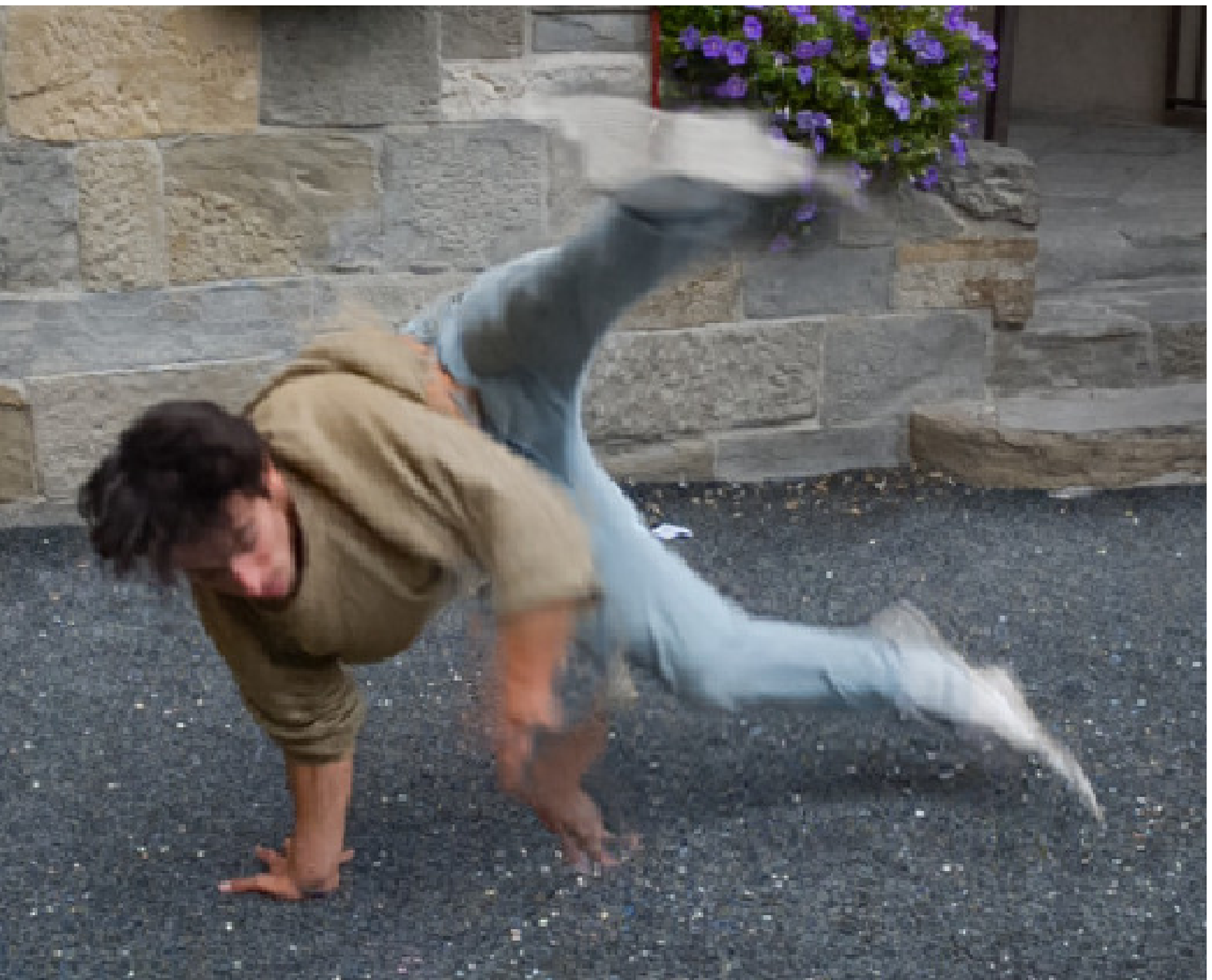}
        &
            \includegraphics[width=\itemwidth]{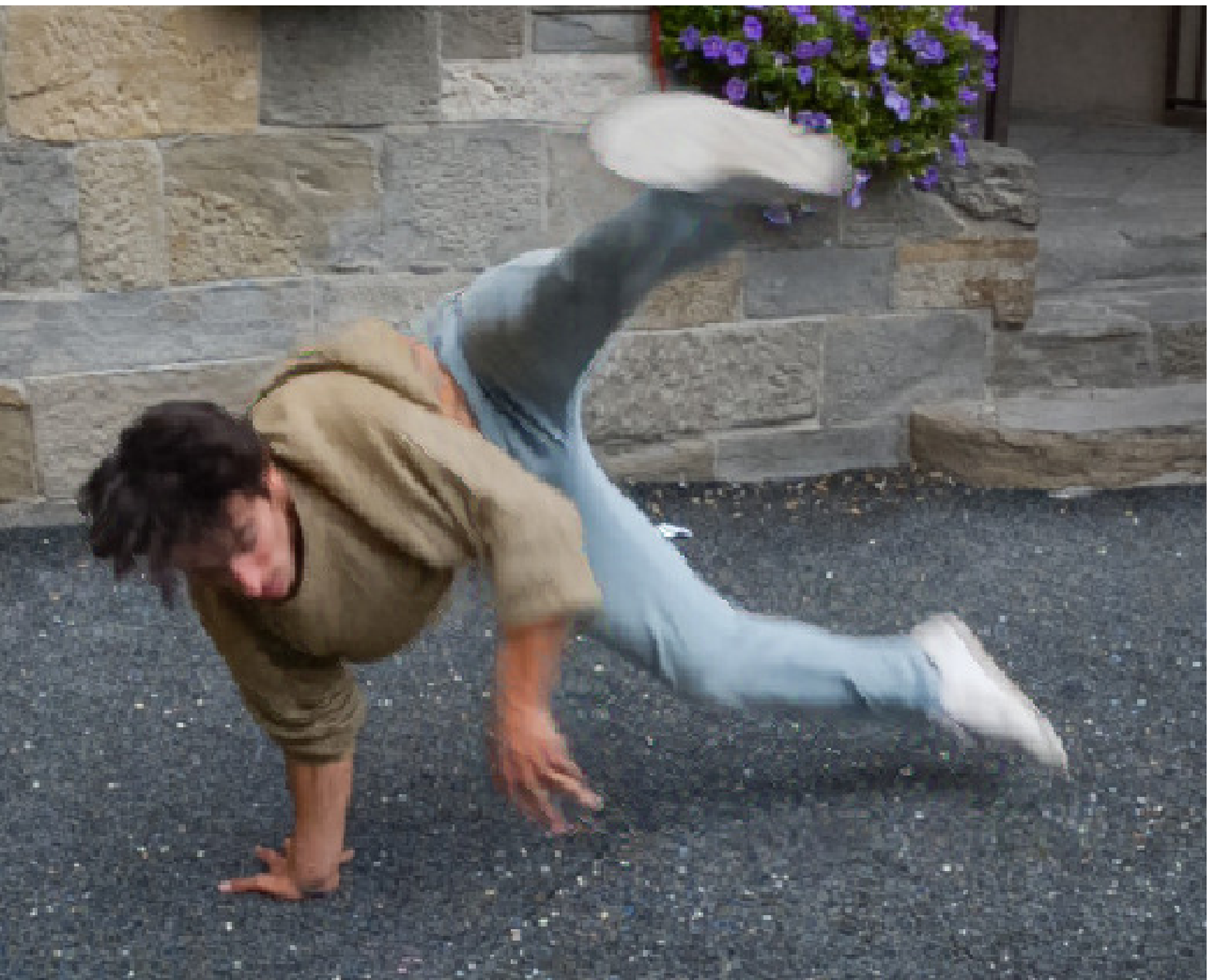}
        \vspace{-0.1cm} \\
            \footnotesize SepConv - $\mathcal{L}_F$~\cite{Niklaus_ICCV_2017}
        &
            \footnotesize Ours - $\mathcal{L}_F$
        \\
    \end{tabularx}\vspace{-0.2cm}
    \caption{A challenging example. Compared to existing frame interpolation methods, our approach can better handle large motion, significant occlusion, and motion blur.}\vspace{-0.3cm}
    \label{fig:teaser}
\end{figure}

Video frame interpolation is one of the basic video processing techniques. It is used to generate intermediate frames between any two consecutive original frames. Video frame interpolation algorithms typically estimate optical flow or its variations and use them to warp and blend original frames to produce interpolation results~\cite{Baker_OTHER_2011, Liu_ICCV_2017, Niklaus_CVPR_2017}. {\let\thefootnote\relax\footnote{\url{http://graphics.cs.pdx.edu/project/ctxsyn}}}

The quality of frame interpolation results depends heavily on that of optical flow. While these years have observed great progress in research on optical flow, challenges such as occlusion and large motion still remain. As reported in recent work, the optical flow accuracy decreases as the motion increases~\cite{Butler_ECCV_2012, Janai_CVPR_2017}. Therefore, many frame interpolation methods estimate bidirectional optical flow between two input frames and use them to handle inaccuracies of motion estimation and occlusion~\cite{Herbst_OTHER_2009,  Raket_OTHER_2012, Zhou_ECCV_2016}. Some of the recent bidirectional approaches, such as~\cite{Zhou_ECCV_2016}, also estimate weight maps to adaptively blend the optical flow-guided warped frames. However, as shown in Figure~\ref{fig:teaser}, blending the warped frames sometimes is limited in handling occlusion and inaccuracies of optical flow, as it requires accurate pixel-wise correspondence between the warped frames.

This paper presents a context-aware frame synthesis approach to high-quality video frame interpolation. Instead of blending pre-warped input frames, our approach adopts a more flexible synthesis method that can better handle inaccuracies of optical flow and occlusion. Specifically, we develop a frame synthesis deep neural network that directly produces an intermediate frame from the pre-warped frames, without being limited to pixel-wise blending. To further improve the interpolation quality, our method employs a pre-trained image classification neural network~\cite{He_CVPR_2016} to extract per-pixel context information from input frames. These context maps are pre-warped together with the input frames guided by bidirectional optical flow. The frame synthesis neural network takes the pre-warped frames and their context maps as input and is able to handle challenging scenarios such as significant occlusion and motion to produce high-quality interpolation results.

Our experiments show that our method is able to handle difficult frame interpolation cases and produces higher quality results than state-of-art approaches~\cite{Liu_ICCV_2017, Meyer_CVPR_2015, Niklaus_CVPR_2017, Xu_PAMI_2012}. On the Middlebury interpolation benchmark, our method generates the best results among all the published ones~\cite{Baker_OTHER_2011}. We attribute the capability of our method to produce high-quality interpolation results to the following factors. First, our frame synthesis neural network is not limited to pixel-wise blending. It is able to make use of neighboring pixels for frame synthesis, which is important to handle occlusion and errors in the optical flow. Second, the extracted and pre-warped context maps provide information in addition to motion and enable our neural network to perform informative frame synthesis. Finally, we would like to acknowledge that our method adopts PWC-Net~\cite{Sun_CORR_2017}, a state-of-the-art optical flow algorithm to estimate the bidirectional flow, which provides a good initialization. Our method is able to leverage the continuing success of optical flow research to further improve our result in the future.

\section{Related Work}
\label{sec:related}
Video frame interpolation is a classic computer vision problem. While it is a constrained problem of novel view interpolation~\cite{Chen_TOG_1993, Kang_OTHER_2006, Szeliski_BOOK_2010}, a variety of dedicated algorithms have been developed for video frame interpolation, which will be the focus of this section.

Classic video frame interpolation algorithms contain two steps: optical flow estimation and frame interpolation~\cite{Baker_OTHER_2011, Werlberger_OTHER_2011, Yu_OTHER_2013}. The quality of frame interpolation largely depends on that of optical flow, which is one of the basic problems in computer vision and has been attracting a significant amount of research effort. These years, the quality of optical flow is consistently improving~\cite{Baker_OTHER_2011, Butler_ECCV_2012, Janai_CVPR_2017, Menze_CVPR_2015}. Similar to many other computer vision problems, deep learning approaches are used in various high-quality optical flow algorithms~\cite{Dosovitskiy_ICCV_2015, Gadot_CVPR_2016, Sun_CORR_2017, Weinzaepfel_ICCV_2013, Xu_CVPR_2017}. However, optical flow is an inherently difficult problem and challenges still remain in scenarios, such as significant occlusion, large motion, lack of texture, and blur. Optical flow results are often error-prone when facing these difficult cases.

Various approaches have been developed to handle the inaccuracies and missing information from optical flow results. For example, Baker~\etal fill the holes in the optical flow results before using them for interpolation. Another category of approaches estimate bidirectional optical flow between two input frames and use them to improve the accuracy and infer missing motion information due to occlusion~\cite{Herbst_OTHER_2009, Raket_OTHER_2012, Zhou_ECCV_2016}. Some recent methods also estimate per-pixel weight maps to better blend the flow-guided pre-warped frames than using global blending coefficients~\cite{Zhou_ECCV_2016}. These methods have been shown effective; however, their performance is sometimes limited by the subsequent interpolation step that blends the pre-warped frames to produce the final result. This paper builds upon these bidirectional flow approaches and extends them by developing a deep frame synthesis neural network that is not limited to pixel-wise blending and thus is more flexible to tolerate errors in optical flow. Moreover, our method extracts and warps pixel-wise contextual maps together with input frames and feeds them to the deep neural network to enable context-aware frame synthesis.

Several recent methods interpolate video frames without estimating optical flow. Meyer~\etal developed a phase based frame interpolation approach that generates intermediate frames through per-pixel phase modification~\cite{Meyer_CVPR_2015}. Long~\etal directly train a deep convolutional neural network that takes two consecutive original frames as input and outputs an intermediate frame without an intermediate motion estimation step. Their results, however, are sometimes blurry~\cite{Long_ECCV_2016}. Niklaus~\etal merge motion estimation and pixel synthesis into a single step of local convolution. They employ a deep convolutional neural network to estimate a pair of convolution kernels for each output pixel and then use them to convolve with input frames to produce an intermediate frame~\cite{Niklaus_CVPR_2017, Niklaus_ICCV_2017}. While their methods can handle occlusion and reasonable size motion, they still cannot handle large motion. Liu~\etal developed a deep neural network to extract voxel flow that is used to sample the space-time video volume to generate the interpolation result~\cite{Liu_ICCV_2017}. As their method samples the $2^3$ spatial-temporal neighborhood according to the voxel flow, it is still limited in accommodating inaccuracies in motion/voxel flow estimation.

\section{Video Frame Interpolation}
\label{sec:method}
\begin{figure*}\centering
    \includegraphics[]{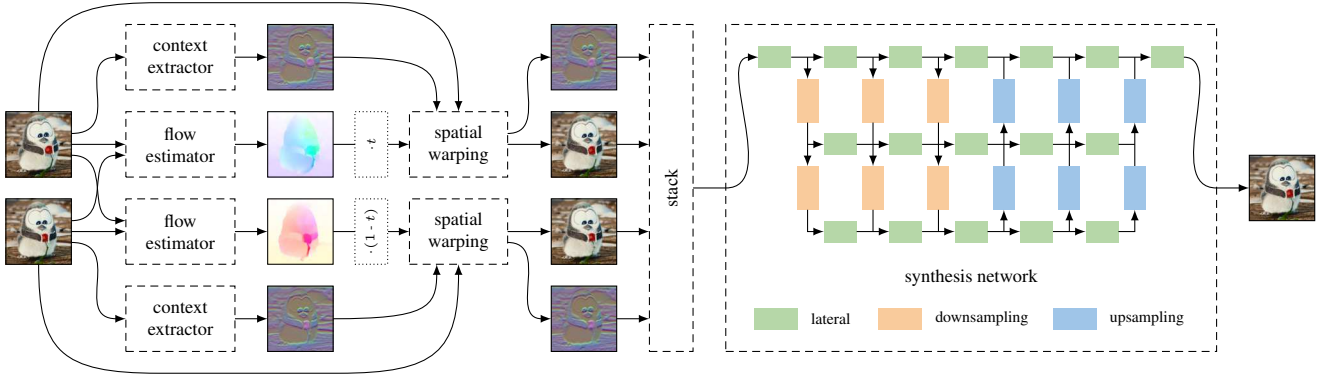}\vspace{-0.15cm}
	\caption{Algorithm overview. Given two consecutive input frames, our method first estimates bidirectional flow between them and extracts per-pixel context maps. Our method then pre-warps the input frames and their corresponding context maps. Finally, the warped frames and context maps are fed into a frame synthesis neural network to generate the interpolation result at a desired temporal position $t \in \left[ 0, 1 \right]$. The synthesis network utilizes a modified GridNet architecture consisting of three rows and six columns. It employs $3 \times 3$ convolutions with per-row channel-sizes of 32, 64, and 96 respectively.}\vspace{-0.3cm}
	\label{fig:architecture}
\end{figure*}

Given two consecutive video frames $I_1$ and $I_2$, our goal is to generate an intermediate frame $\hat{I}_t$ at the temporal location \emph{t} in between the two input frames. Our method works in three stages as illustrated in Figure~\ref{fig:architecture}. We first estimate bidirectional optical flow between $I_1$ and $I_2$ and extract pixel-wise context maps. We then warp the input frames together with their context maps according to the optical flow. We finally feed them into a deep frame synthesis neural network to generate the interpolation result.

We estimate the bidirectional optical flow $F_{1 \rightarrow 2}$ and $F_{2 \rightarrow 1}$ between the two frames using the recent PWC-Net method~\cite{Sun_CORR_2017}. This method utilizes a multi-scale feature pyramid in combination with warping and cost volumes. It performs well in standard benchmarks while at the same time being computationally efficient.

Guided by the bidirectional flow, we pre-warp the input frames. Specifically, we employ forward warping that uses optical flow $F_{1 \rightarrow 2}$ to warp input frame $I_1$ to the target location and obtain a pre-warped frame $\hat{I}_t^1$. During forward warping, we measure the flow quality by checking the brightness constancy and discard contributions from flow vectors that significantly violate this constraint. We warp input frame $I_2$ and generate the pre-warped frame $\hat{I}_t^2$ in the same way. Note that this approach can lead to holes in the warped output, mostly due to occlusion. We chose forward warping to allow the synthesis network to identify occluded regions and to avoid confusing the synthesis network with heuristic measures for filling in missing content.

\begin{figure}\centering
    \includegraphics[]{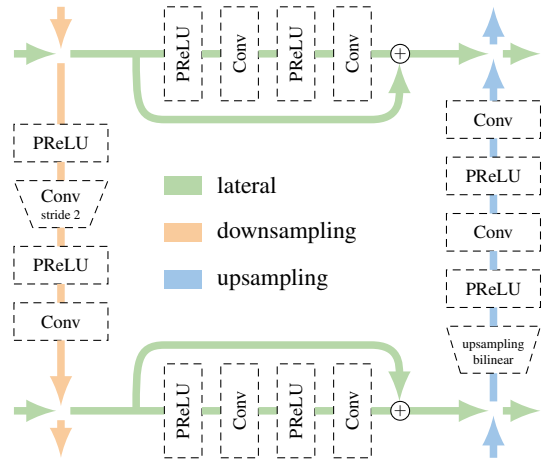}\vspace{-0.1cm}
    \caption{Building block of our frame synthesis neural network in Figure~\ref{fig:architecture}, adapted from the GridNet architecture.}\vspace{-0.4cm}
    \label{fig:blocks}
\end{figure}

\vspace{0.05in}
\subsection{Context-aware Frame Synthesis}\label{sec:network}
 
Given two pre-warped frames $\hat{I}_t^1$ and $\hat{I}_t^2$, existing bidirectional methods combine them through weighted blending to obtain the interpolation result $\hat{I}_t$. This pixel-wise blending approach requires pixel-wise accuracy of optical flow. We develop a more flexible approach by training a synthesis neural network that takes the two pre-warped images as input and directly generates the final output, without resorting to pixel-wise blending. In this way, our method can better tolerate inaccuracies of optical flow estimation.

Generating the final interpolation result solely from the two pre-warped frames has a limitation in that rich contextual information from the original frames is lost. Note that the contextual information for each pixel in pre-warped frames could be compromised due to errors in optical flow. We therefore extract per-pixel context maps from the original input frames and warp them together with the input frames, before feeding them into the synthesis network as shown in Figure~\ref{fig:architecture}. We chose to extract the contextual information by utilizing the response of the \verb|conv1| layer from ResNet-18~\cite{He_CVPR_2016}. Each pixel in the input frame accordingly has a contextual vector that describes its $7 \times 7$ neighborhood. Note that we modify the stride of the \verb|conv1| layer to be $1$ instead of $2$ such that the context map has the same size as its corresponding input frame.

We extend the GridNet architecture to generate the final interpolation result from the two pre-warped frames and their context maps~\cite{Fourure_BMVC_2017}. Instead of having a single sequence of consecutive layers like typical neural networks, a GridNet processes features in a grid of rows and columns as shown in Figure~\ref{fig:architecture}. The layers in each row form a stream in which the feature resolution is kept constant. Each stream processes information at a different scale and the columns connect the streams to exchange information by using downsampling and upsampling layers. This generalizes typical encoder-decoder architectures in which features are processed along a single path~\cite{Long_CVPR_2015, Pohlen_CVPR_17}. In comparison, a GridNet learns how information at different scales should be combined on its own, making it well-suited for pixel-wise problems where global low-resolution information guides local high-resolution predictions. We modified the horizontal and vertical connections of the GridNet architecture as shown in Figure~\ref{fig:blocks}. Specifically, we follow recent findings in image enhancement tasks and do not use Batch Normalization~\cite{Ioffe_ICML_2015, Nah_CVPR_2017, Lim_CVPR_2017}. Furthermore, we incorporate parametric rectified linear units for improved training and use bilinear upsampling to avoid checkerboard artifacts~\cite{He_ICCV_2015, Odena_OTHER_2016}. Note that our configuration of three streams, and thus three scales, leads to a relatively small receptive field of the network~\cite{Luo_NIPS_2016}. However, it fits our problem well since pre-warping already compensates for motion.

\vspace{0.05in}
\noindent\textbf{Loss functions.} We consider various loss functions that measure the difference between the interpolated frame $\hat{I}$ and its ground truth $I_{gt}$. For the color-based loss function, we, in accordance with reports that $\ell_2$ leads to blurry results~\cite{Goroshin_NIPS_2015, Long_ECCV_2016, Mathieu_ICLR_2016, Ranzato_CORR_2014, Srivastava_ICML_2015}, employ a $\ell_1$-based loss function as follows.
\begin{equation}
    \mathcal{L}_1 = \left\| \hat{I} - I_{gt} \right\| _1
\end{equation}
We also consider a feature-based loss that measures perceptual difference~\cite{Chen_ICCV_2017, Dosovitskiy_NIPS_2016, Johnson_ECCV_2016, Ledig_CORR_2016, Sajjadi_CORR_2016, Zhu_ECCV_2016}. Specifically, we follow Niklaus~\etal~\cite{Niklaus_ICCV_2017} and utilize the response of the \verb|relu4_4| layer from VGG-19~\cite{Simonyan_CORR_2014} to extract features $\phi$ and measure their difference as follows.
\begin{equation}
    \mathcal{L}_F = \left\| \phi(\hat{I}) - \phi(I_{gt}) \right\| _2^2
\end{equation}
Another alternative that we adopt measures the difference between Laplacian pyramids~\cite{Bojanowski_CORR_2017}. This multi-scale loss separates local and global features, thus potentially providing a better loss function for synthesis tasks. By denoting the $i$-th layer of a Laplacian pyramid representation of an image $I$ as $L^{i}(I)$, we define the loss as follows.
\begin{equation}
    \mathcal{L}_{\textit{Lap}} = \sum_{i = 1}^{5} \thinspace 2^{i - 1} \left\| L^{i}(\hat{I}) - L^{i}(I_{gt}) \right\| _1
\end{equation}
This loss function takes the difference between two Laplacian pyramid representations with five layers. Additionally, we scale the contributions from the deeper levels in order to partially account for their reduced spatial support. We will discuss how these different loss functions affect the interpolation results in Section~\ref{sec:exp}. Briefly, the feature loss tends to produce visually more pleasing results while the Laplacian and the $\ell_1$-based loss produce better quantitative results.

\subsection{Training}

We train our frame synthesis neural network on examples of size $256 \times 256$ using AdaMax with $\beta_1 = 0.9$, $\beta_2 = 0.999$, a learning rate of $0.001$, and a mini-batch size of $8$ samples~\cite{Kingma_CORR_2014}. We eventually collect $50,000$ training examples and train the neural network on these for $50$ epochs.

\vspace{0.05in}
\noindent\textbf{Training dataset.} We collected training samples from videos by splitting each video into sets of three frames, such that the center frame in each triplet serves as ground truth. We then extracted patches with a size of $300 \times 300$ from these frame triplets, allowing us to only select patches with useful information~\cite{Bansal_CORR_2017}. We prioritized patches that contain sufficiently large motion as well as high-frequency details. To determine the former, we estimated the optical flow between the first and the last patch in each triplet using DIS~\cite{Kroeger_ECCV_216}. We followed the methodology of Niklaus~\etal to select raw videos and extract $50,000$ samples with a resolution of $300 \times 300$ pixels~\cite{Niklaus_ICCV_2017}. Specifically, we obtained high-quality videos from YouTube with the extracted samples having an estimated average optical flow of $4$ pixels. The largest optical flow is is estimated to be $41$ pixels and about $5\%$ of the pixels have an estimated optical flow of at least $21$ pixels.

\vspace{0.05in}
\noindent\textbf{Data augmentation.} While the raw samples in the training dataset have a resolution of $300 \times 300$ pixels, we only use patches with a size of $256 \times 256$ during training. This allows us to augment the training data on the fly by choosing a different randomly cropped patch from a training sample each time it is being used. In order to eliminate potential priors, we randomly flip each patch vertically or horizontally and randomly swap the temporal order.

\subsection{Implementation}

We implemented our approach using PyTorch. In our implementation of PWC-Net~\cite{Sun_CORR_2017}, we realized the necessary cost volume layer using CUDA and utilize the grid sampler from cuDNN in order to perform the involved warping. Wherever available, we utilized optimized cuDNN layers to implement the synthesis network~\cite{Chetlur_CORR_2014}. We implemented the pre-warping algorithm using CUDA and leverage atomic operations to efficiently deal with race conditions. Fully training the synthesis network using a Nvidia Titan X (Pascal) takes about two days. On this graphics card, it takes $0.77$ seconds to interpolate a $1920 \times 1080$ frame and $0.36$ seconds to interpolate a $1280 \times 720$ frame. This includes all steps, the bidirectional optical flow estimation, the context extraction, the pre-warping, and the guided synthesis.

\vspace{0.05in}
\noindent\textbf{Instance normalization.} We have found instance normalization to improve the interpolation quality~\cite{Ulyanov_CORR_2016}. Specifically, we jointly normalize the two input frames and reverse this normalization on the resulting output from the synthesis network. We likewise jointly normalize the extracted context information but do not need to reverse this operation afterwards. There are two arguments that support this step. First, it removes instance-specific contrast information, removing one possible type of variation in the input. Second, it ensures that the color space and the context space have a similar range, which helps to train the synthesis network.

\section{Experiments}
\label{sec:exp}
To evaluate our method, we quantitatively and qualitatively compare it with several baselines as well as representative state-of-the-art video frame interpolation methods. Please also refer to the supplementary video demo to examine the visual quality of our results.

\begin{table}\centering
    \setlength{\tabcolsep}{0.0cm}
    \renewcommand{\arraystretch}{1.1}
    
    \newcommand{\middSec}[1]{\tiny #1}
    \newcommand{\middAll}[1]{\scalebox{0.83}[1.0]{$\uline{ #1 }$}}
    \newcommand{\middALL}[1]{\scalebox{0.83}[1.0]{$\uline{\bm{ #1 }}$}}
    \newcommand{\middOth}[1]{\scalebox{0.83}[1.0]{$ #1 $}}
    \newcommand{\middOTH}[1]{\scalebox{0.83}[1.0]{$\bm{ #1 }$}}
    \scriptsize
    \begin{tabularx}{\columnwidth}{@{\hspace{0.1cm}} X @{\hspace{0.0cm}} c @{\hspace{0.1cm}} c @{\hspace{0.0cm}} c @{\hspace{0.225cm}} r @{\hspace{0.225cm}} c @{\hspace{0.1cm}} c @{\hspace{0.0cm}} l @{\hspace{0.225cm}} c @{\hspace{0.1cm}} c @{\hspace{0.0cm}} l @{\hspace{0.225cm}} c @{\hspace{0.1cm}} c @{\hspace{0.0cm}} l @{\hspace{0.1cm}}}
        \toprule
            & \multicolumn{2}{c}{\sc Average} &&& \multicolumn{2}{c}{Laboratory} && \multicolumn{2}{c}{Synthetic} && \multicolumn{2}{c}{Real World} &
        \\ \cmidrule{2-3} \cmidrule{6-7} \cmidrule{9-10} \cmidrule{12-13}
            & \middSec{PSNR} & \middSec{SSIM} &&& \middSec{PSNR} & \middSec{SSIM} && \middSec{PSNR} & \middSec{SSIM} && \middSec{PSNR} & \middSec{SSIM} &
        \\ \midrule
            Ours - $\mathcal{L}_{\textit{Lap}}$ & \middOTH{36.93} & \middOTH{0.964} &&\multicolumn{1}{|c}{}& \middOTH{40.09} & \middOTH{0.971} && \middOTH{38.89} & \middOTH{0.981} && \middOTH{31.79} & \middOTH{0.939} &
        \\
            Ours - $\mathcal{L}_{1}$ & \middOth{36.71} & \middOth{0.963} &&\multicolumn{1}{|c}{}& \middOth{39.90} & \middOTH{0.971} && \middOth{38.50} & \middOth{0.980} && \middOth{31.74} & \middOTH{0.939} &
        \\
            Ours - $\mathcal{L}_{F}$ & \middOth{35.95} & \middOth{0.959} &&\multicolumn{1}{|c}{}& \middOth{39.21} & \middOth{0.968} && \middOth{37.17} & \middOth{0.975} && \middOth{31.46} & \middOth{0.933} &
        \\
            SepConv - $\mathcal{L}_{1}$ & \middOth{35.73} & \middOth{0.959} &&\multicolumn{1}{|c}{}& \middOth{39.93} & \middOTH{0.971} && \middOth{36.78} & \middOth{0.974} && \middOth{30.47} & \middOth{0.932} &
        \\
            SepConv - $\mathcal{L}_{F}$ & \middOth{35.03} & \middOth{0.954} &&\multicolumn{1}{|c}{}& \middOth{39.49} & \middOth{0.968} && \middOth{35.59} & \middOth{0.967} && \middOth{30.02} & \middOth{0.927} &
        \\
            MDP-Flow2 & \middOth{34.81} & \middOth{0.953} &&\multicolumn{1}{|c}{}& \middOth{38.40} & \middOth{0.966} && \middOth{35.07} & \middOth{0.959} && \middOth{30.97} & \middOth{0.934} &
        \\
            DeepFlow2 & \middOth{34.34} & \middOth{0.951} &&\multicolumn{1}{|c}{}& \middOth{38.33} & \middOth{0.966} && \middOth{34.63} & \middOth{0.956} && \middOth{30.07} & \middOth{0.932} &
        \\
            Meyer~\etal & \middOth{30.89} & \middOth{0.883} &&\multicolumn{1}{|c}{}& \middOth{34.22} & \middOth{0.912} && \middOth{27.88} & \middOth{0.813} && \middOth{30.58} & \middOth{0.925} &
        \\ \bottomrule
    \end{tabularx}\vspace{-0.2cm}
    \caption{Evaluation of the loss functions.}\vspace{-0.2cm}
    \label{tbl:public}
\end{table}

\begin{table}\centering
    \setlength{\tabcolsep}{0.0cm}
    \renewcommand{\arraystretch}{1.1}
    
    \newcommand{\middSec}[1]{\tiny #1}
    \newcommand{\middAll}[1]{\scalebox{0.83}[1.0]{$\uline{ #1 }$}}
    \newcommand{\middALL}[1]{\scalebox{0.83}[1.0]{$\uline{\bm{ #1 }}$}}
    \newcommand{\middOth}[1]{\scalebox{0.83}[1.0]{$ #1 $}}
    \newcommand{\middOTH}[1]{\scalebox{0.83}[1.0]{$\bm{ #1 }$}}
    \scriptsize
    \begin{tabularx}{\columnwidth}{@{\hspace{0.1cm}} X @{\hspace{0.0cm}} c @{\hspace{0.1cm}} c @{\hspace{0.0cm}} c @{\hspace{0.225cm}} r @{\hspace{0.225cm}} c @{\hspace{0.1cm}} c @{\hspace{0.0cm}} l @{\hspace{0.225cm}} c @{\hspace{0.1cm}} c @{\hspace{0.0cm}} l @{\hspace{0.225cm}} c @{\hspace{0.1cm}} c @{\hspace{0.0cm}} l @{\hspace{0.1cm}}}
        \toprule
            & \multicolumn{2}{c}{\sc Average} &&& \multicolumn{2}{c}{Laboratory} && \multicolumn{2}{c}{Synthetic} && \multicolumn{2}{c}{Real World} &
        \\ \cmidrule{2-3} \cmidrule{6-7} \cmidrule{9-10} \cmidrule{12-13}
            & \middSec{PSNR} & \middSec{SSIM} &&& \middSec{PSNR} & \middSec{SSIM} && \middSec{PSNR} & \middSec{SSIM} && \middSec{PSNR} & \middSec{SSIM} &
        \\ \midrule
            net. w/ context & \middOTH{36.93} & \middOTH{0.964} &&\multicolumn{1}{|c}{}& \middOTH{40.09} & \middOTH{0.971} && \middOTH{38.89} & \middOTH{0.981} && \middOTH{31.79} & \middOTH{0.939} &
        \\
            net. w/o context & \middOth{35.48} & \middOth{0.958} &&\multicolumn{1}{|c}{}& \middOth{39.07} & \middOth{0.968} && \middOth{35.69} & \middOth{0.967} && \middOth{31.68} & \middOTH{0.939} &
        \\
            bidirect. blending & \middOth{34.71} & \middOth{0.952} &&\multicolumn{1}{|c}{}& \middOth{38.25} & \middOth{0.966} && \middOth{34.27} & \middOth{0.953} && \middOth{31.59} & \middOth{0.938} &
        \\
            forward blending & \middOth{34.01} & \middOth{0.949} &&\multicolumn{1}{|c}{}& \middOth{37.94} & \middOth{0.965} && \middOth{33.38} & \middOth{0.950} && \middOth{30.70} & \middOth{0.933} &
        \\ \bottomrule
    \end{tabularx}\vspace{-0.2cm}
    \caption{Frame synthesis network vs pixel-wise blending.}\vspace{-0.5cm}
    \label{tbl:bidirectional}
\end{table}

Since the interpolation category of the Middlebury optical flow benchmark is typically used for assessing frame interpolation methods~\cite{Baker_OTHER_2011, Meyer_CVPR_2015, Niklaus_CVPR_2017, Niklaus_ICCV_2017}, we compare our approach with the methods that perform best on this interpolation benchmark and that are publicly available. Specifically, we select MDP-Flow2~\cite{Xu_PAMI_2012}, a classic optical flow method, as it ranks first among all the published methods on the Middlebury benchmark. We also select DeepFlow2~\cite{Weinzaepfel_ICCV_2013} as a representative deep learning-based optical flow algorithm. For these optical flow algorithms, we use the algorithm from Baker~\etal to produce interpolation results~\cite{Baker_OTHER_2011}. We furthermore select the recent SepConv~\cite{Niklaus_ICCV_2017} method which is based on adaptive separable convolutions, as well as the the phase-based interpolation approach from Meyer~\etal~\cite{Meyer_CVPR_2015}. In ablation experiments, we will additionally compare several variations of our proposed approach, for example, a version that does not use contextual information.

\begin{figure}\centering
    \setlength{\tabcolsep}{0.0cm}
    \setlength{\itemwidth}{2.75cm}

    \begin{tabularx}{\textwidth}{c @{\hspace{0.05cm}} c @{\hspace{0.05cm}} c}
            \includegraphics[width=\itemwidth]{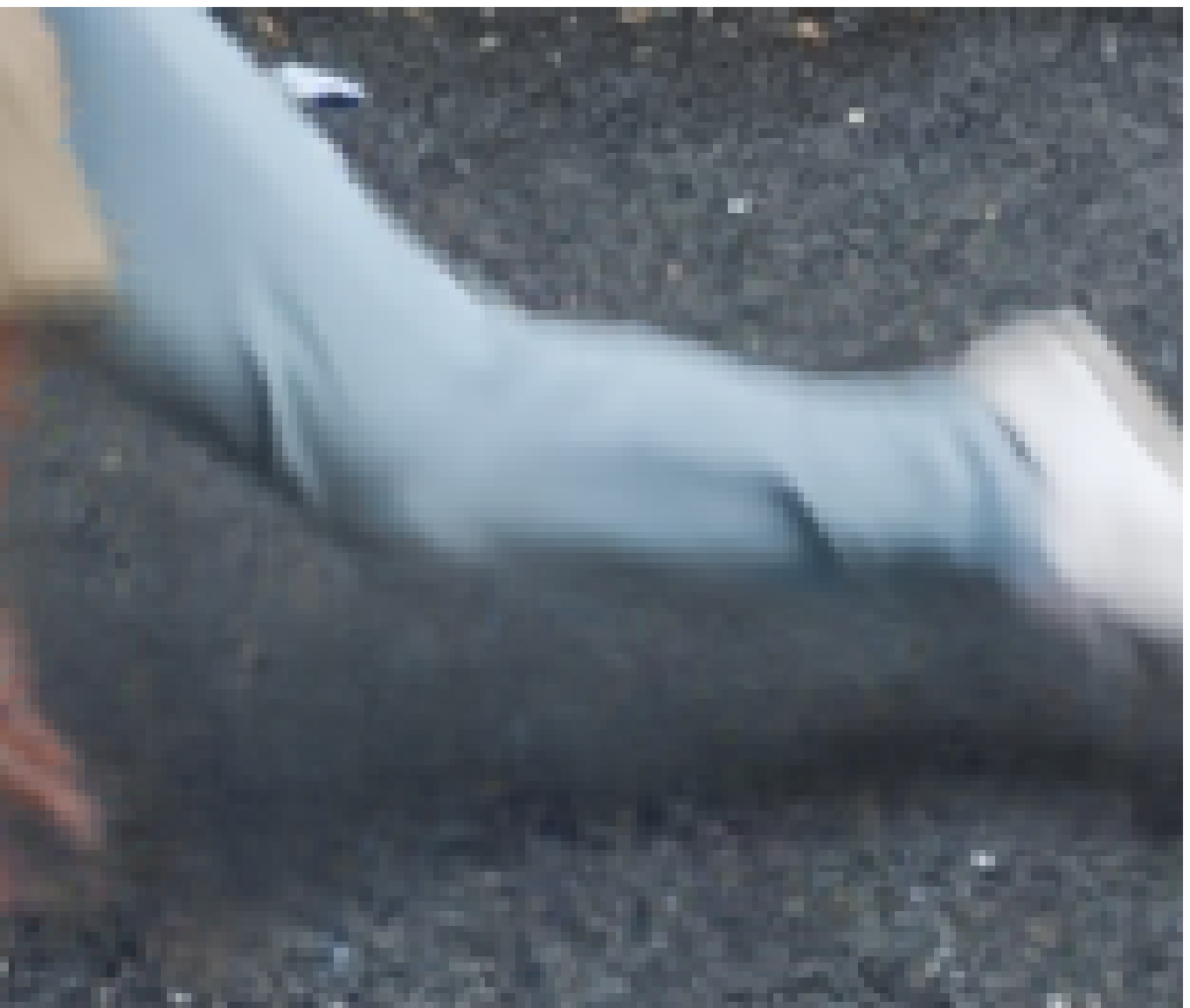}
        &
            \includegraphics[width=\itemwidth]{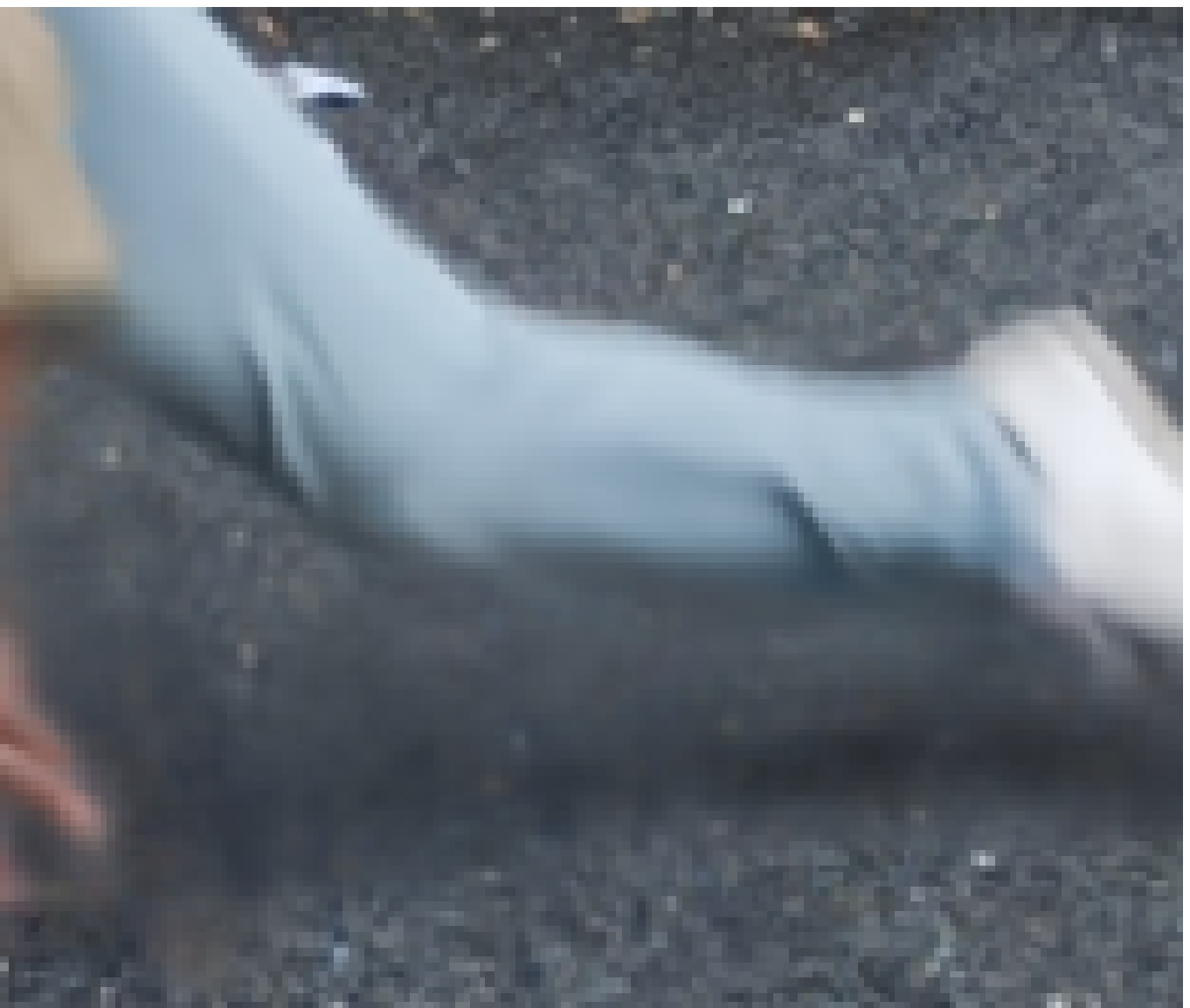}
        &
            \includegraphics[width=\itemwidth]{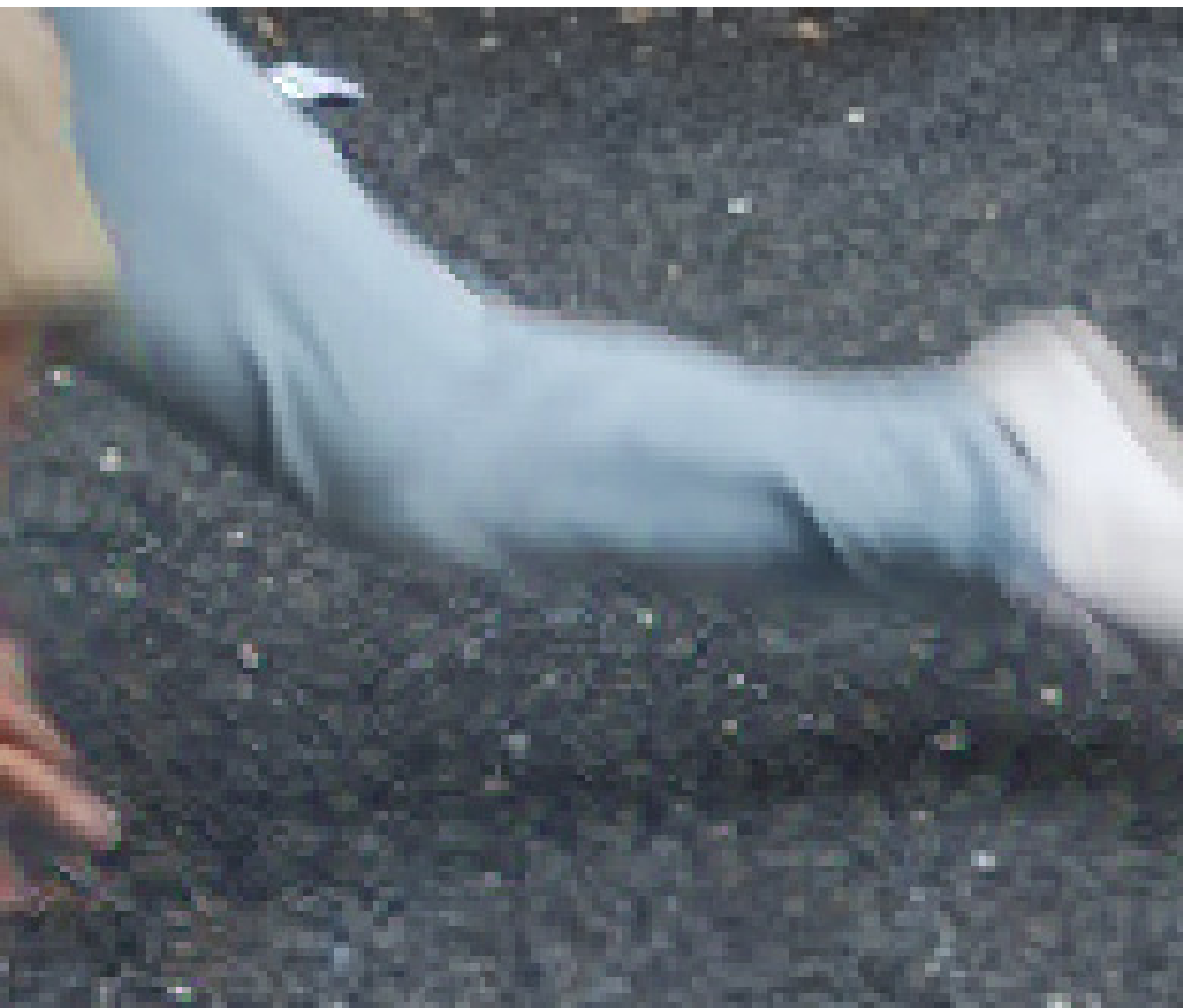}
        \vspace{-0.1cm} \\
    \end{tabularx}
    \begin{tabularx}{\textwidth}{c @{\hspace{0.05cm}} c @{\hspace{0.05cm}} c}
            \includegraphics[width=\itemwidth]{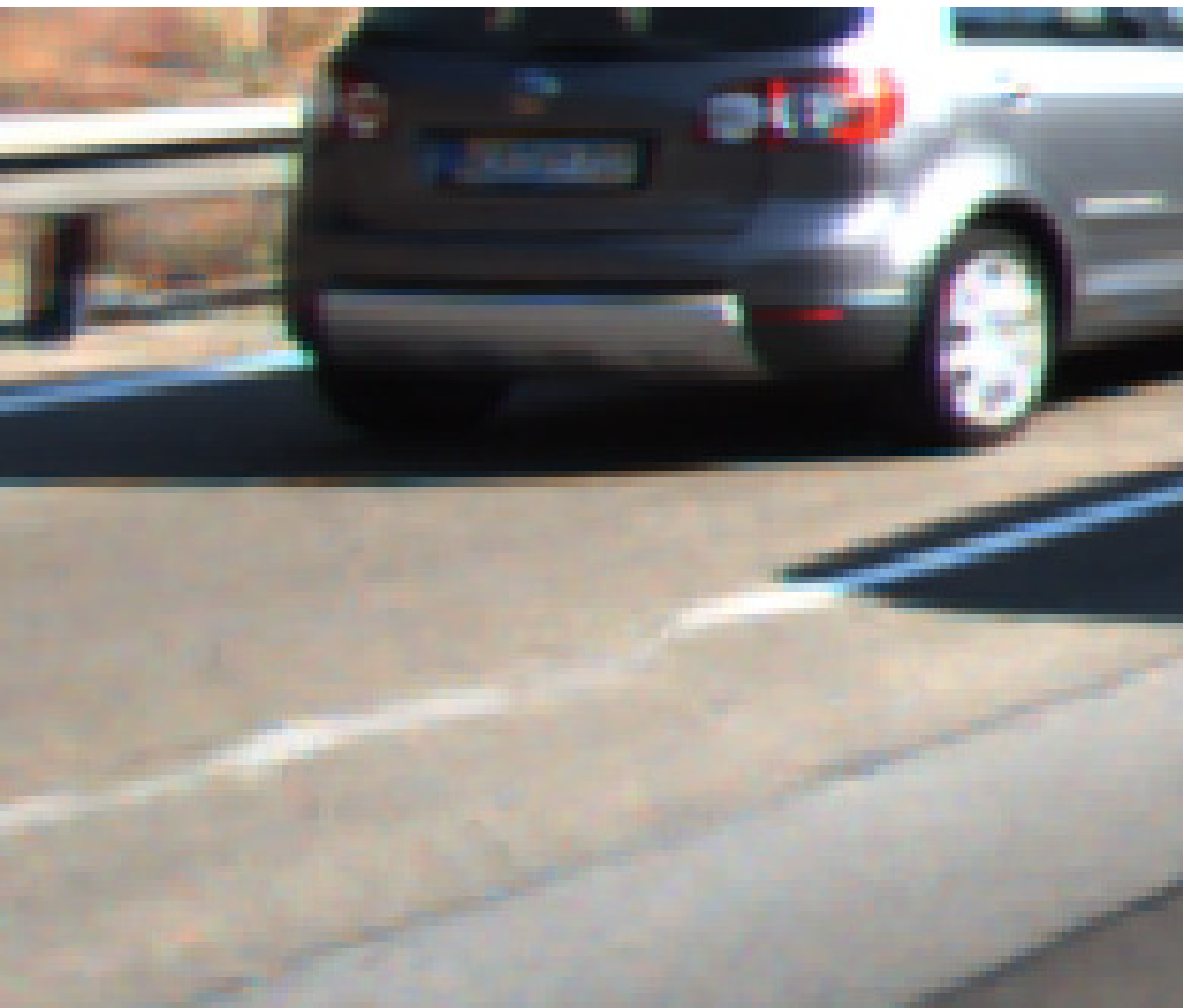}
        &
            \includegraphics[width=\itemwidth]{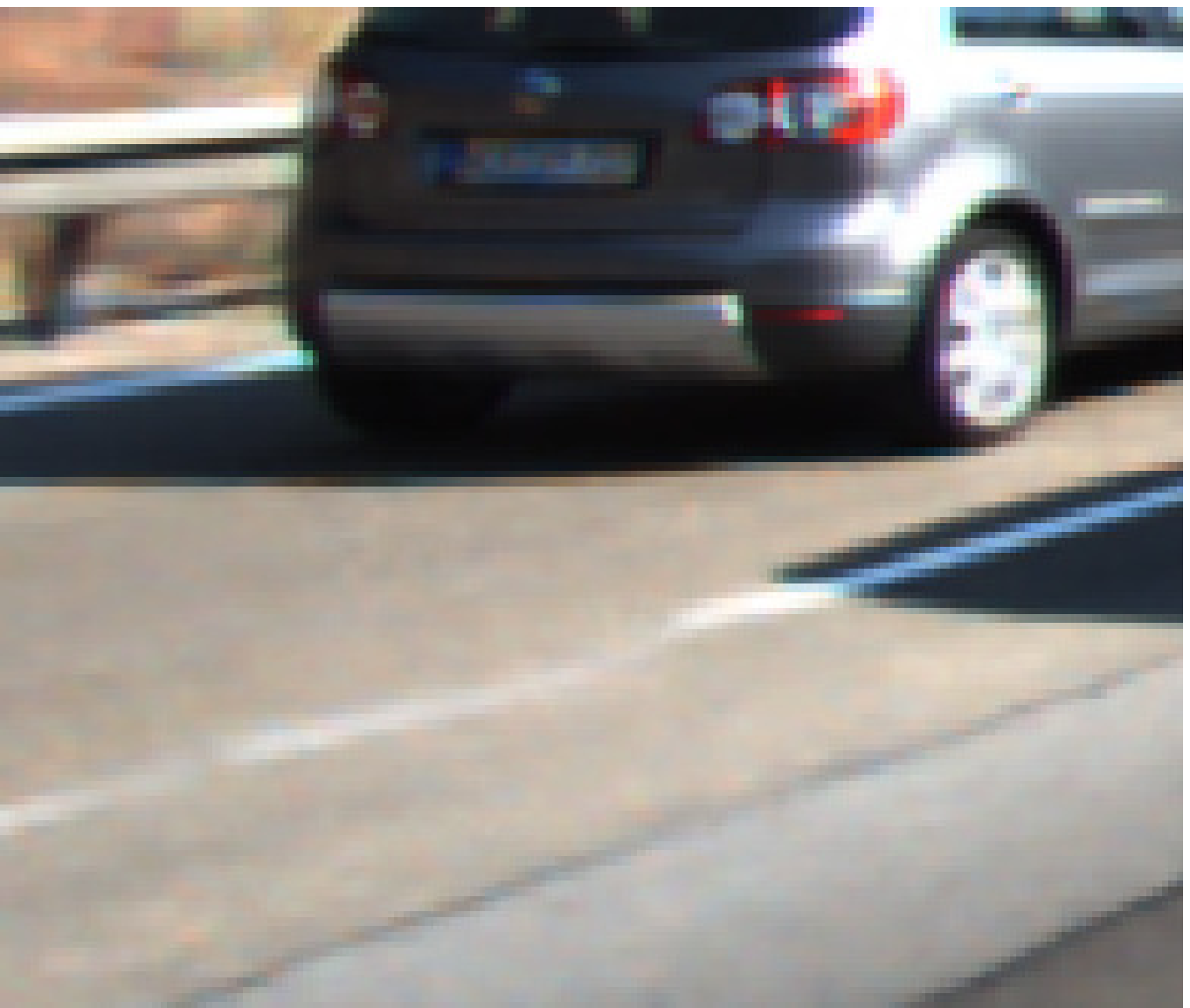}
        &
            \includegraphics[width=\itemwidth]{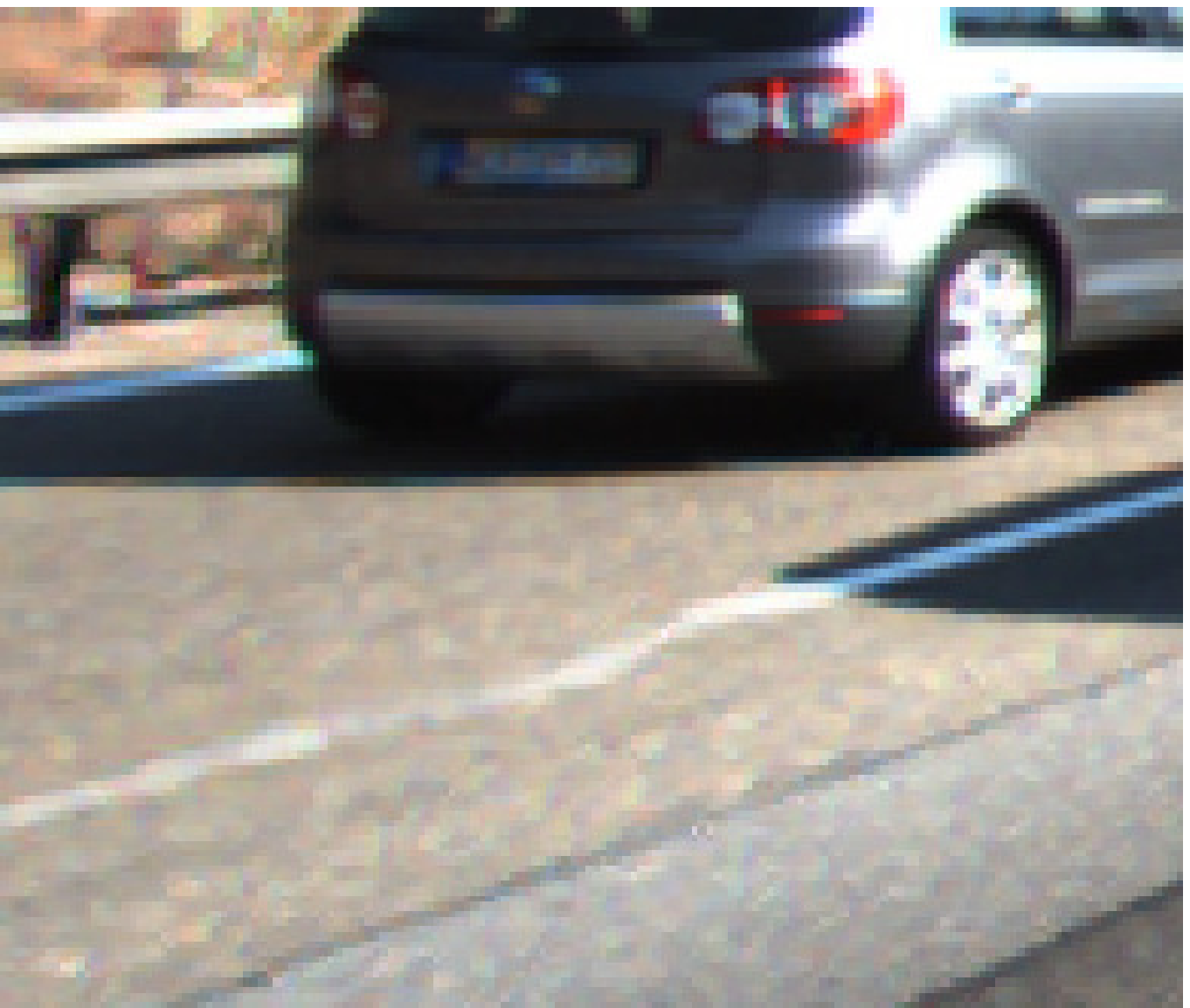}
        \vspace{-0.1cm} \\
            \footnotesize Ours - $\mathcal{L}_1$
        &
            \footnotesize Ours - $\mathcal{L}_{\textit{Lap}}$
        &
            \footnotesize Ours - $\mathcal{L}_F$
        \\
    \end{tabularx}\vspace{-0.2cm}
    \caption{Examples of using different loss functions.}\vspace{-0.2cm}
    \label{fig:loss}
\end{figure}

\begin{figure}\centering
    \setlength{\tabcolsep}{0.0cm}
    \setlength{\itemwidth}{2.75cm}

    \begin{tabularx}{\textwidth}{c @{\hspace{0.05cm}} c @{\hspace{0.05cm}} c @{\hspace{0.05cm}} c}
            \includegraphics[width=\itemwidth, trim={0.0cm 0.0cm 0.0cm 1.0cm}, clip]{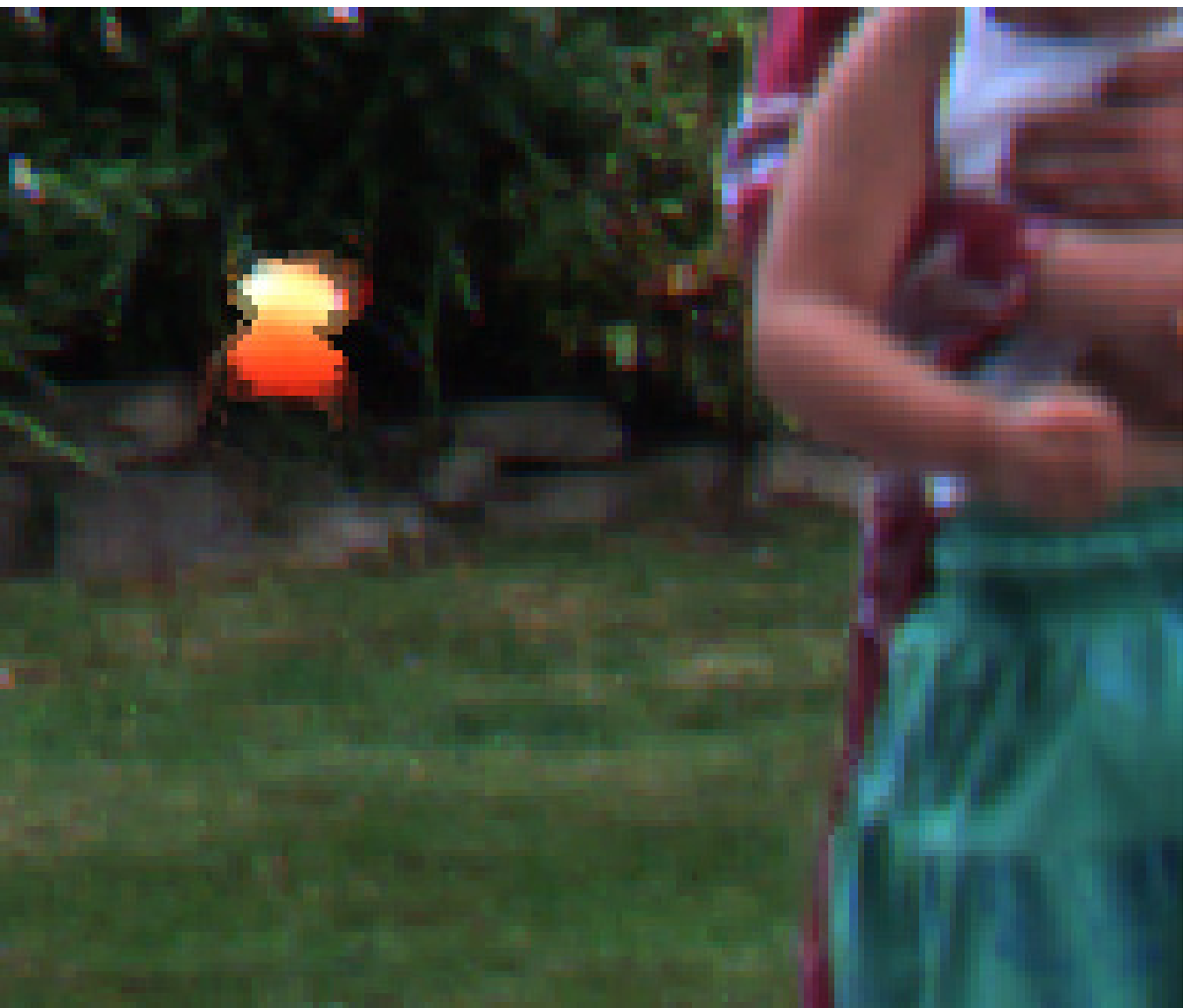}
        &
            \includegraphics[width=\itemwidth, trim={0.0cm 0.0cm 0.0cm 1.0cm}, clip]{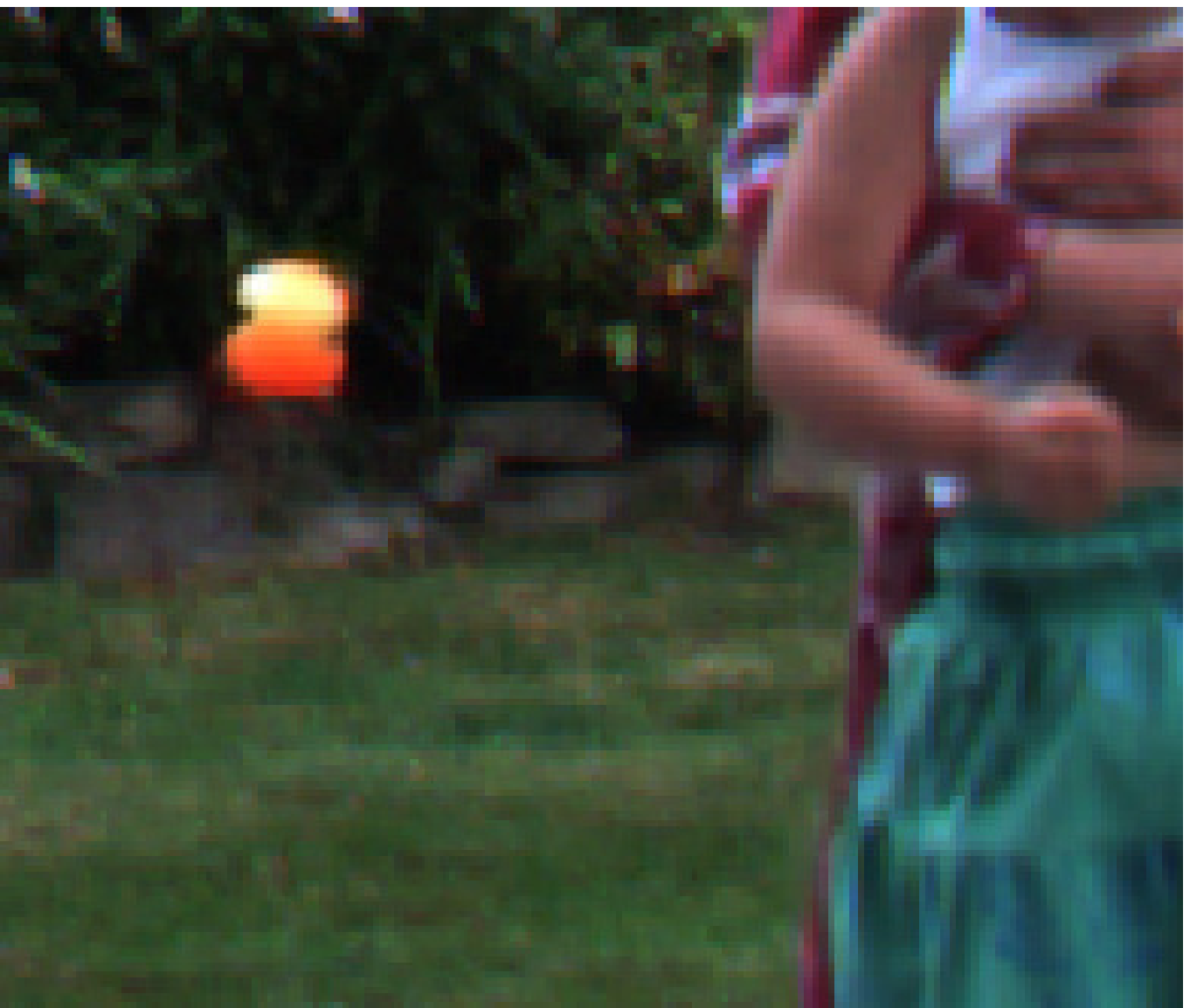}
        &
            \includegraphics[width=\itemwidth, trim={0.0cm 0.0cm 0.0cm 1.0cm}, clip]{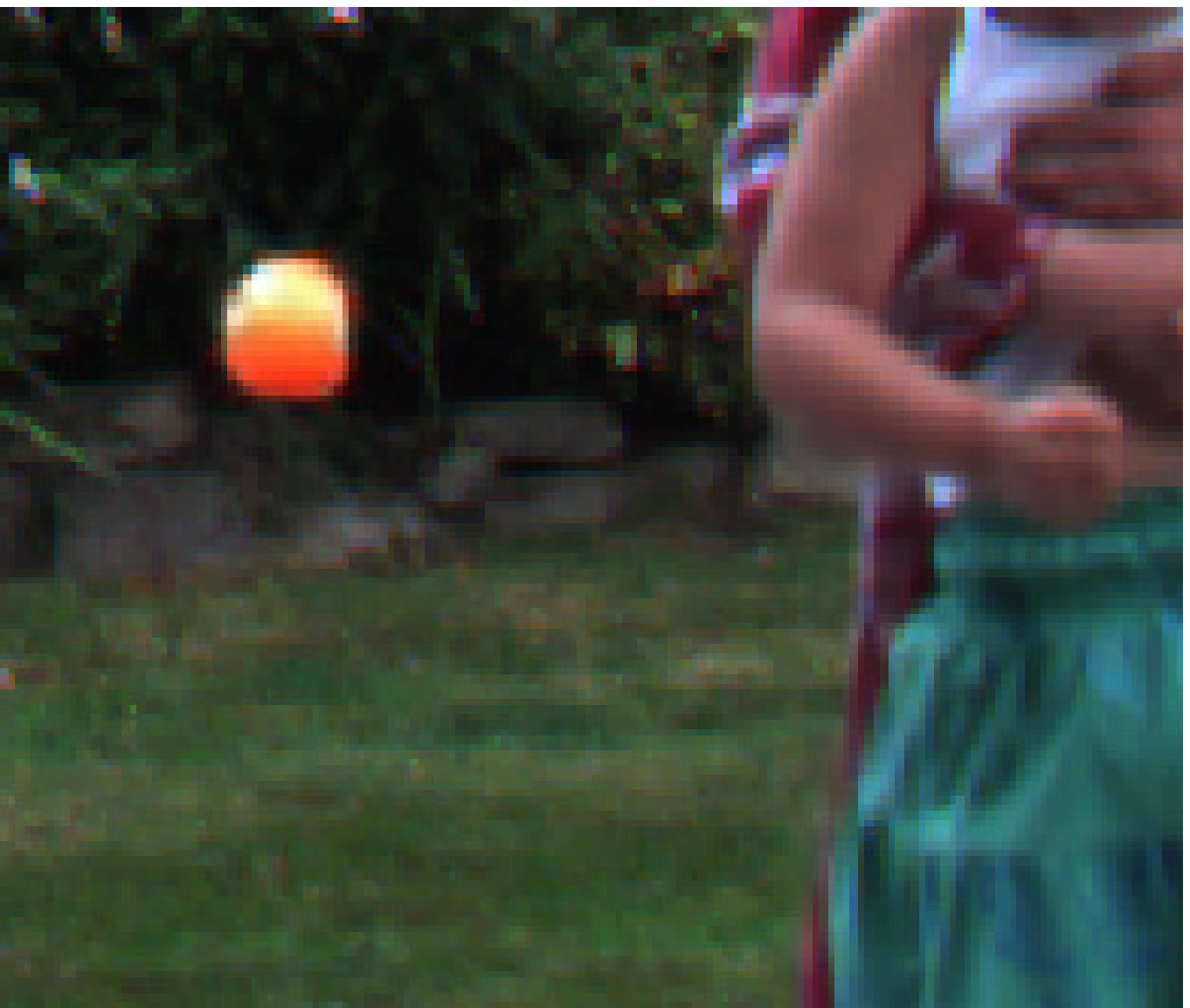}
        \vspace{-0.1cm} \\
    \end{tabularx}
    \begin{tabularx}{\textwidth}{c @{\hspace{0.05cm}} c @{\hspace{0.05cm}} c @{\hspace{0.05cm}} c}
            \includegraphics[width=\itemwidth, trim={0.0cm 0.0cm 0.0cm 1.0cm}, clip]{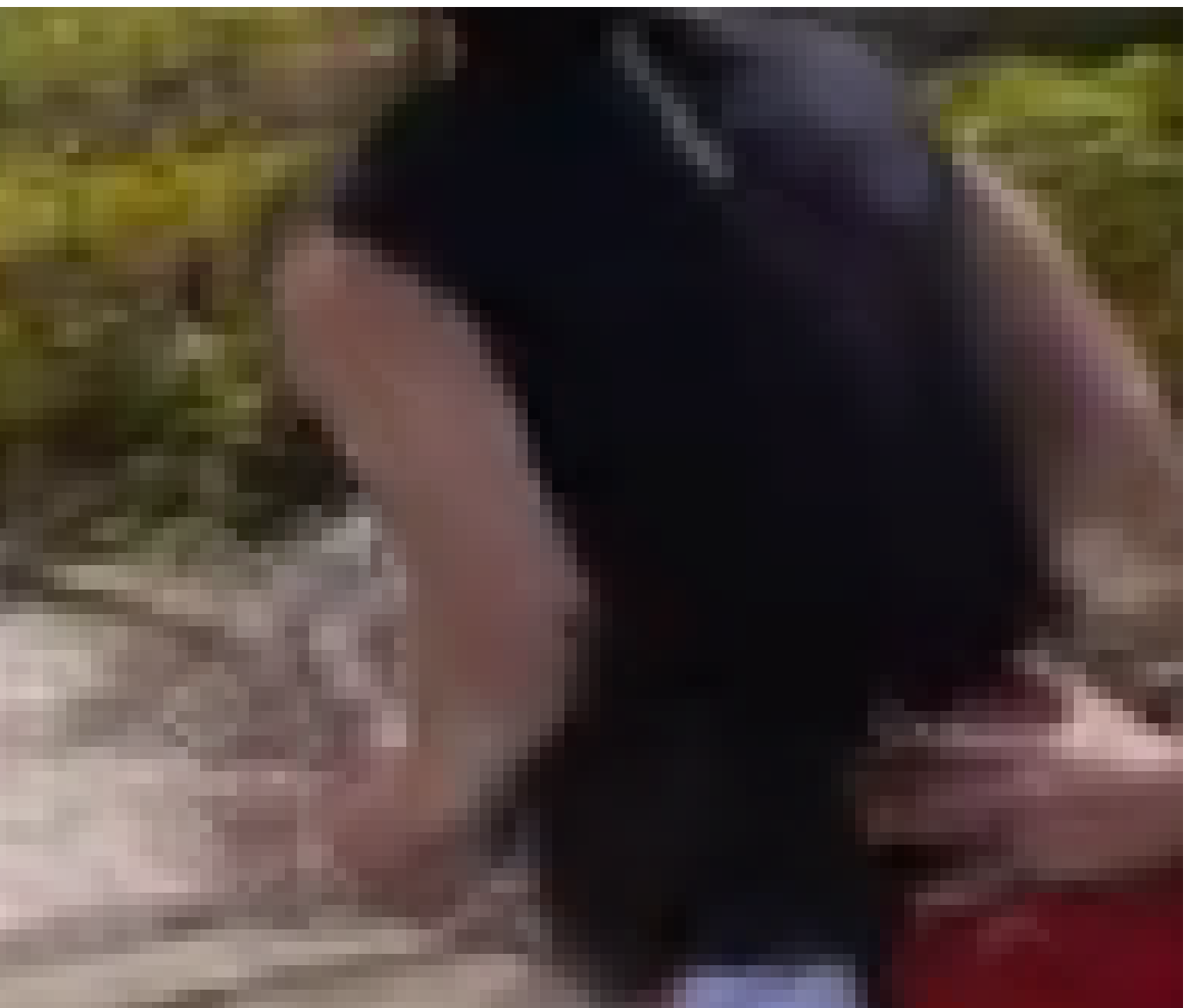}
        &
            \includegraphics[width=\itemwidth, trim={0.0cm 0.0cm 0.0cm 1.0cm}, clip]{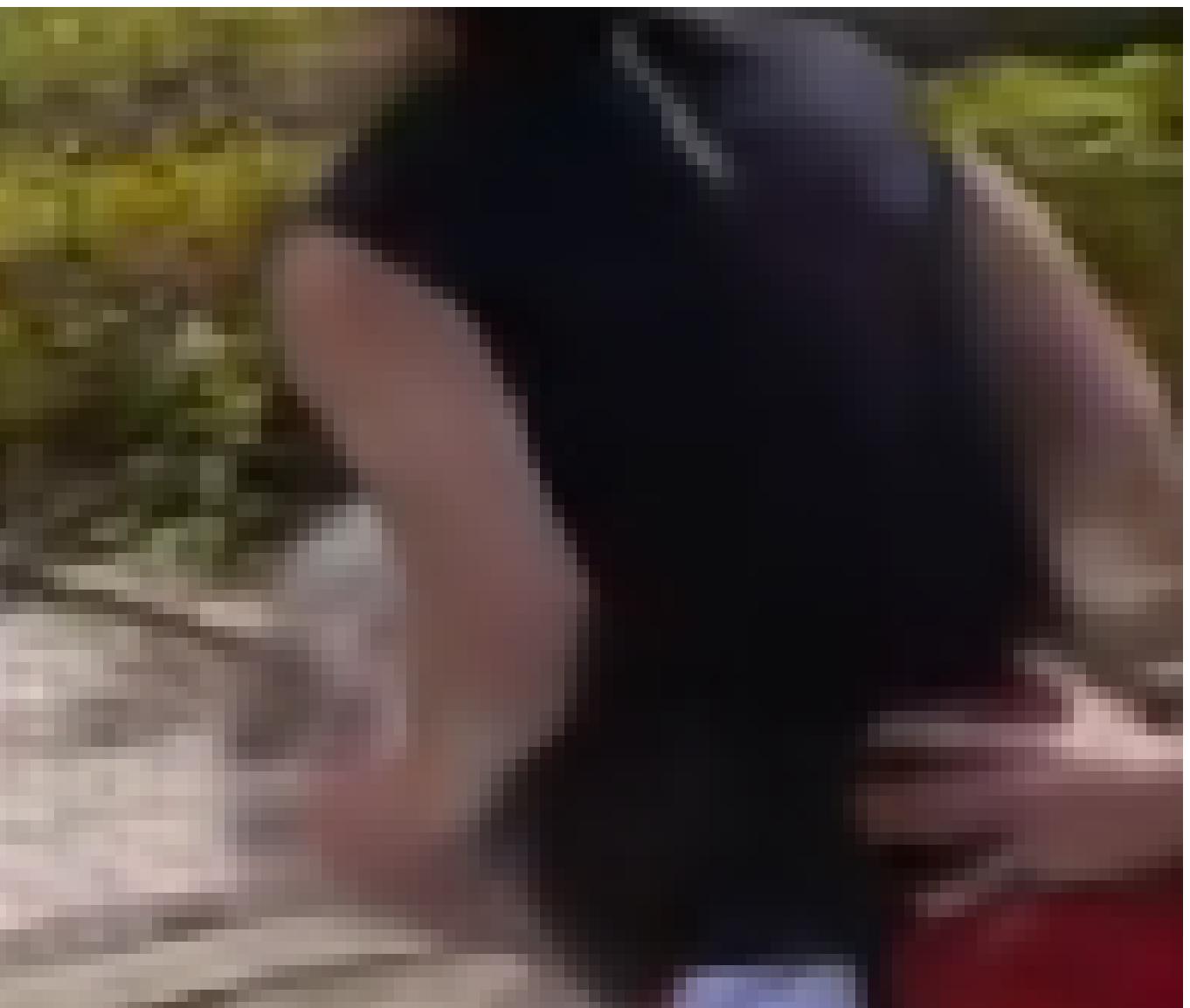}
        &
            \includegraphics[width=\itemwidth, trim={0.0cm 0.0cm 0.0cm 1.0cm}, clip]{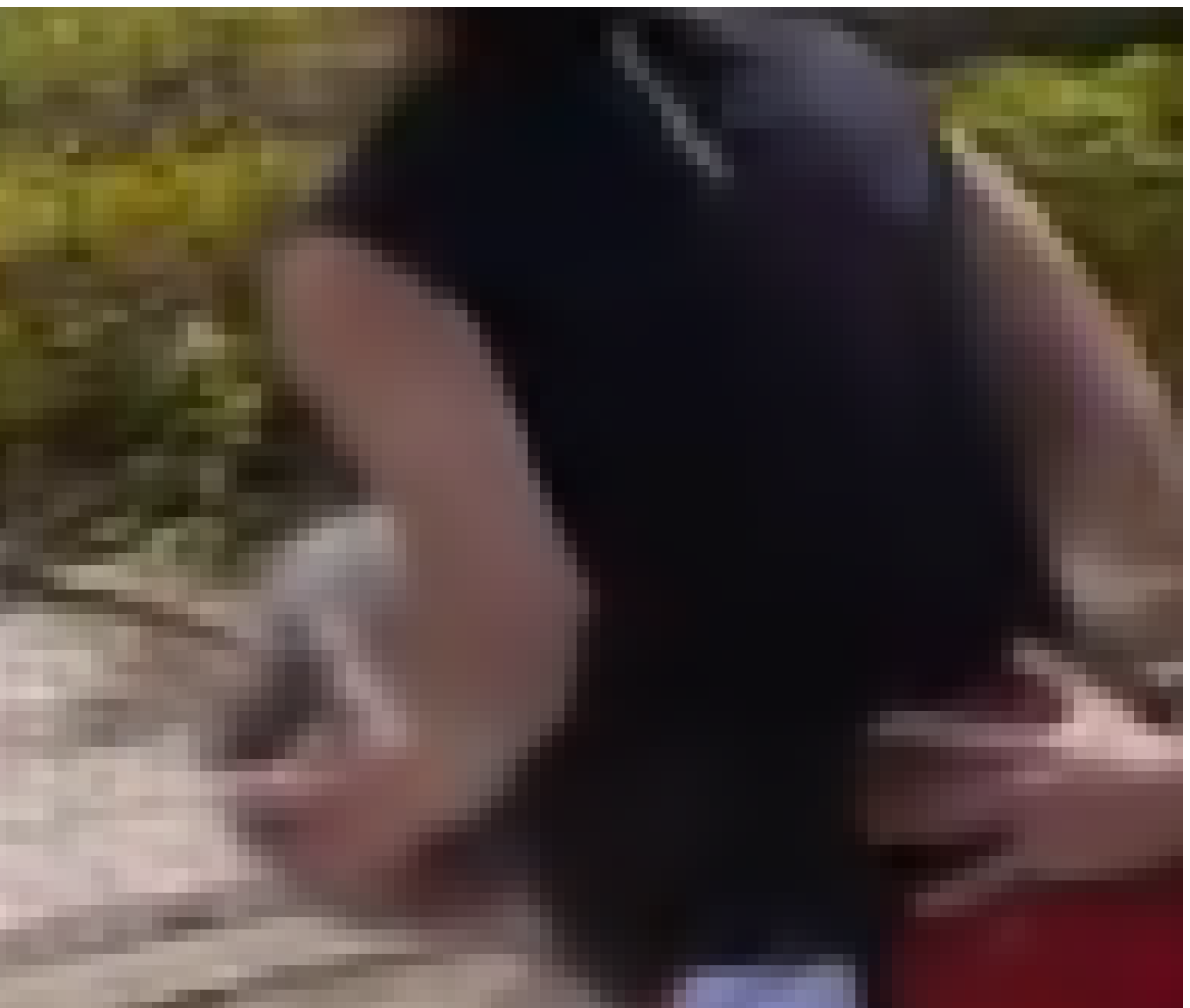}
        \vspace{-0.1cm} \\
            \footnotesize bidirect. blending
        &
            \footnotesize new w/o context
        &
            \footnotesize net. w/ context
        \\
    \end{tabularx}\vspace{-0.2cm}
    \caption{Examples for the ablation experiments.}\vspace{-0.5cm}
    \label{fig:ablation}
\end{figure}

\subsection{Ablation experiments}\label{subsec:ablation}

We first evaluate the different loss functions for training our frame synthesis neural network. We then discuss a few design choices and compare our method to several baseline versions in order to assess their effect.

We conduct the ablation experiments quantitatively and use the examples from the Middlebury optical flow benchmark that have publicly available ground truth interpolation results~\cite{Baker_OTHER_2011}. There are twelve such examples, which were either obtained in a lab environment with controlled lighting, synthetically rendered, or acquired by filming real-world scenes. We assess each category, consisting of four examples, separately and measure PSNR as well as SSIM~\cite{Wang_TIP_2004}.

\begin{table*}\begin{minipage}[b]{.4797\textwidth}
    \setlength{\tabcolsep}{0.0cm}
    \renewcommand{\arraystretch}{1.1}
    
    \newcommand{\middSec}[1]{\tiny #1}
    \newcommand{\middAll}[1]{\scalebox{0.83}[1.0]{$\uline{ #1 }$}}
    \newcommand{\middALL}[1]{\scalebox{0.83}[1.0]{$\uline{\bm{ #1 }}$}}
    \newcommand{\middOth}[1]{\scalebox{0.83}[1.0]{$ #1 $}}
    \newcommand{\middOTH}[1]{\scalebox{0.83}[1.0]{$\bm{ #1 }$}}
    \scriptsize
    \begin{tabularx}{\columnwidth}{@{\hspace{0.1cm}} X @{\hspace{0.0cm}} c @{\hspace{0.1cm}} c @{\hspace{0.0cm}} c @{\hspace{0.225cm}} r @{\hspace{0.225cm}} c @{\hspace{0.1cm}} c @{\hspace{0.0cm}} l @{\hspace{0.225cm}} c @{\hspace{0.1cm}} c @{\hspace{0.0cm}} l @{\hspace{0.225cm}} c @{\hspace{0.1cm}} c @{\hspace{0.0cm}} l @{\hspace{0.1cm}}}
        \toprule
            & \multicolumn{2}{c}{\sc Average} &&& \multicolumn{2}{c}{Laboratory} && \multicolumn{2}{c}{Synthetic} && \multicolumn{2}{c}{Real World} &
        \\ \cmidrule{2-3} \cmidrule{6-7} \cmidrule{9-10} \cmidrule{12-13}
            & \middSec{PSNR} & \middSec{SSIM} &&& \middSec{PSNR} & \middSec{SSIM} && \middSec{PSNR} & \middSec{SSIM} && \middSec{PSNR} & \middSec{SSIM} &
        \\ \midrule
            ResNet-18 - \tiny \verb|conv1| & \middOTH{36.93} & \middOTH{0.964} &&\multicolumn{1}{|c}{}& \middOTH{40.09} & \middOTH{0.971} && \middOTH{38.89} & \middOTH{0.981} && \middOth{31.79} & \middOth{0.939} &
        \\
            VGG-19 - \tiny \verb|conv1_2| & \middOth{36.77} & \middOTH{0.964} &&\multicolumn{1}{|c}{}& \middOth{39.97} & \middOTH{0.971} && \middOth{38.53} & \middOth{0.980} && \middOTH{31.81} & \middOTH{0.940} &
        \\
            VGG-19 - \tiny \verb|conv1_1| & \middOth{36.69} & \middOth{0.963} &&\multicolumn{1}{|c}{}& \middOth{39.90} & \middOTH{0.971} && \middOth{38.37} & \middOth{0.979} && \middOth{31.80} & \middOTH{0.940} &
        \\
            no context & \middOth{35.48} & \middOth{0.958} &&\multicolumn{1}{|c}{}& \middOth{39.07} & \middOth{0.968} && \middOth{35.69} & \middOth{0.967} && \middOth{31.68} & \middOth{0.939} &
        \\ \bottomrule
    \end{tabularx}\vspace{-0.2cm}
    \caption{Effect of contextual information.}\vspace{-0.1cm}
    \label{tbl:context}
\end{minipage}\qquad\begin{minipage}[b]{.4797\textwidth}
    \setlength{\tabcolsep}{0.0cm}
    \renewcommand{\arraystretch}{1.1}
    
    \newcommand{\middSec}[1]{\tiny #1}
    \newcommand{\middAll}[1]{\scalebox{0.83}[1.0]{$\uline{ #1 }$}}
    \newcommand{\middALL}[1]{\scalebox{0.83}[1.0]{$\uline{\bm{ #1 }}$}}
    \newcommand{\middOth}[1]{\scalebox{0.83}[1.0]{$ #1 $}}
    \newcommand{\middOTH}[1]{\scalebox{0.83}[1.0]{$\bm{ #1 }$}}
    \scriptsize
    \begin{tabularx}{\columnwidth}{@{\hspace{0.1cm}} X @{\hspace{0.0cm}} c @{\hspace{0.1cm}} c @{\hspace{0.0cm}} c @{\hspace{0.225cm}} r @{\hspace{0.225cm}} c @{\hspace{0.1cm}} c @{\hspace{0.0cm}} l @{\hspace{0.225cm}} c @{\hspace{0.1cm}} c @{\hspace{0.0cm}} l @{\hspace{0.225cm}} c @{\hspace{0.1cm}} c @{\hspace{0.0cm}} l @{\hspace{0.1cm}}}
        \toprule
            & \multicolumn{2}{c}{\sc Average} &&& \multicolumn{2}{c}{Laboratory} && \multicolumn{2}{c}{Synthetic} && \multicolumn{2}{c}{Real World} &
        \\ \cmidrule{2-3} \cmidrule{6-7} \cmidrule{9-10} \cmidrule{12-13}
            & \middSec{PSNR} & \middSec{SSIM} &&& \middSec{PSNR} & \middSec{SSIM} && \middSec{PSNR} & \middSec{SSIM} && \middSec{PSNR} & \middSec{SSIM} &
        \\ \midrule
            flow of PWC-Net & \middOTH{36.93} & \middOTH{0.964} &&\multicolumn{1}{|c}{}& \middOTH{40.09} & \middOTH{0.971} && \middOTH{38.89} & \middOTH{0.981} && \middOTH{31.79} & \middOTH{0.939} &
        \\
            flow of SPyNet & \middOth{35.78} & \middOth{0.959} &&\multicolumn{1}{|c}{}& \middOth{39.89} & \middOth{0.971} && \middOth{36.87} & \middOth{0.971} && \middOth{30.58} & \middOth{0.935} &
        \\
            H.264 motion vec. & \middOth{35.44} & \middOth{0.957} &&\multicolumn{1}{|c}{}& \middOth{39.42} & \middOth{0.968} && \middOth{36.83} & \middOth{0.975} && \middOth{30.07} & \middOth{0.927} &
        \\
            no flow / warping & \middOth{34.98} & \middOth{0.955} &&\multicolumn{1}{|c}{}& \middOth{38.67} & \middOth{0.963} && \middOth{35.98} & \middOth{0.966} && \middOth{30.30} & \middOth{0.935} &
        \\ \bottomrule
    \end{tabularx}\vspace{-0.2cm}
    \caption{Effect of different optical flow algorithms.}\vspace{-0.1cm}
    \label{tbl:flow}
\end{minipage}\end{table*}

\begin{table*}\centering
    \setlength{\tabcolsep}{0.0cm}
    \renewcommand{\arraystretch}{1.1}
    
    \newcommand{\middSec}[1]{\scriptsize #1}
    \newcommand{\middAll}[1]{\scalebox{0.75}[1.0]{$\uline{ #1 }$}}
    \newcommand{\middALL}[1]{\scalebox{0.75}[1.0]{$\uline{\bm{ #1 }}$}}
    \newcommand{\middOth}[1]{\scalebox{0.75}[1.0]{$ #1 $}}
    \newcommand{\middOTH}[1]{\scalebox{0.75}[1.0]{$\bm{ #1 }$}}
    \scriptsize
    \begin{tabularx}{\textwidth}{@{\hspace{0.1cm}} X @{\hspace{0.0cm}} c @{\hspace{0.1cm}} c @{\hspace{0.1cm}} c @{\hspace{0.0cm}} c @{\hspace{0.225cm}} r @{\hspace{0.225cm}} c @{\hspace{0.1cm}} c @{\hspace{0.1cm}} c @{\hspace{0.0cm}} l @{\hspace{0.225cm}} c @{\hspace{0.1cm}} c @{\hspace{0.1cm}} c @{\hspace{0.0cm}} l @{\hspace{0.225cm}} c @{\hspace{0.1cm}} c @{\hspace{0.1cm}} c @{\hspace{0.0cm}} l @{\hspace{0.225cm}} c @{\hspace{0.1cm}} c @{\hspace{0.1cm}} c @{\hspace{0.0cm}} l @{\hspace{0.225cm}} c @{\hspace{0.1cm}} c @{\hspace{0.1cm}} c @{\hspace{0.0cm}} l @{\hspace{0.225cm}} c @{\hspace{0.1cm}} c @{\hspace{0.1cm}} c @{\hspace{0.0cm}} l @{\hspace{0.225cm}} c @{\hspace{0.1cm}} c @{\hspace{0.1cm}} c @{\hspace{0.0cm}} l @{\hspace{0.225cm}} c @{\hspace{0.1cm}} c @{\hspace{0.1cm}} c @{\hspace{0.0cm}} l @{\hspace{0.1cm}}}
        \toprule
            & \multicolumn{3}{c}{\sc Average} &&& \multicolumn{3}{c}{Mequon} && \multicolumn{3}{c}{Schefflera} && \multicolumn{3}{c}{Urban} && \multicolumn{3}{c}{Teddy} && \multicolumn{3}{c}{Backyard} && \multicolumn{3}{c}{Basketball} && \multicolumn{3}{c}{Dumptruck} && \multicolumn{3}{c}{Evergreen} &
        \\ \cmidrule{2-4} \cmidrule{7-9} \cmidrule{11-13} \cmidrule{15-17} \cmidrule{19-21} \cmidrule{23-25} \cmidrule{27-29} \cmidrule{31-33} \cmidrule{35-37}
            & \middSec{all} & \middSec{disc.} & \middSec{unt.} &&& \middSec{all} & \middSec{disc.} & \middSec{unt.} && \middSec{all} & \middSec{disc.} & \middSec{unt.} && \middSec{all} & \middSec{disc.} & \middSec{unt.} && \middSec{all} & \middSec{disc.} & \middSec{unt.} && \middSec{all} & \middSec{disc.} & \middSec{unt.} && \middSec{all} & \middSec{disc.} & \middSec{unt.} && \middSec{all} & \middSec{disc.} & \middSec{unt.} && \middSec{all} & \middSec{disc.} & \middSec{unt.} &
        \\ \midrule
            Ours - $\mathcal{L}_{\textit{Lap}}$ & \middALL{5.28} & \middOTH{8.00} & \middOth{2.19} &&\multicolumn{1}{|c}{}& \middALL{2.24} & \middOTH{3.72} & \middOTH{1.04} && \middALL{2.96} & \middOTH{4.16} & \middOth{1.35} && \middAll{4.32} & \middOTH{3.42} & \middOth{3.18} && \middALL{4.21} & \middOTH{5.46} & \middOTH{3.00} && \middALL{9.6} & \middOTH{11.9} & \middOth{3.46} && \middALL{5.22} & \middOTH{9.8} & \middOth{2.22} && \middAll{7.02} & \middOTH{15.4} & \middOth{1.58} && \middAll{6.66} & \middOTH{10.2} & \middOth{1.69} &
        \\
            SepConv - $\mathcal{L}_1$ & \middAll{5.61} & \middOth{8.74} & \middOth{2.33} &&\multicolumn{1}{|c}{}& \middAll{2.52} & \middOth{4.83} & \middOth{1.11} && \middAll{3.56} & \middOth{5.04} & \middOth{1.90} && \middAll{4.17} & \middOth{4.15} & \middOth{2.86} && \middAll{5.41} & \middOth{6.81} & \middOth{3.88} && \middAll{10.2} & \middOth{12.8} & \middOth{3.37} && \middAll{5.47} & \middOth{10.4} & \middOTH{2.21} && \middAll{6.88} & \middOth{15.6} & \middOth{1.72} && \middALL{6.63} & \middOth{10.3} & \middOTH{1.62} &
        \\
            SepConv - $\mathcal{L}_F$ & \middAll{5.81} & \middOth{9.04} & \middOth{2.40} &&\multicolumn{1}{|c}{}& \middAll{2.60} & \middOth{5.00} & \middOth{1.19} && \middAll{3.87} & \middOth{5.50} & \middOth{2.07} && \middAll{4.38} & \middOth{4.29} & \middOth{2.73} && \middAll{5.78} & \middOth{7.16} & \middOth{3.94} && \middAll{10.1} & \middOth{12.7} & \middOth{3.39} && \middAll{5.98} & \middOth{11.4} & \middOth{2.42} && \middALL{6.85} & \middOth{15.5} & \middOth{1.78} && \middAll{6.90} & \middOth{10.8} & \middOth{1.65} &
        \\
            MDP-Flow2 & \middAll{5.83} & \middOth{9.69} & \middOth{2.15} &&\multicolumn{1}{|c}{}& \middAll{2.89} & \middOth{5.38} & \middOth{1.19} && \middAll{3.47} & \middOth{5.07} & \middOTH{1.26} && \middAll{3.66} & \middOth{6.10} & \middOth{2.48} && \middAll{5.20} & \middOth{7.48} & \middOth{3.14} && \middAll{10.2} & \middOth{12.8} & \middOth{3.61} && \middAll{6.13} & \middOth{11.8} & \middOth{2.31} && \middAll{7.36} & \middOth{16.8} & \middOTH{1.49} && \middAll{7.75} & \middOth{12.1} & \middOth{1.69} &
        \\
            DeepFlow2 & \middAll{6.02} & \middOth{9.94} & \middOTH{2.06} &&\multicolumn{1}{|c}{}& \middAll{2.99} & \middOth{5.65} & \middOth{1.22} && \middAll{3.88} & \middOth{5.79} & \middOth{1.48} && \middALL{3.62} & \middOth{6.03} & \middOTH{1.34} && \middAll{5.38} & \middOth{7.44} & \middOth{3.22} && \middAll{11.0} & \middOth{13.8} & \middOth{3.67} && \middAll{5.83} & \middOth{11.2} & \middOth{2.25} && \middAll{7.60} & \middOth{17.4} & \middOth{1.50} && \middAll{7.82} & \middOth{12.2} & \middOth{1.77} &
        \\ \bottomrule
    \end{tabularx}\vspace{-0.2cm}
    \caption{Evaluation on the Middlebury benchmark. \emph{disc.}: regions with discontinuous motion. \emph{unt.}: textureless regions.}\vspace{-0.4cm}
    \label{tbl:hidden}
\end{table*}

\begin{figure*}\centering
    \setlength{\tabcolsep}{0.0cm}
    \setlength{\itemwidth}{2.14cm}

    \begin{tabularx}{\textwidth}{c @{\hspace{0.05cm}} c @{\hspace{0.05cm}} c @{\hspace{0.05cm}} c @{\hspace{0.05cm}} c @{\hspace{0.05cm}} c @{\hspace{0.05cm}} c @{\hspace{0.05cm}} c}
            \includegraphics[width=\itemwidth]{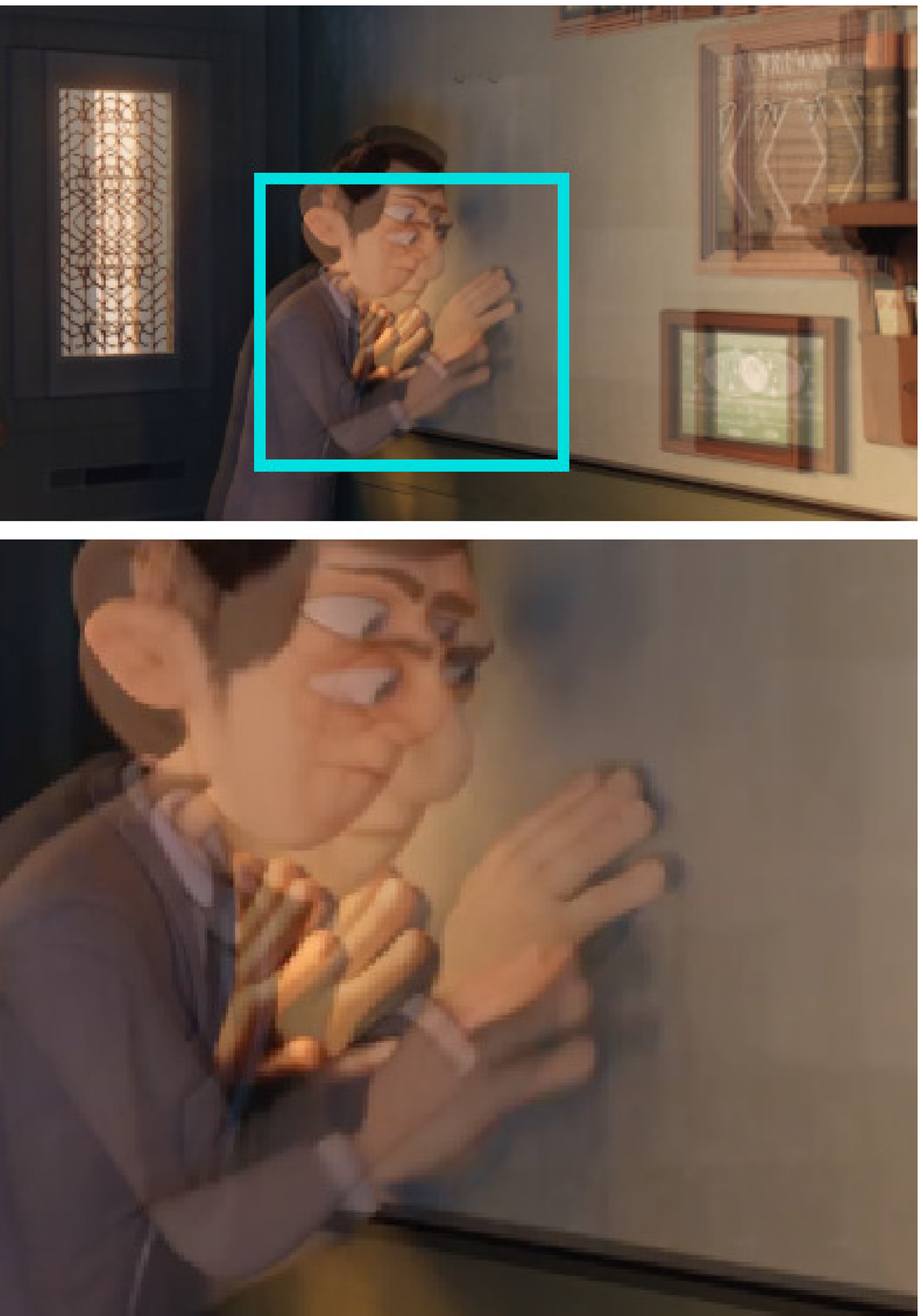}
        &
            \includegraphics[width=\itemwidth]{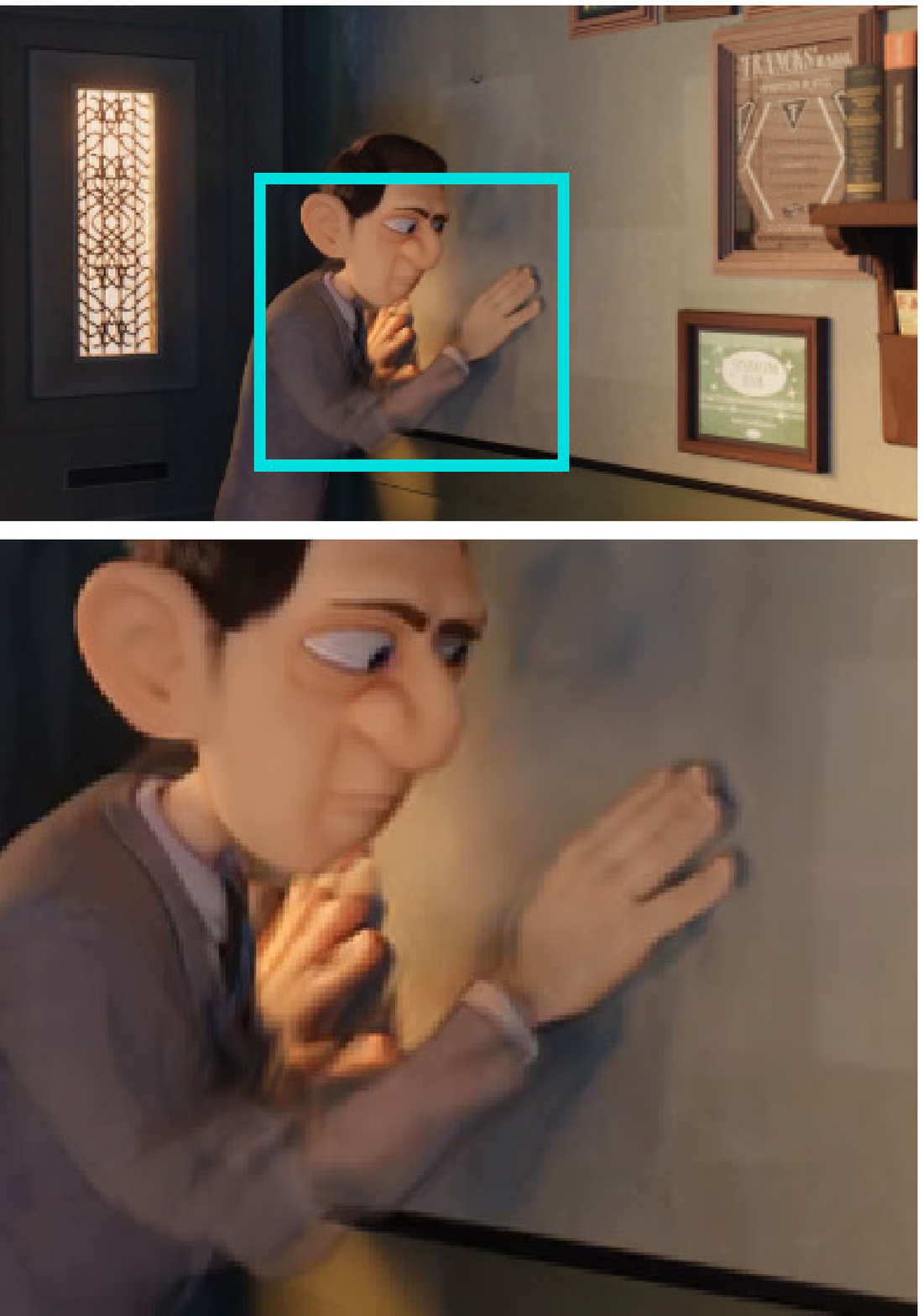}
        &
            \includegraphics[width=\itemwidth]{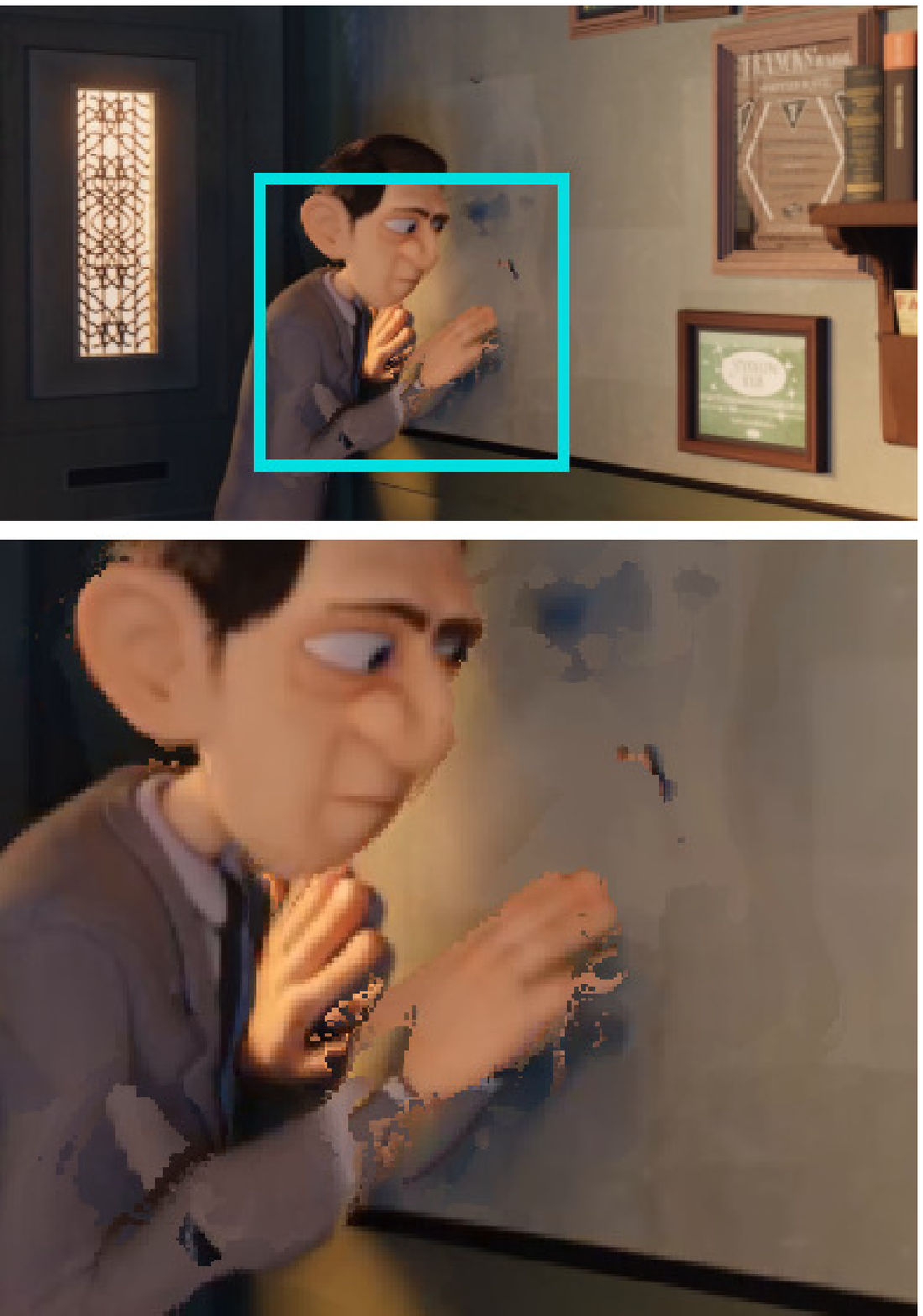}
        &
            \includegraphics[width=\itemwidth]{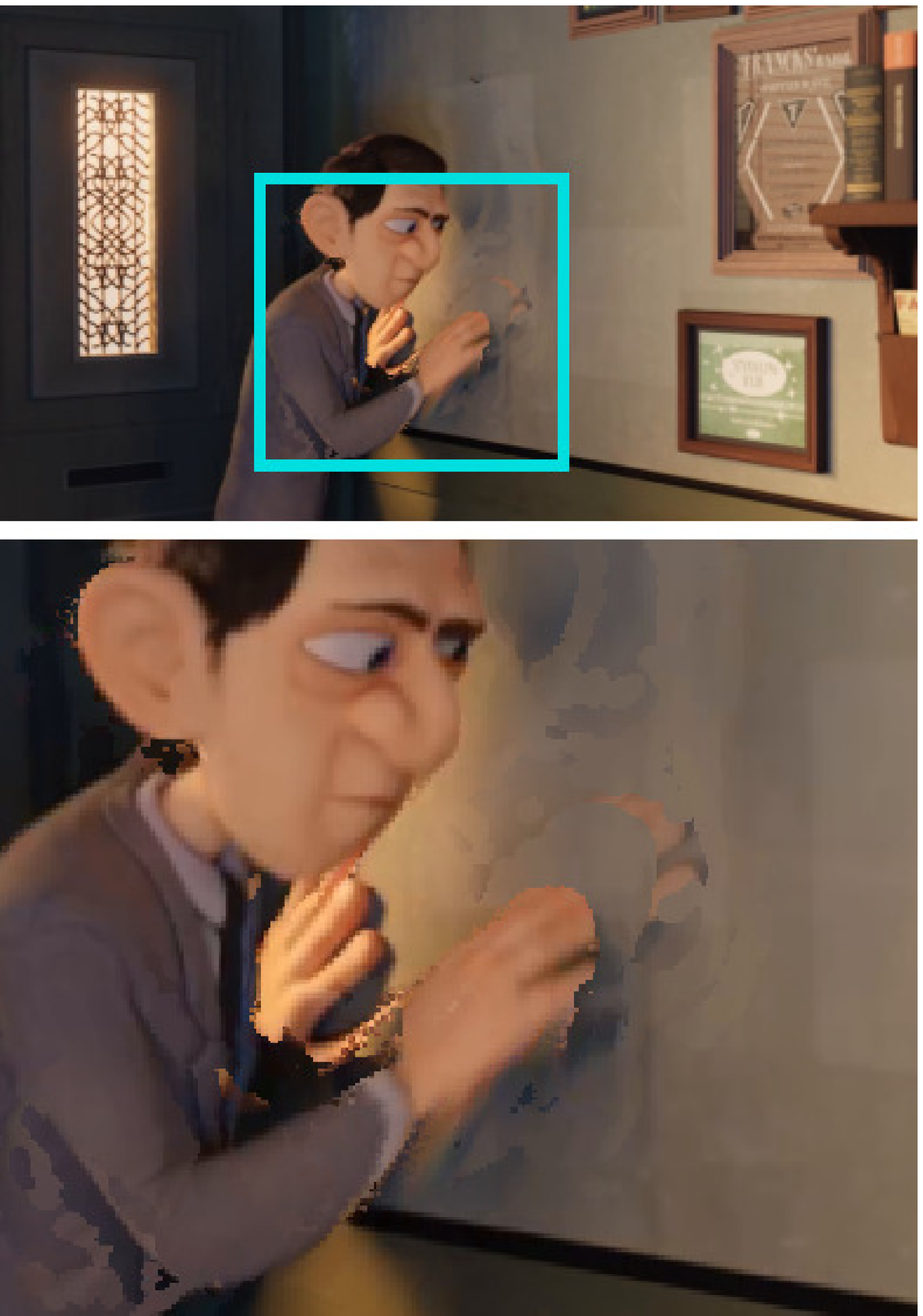}
        &
            \includegraphics[width=\itemwidth]{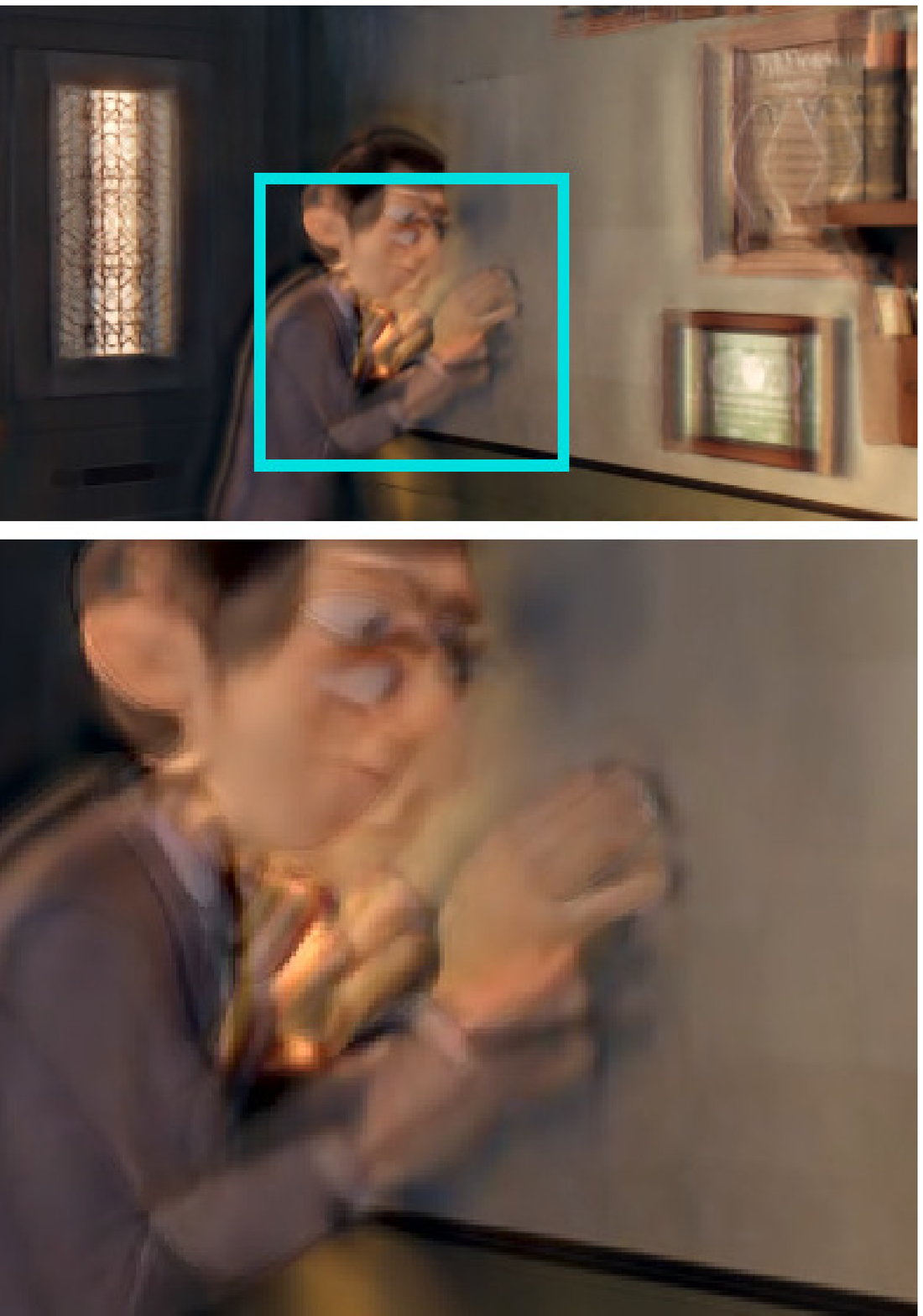}
        &
            \includegraphics[width=\itemwidth]{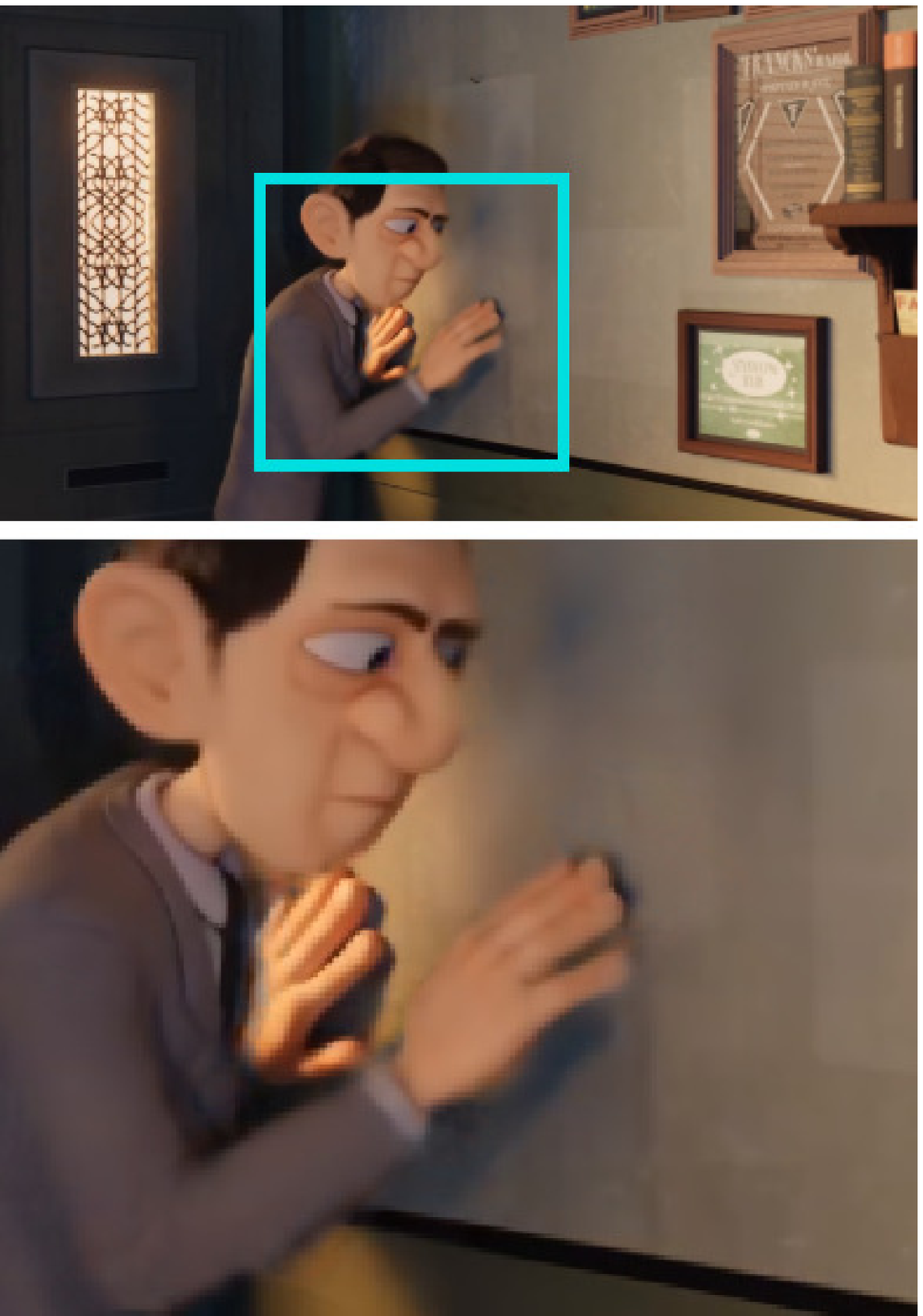}
        &
            \includegraphics[width=\itemwidth]{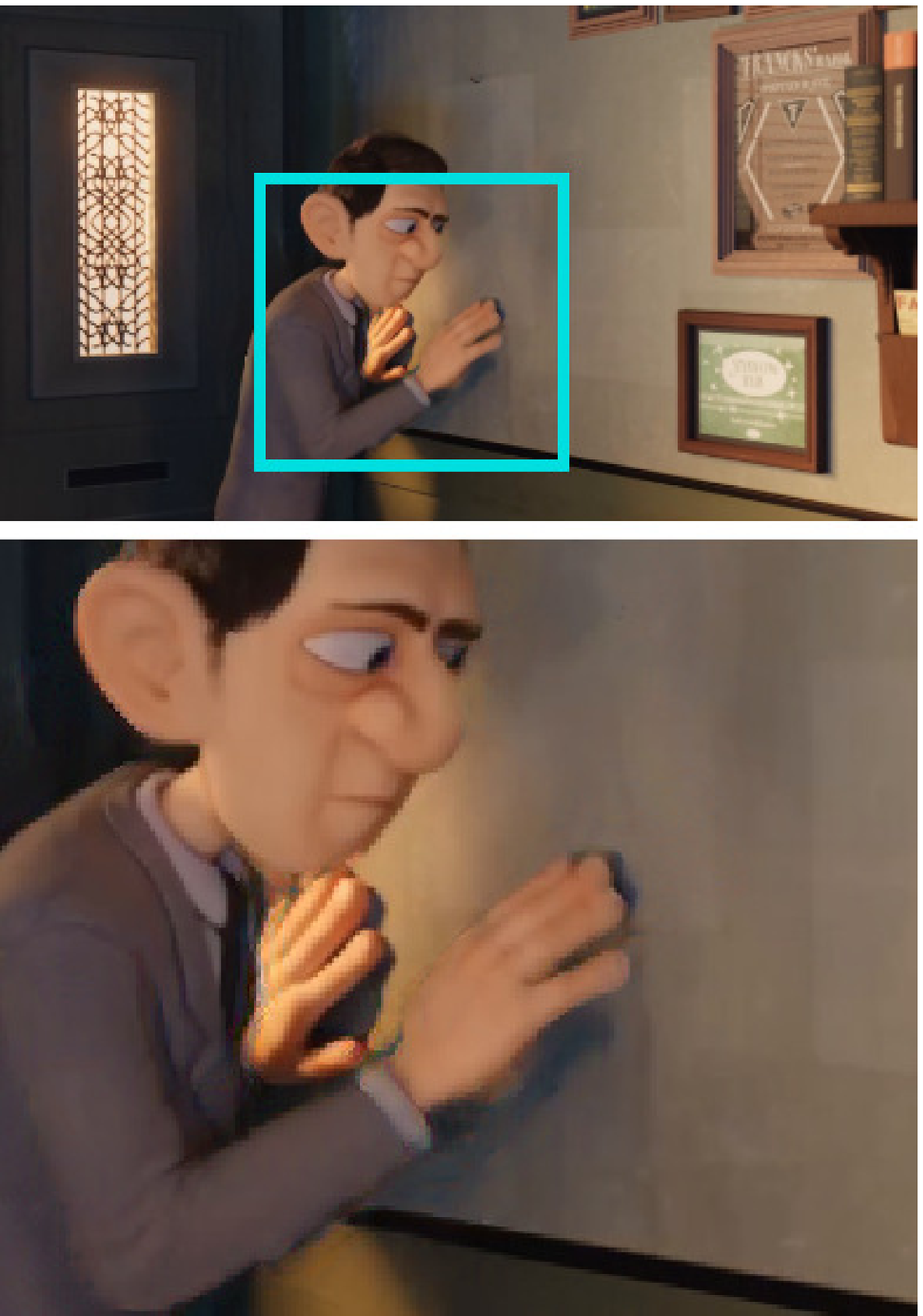}
        &
            \includegraphics[width=\itemwidth]{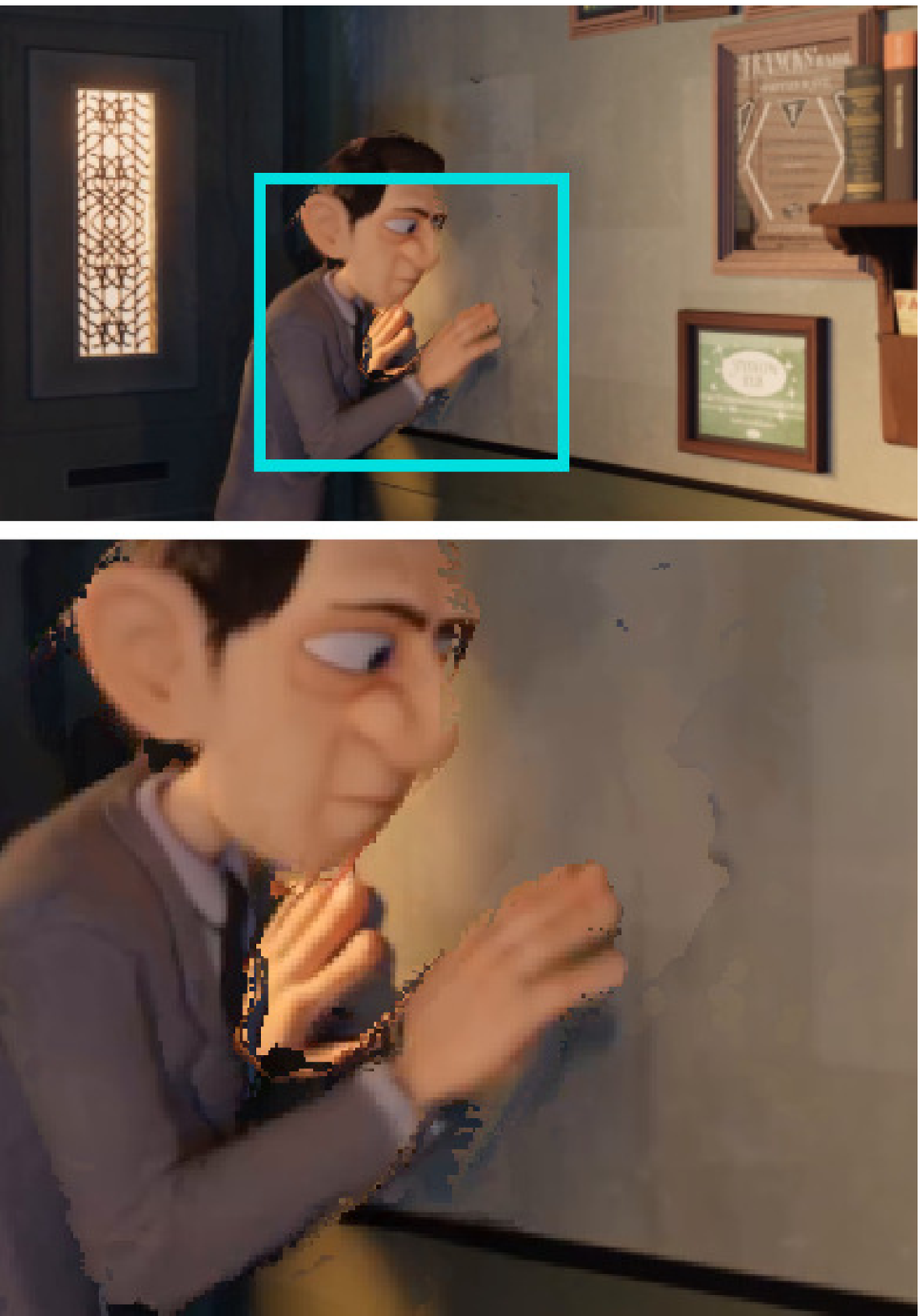}
        \vspace{-0.1cm} \\
    \end{tabularx}
    \begin{tabularx}{\textwidth}{c @{\hspace{0.05cm}} c @{\hspace{0.05cm}} c @{\hspace{0.05cm}} c @{\hspace{0.05cm}} c @{\hspace{0.05cm}} c @{\hspace{0.05cm}} c @{\hspace{0.05cm}} c}
            \includegraphics[width=\itemwidth]{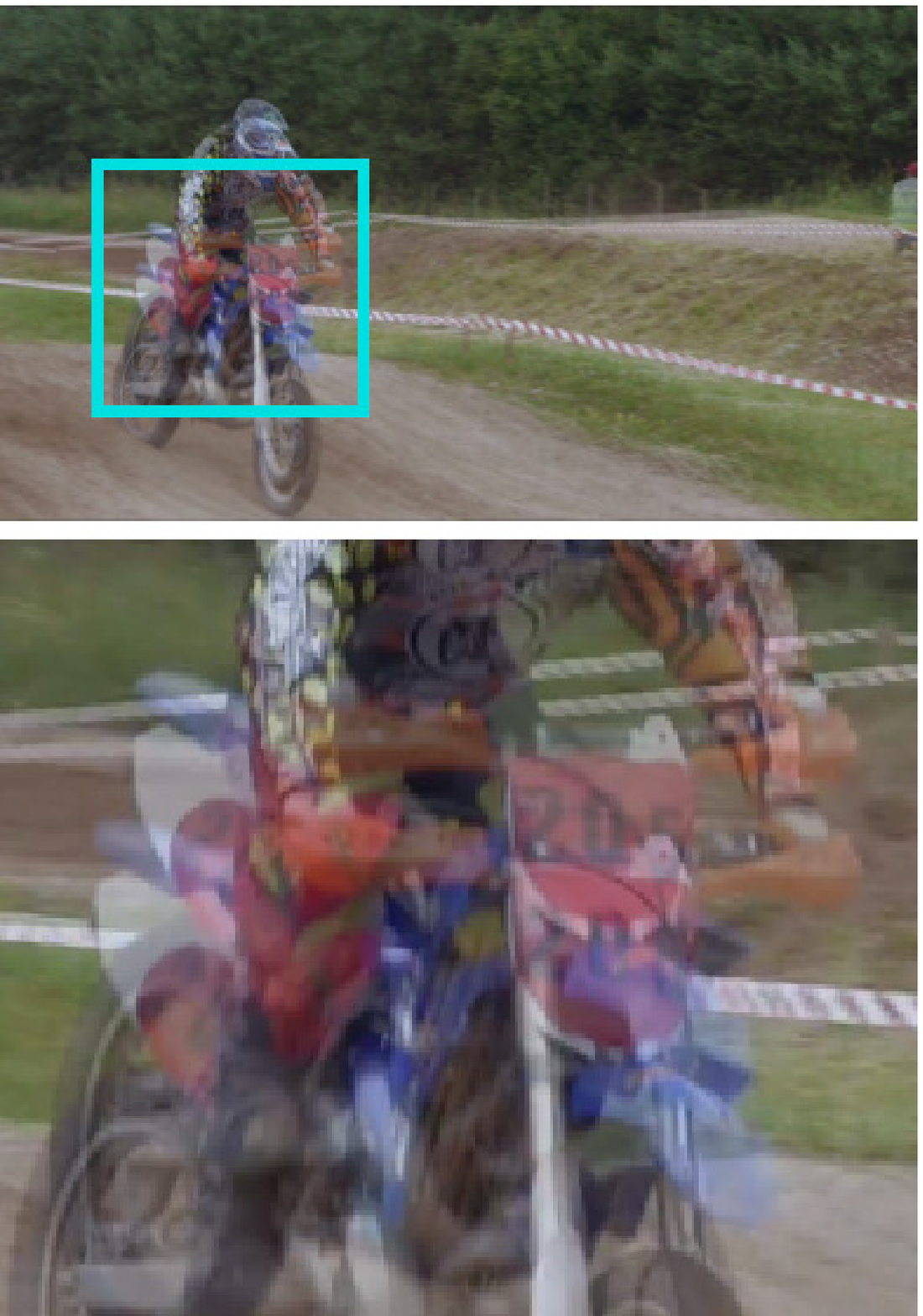}
        &
            \includegraphics[width=\itemwidth]{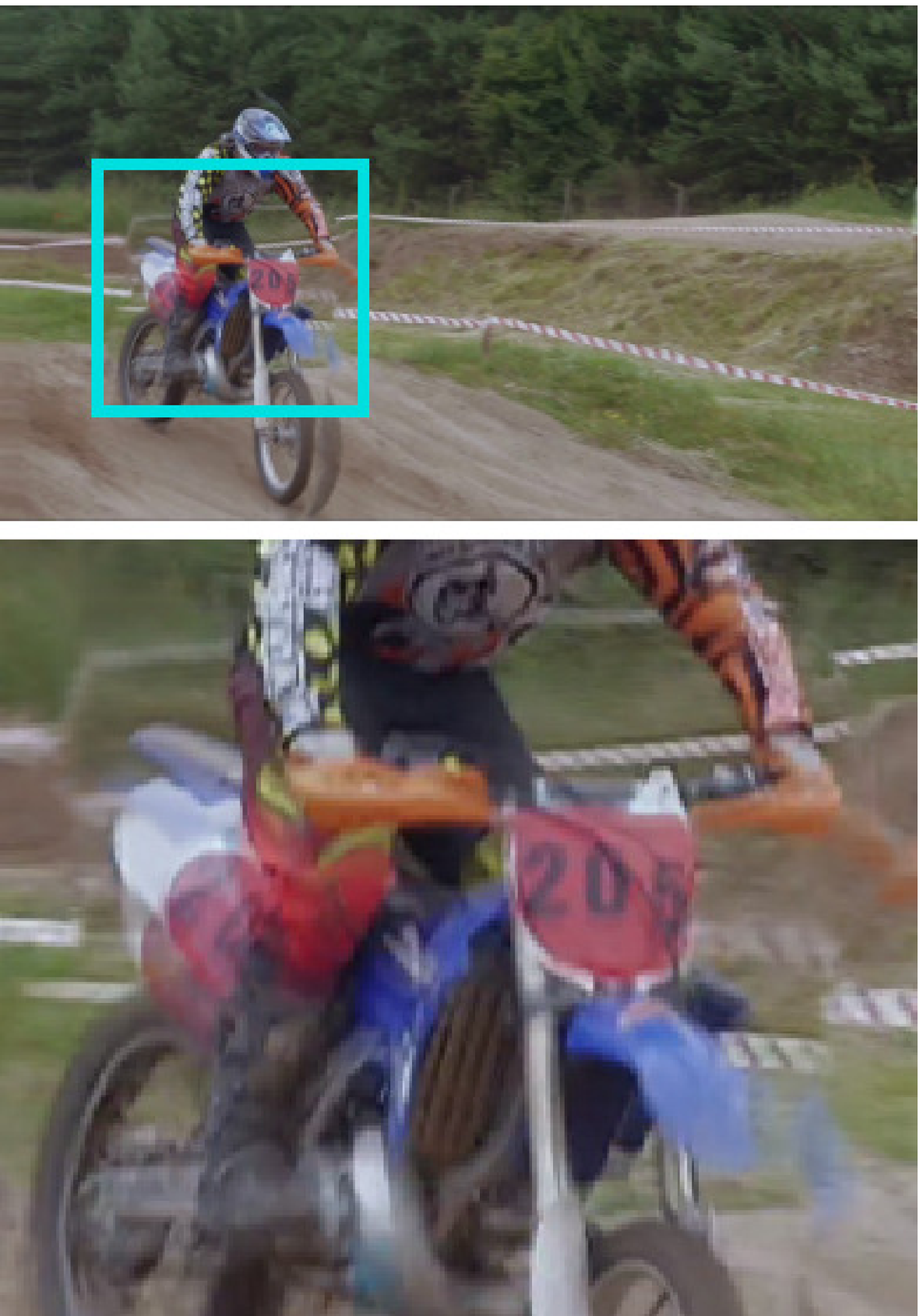}
        &
            \includegraphics[width=\itemwidth]{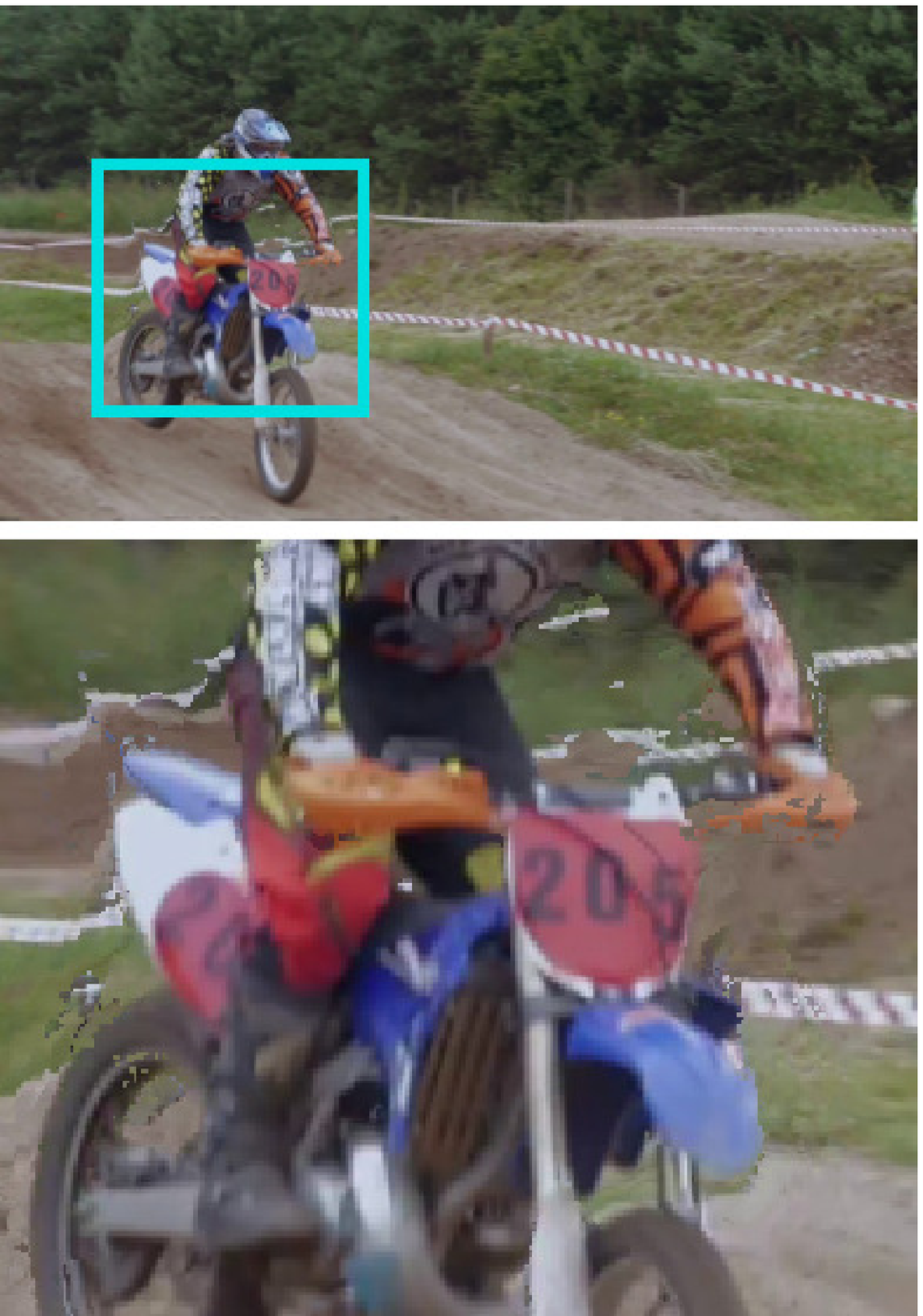}
        &
            \includegraphics[width=\itemwidth]{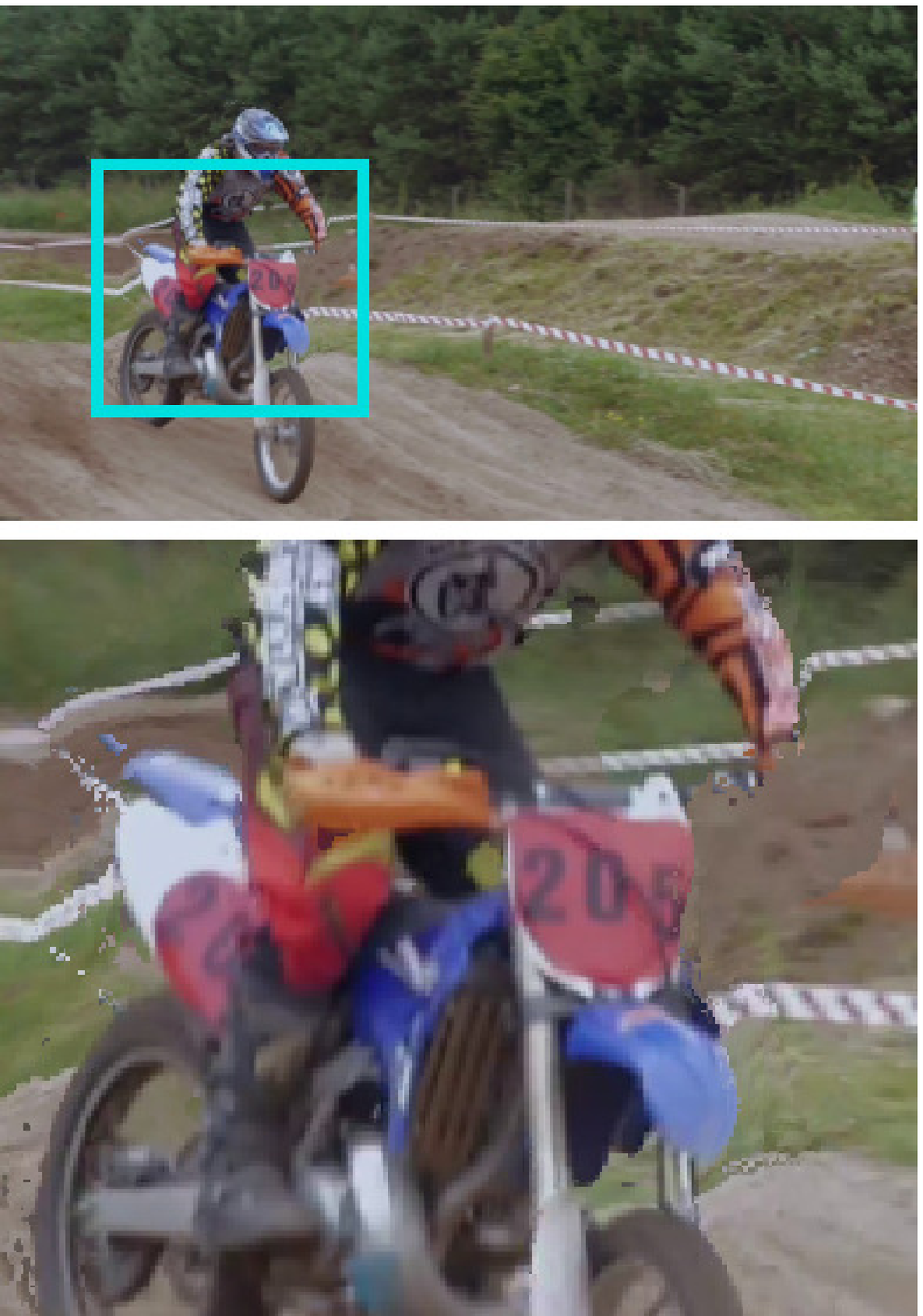}
        &
            \includegraphics[width=\itemwidth]{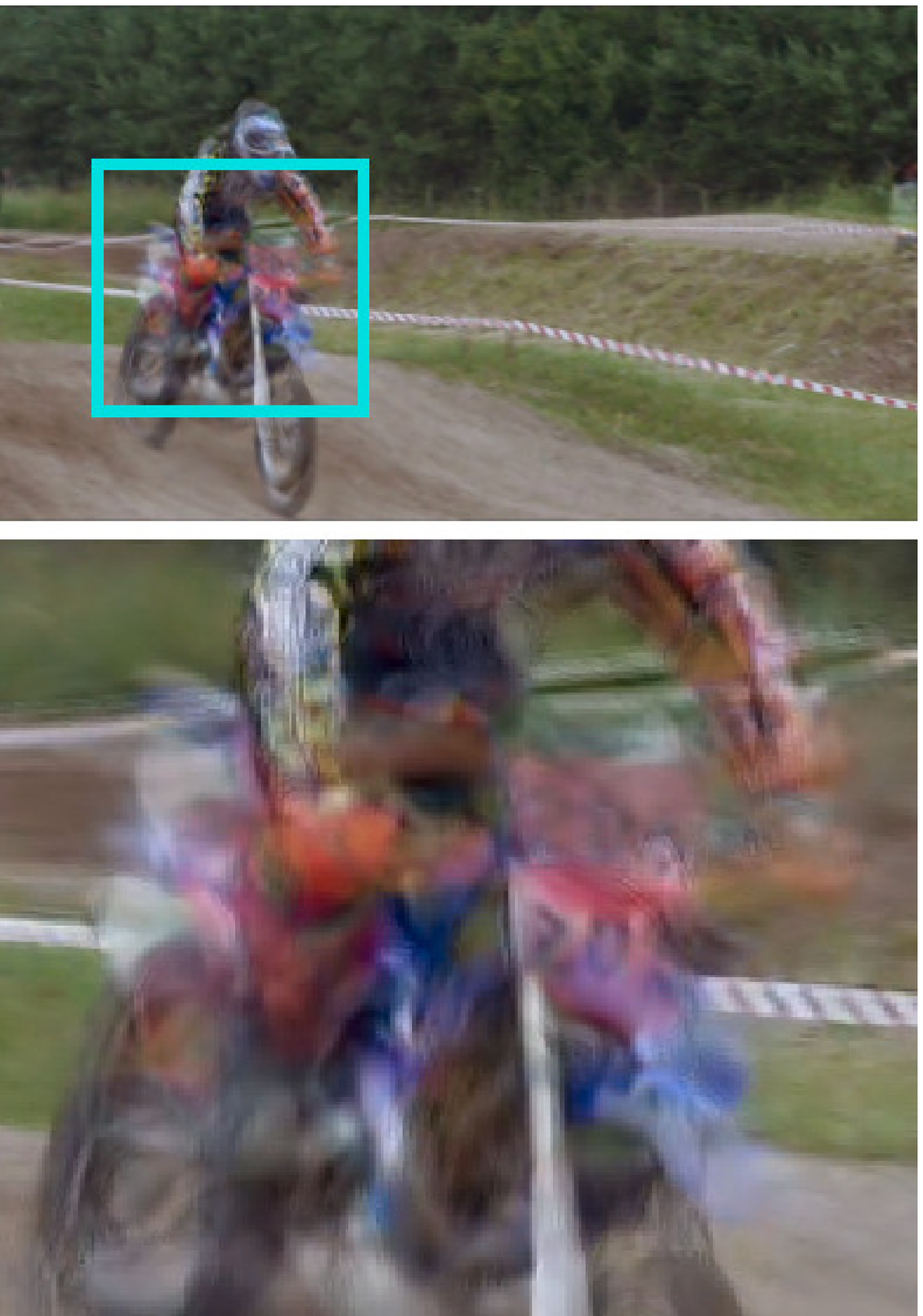}
        &
            \includegraphics[width=\itemwidth]{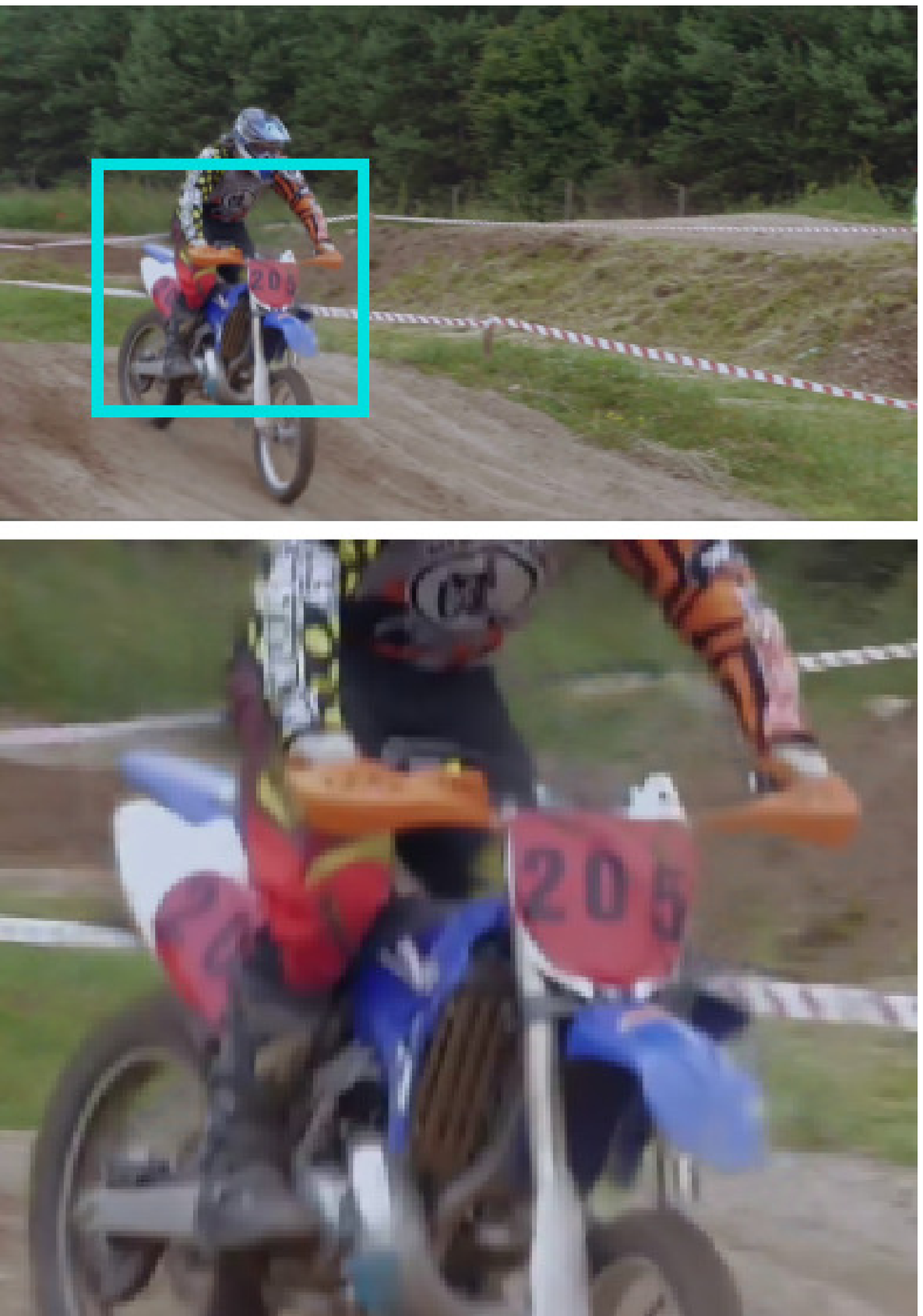}
        &
            \includegraphics[width=\itemwidth]{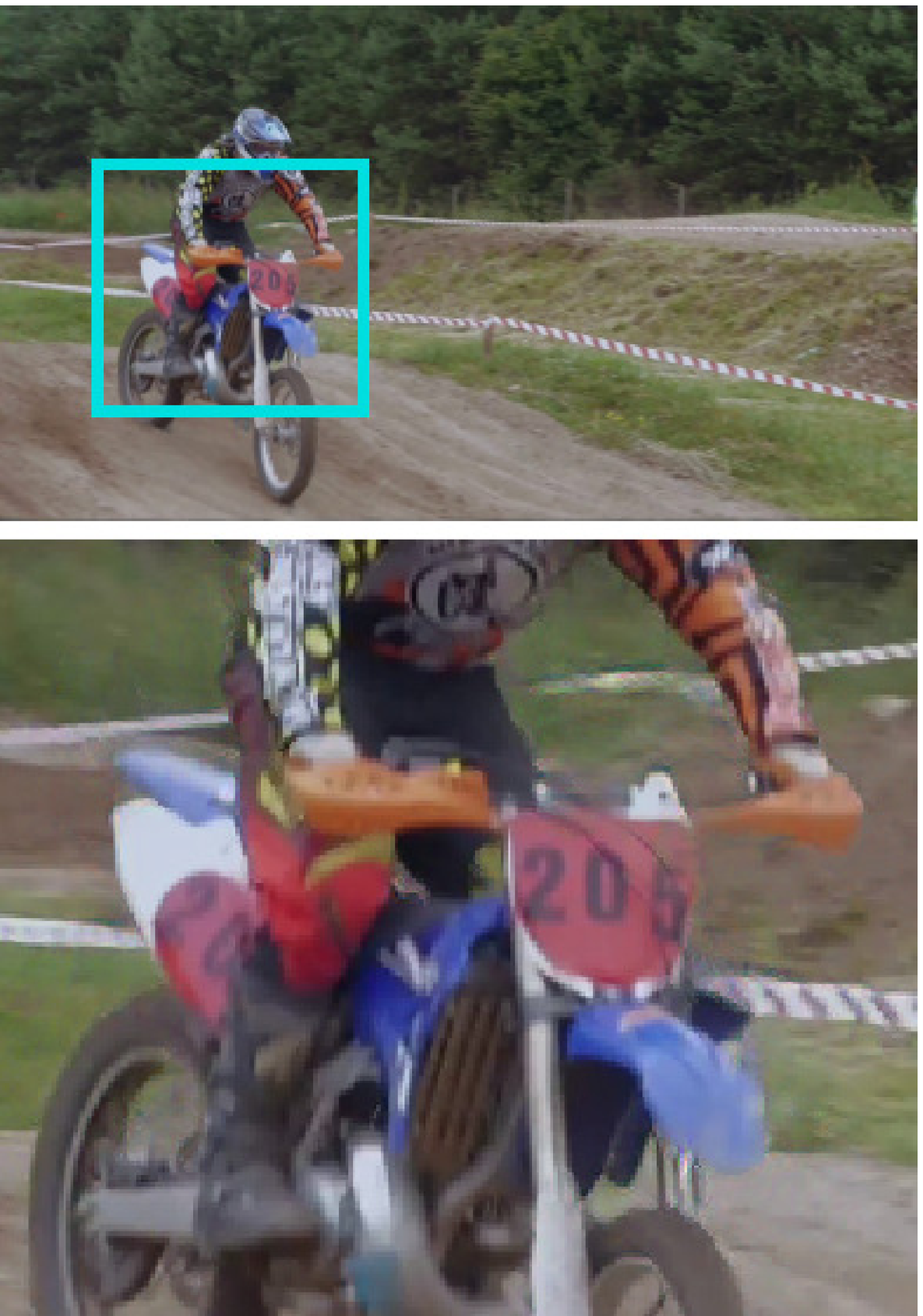}
        &
            \includegraphics[width=\itemwidth]{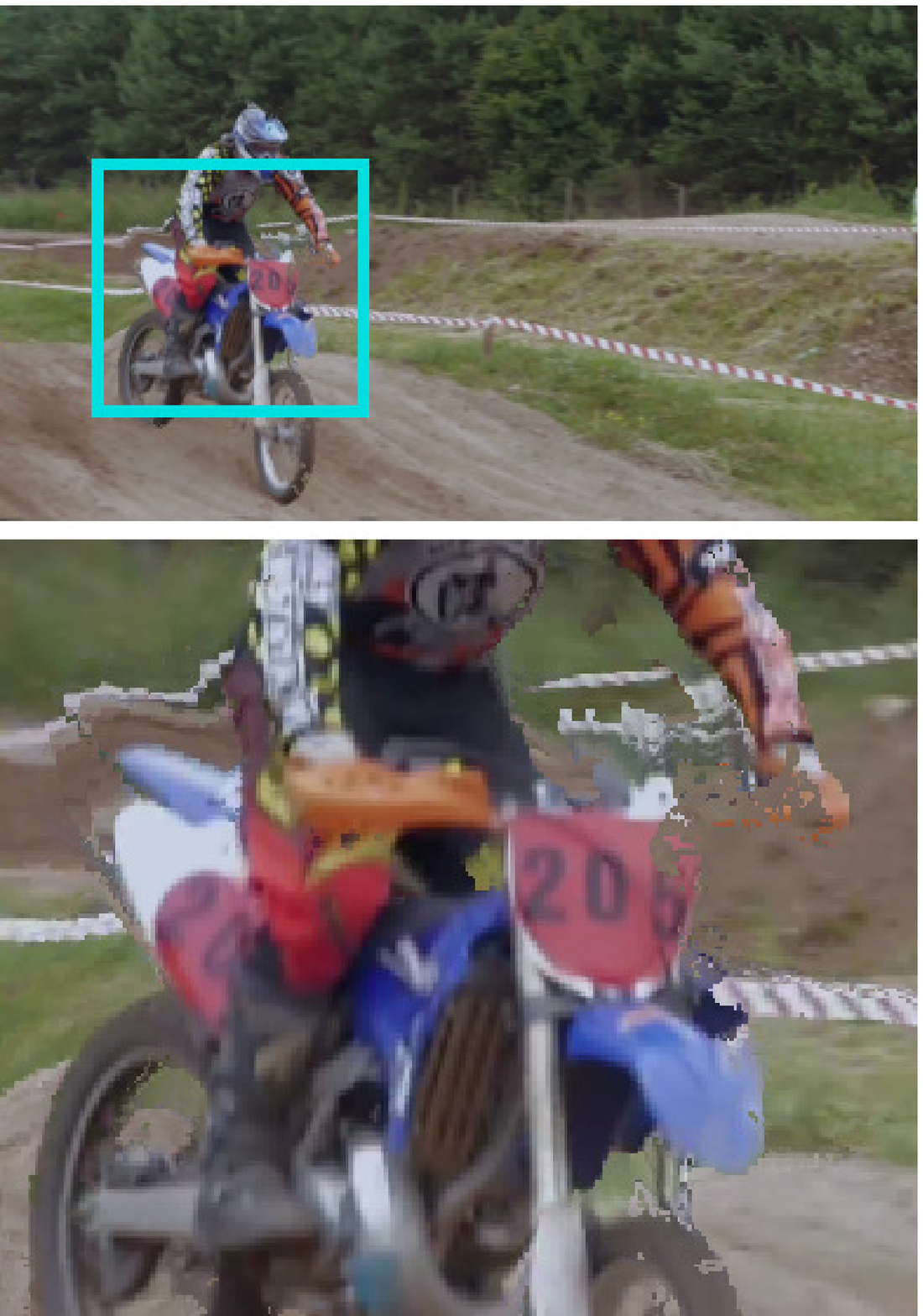}
        \vspace{-0.1cm} \\
            \footnotesize Overlayed input
        &
            \footnotesize SepConv - $\mathcal{L}_F$
        &
            \footnotesize MDP-Flow2
        &
            \footnotesize DeepFlow2
        &
            \footnotesize Meyer~\etal
        &
            \footnotesize Ours - $\mathcal{L}_{\textit{Lap}}$
        &
            \footnotesize Ours - $\mathcal{L}_F$
        &
            \footnotesize PWC-Net
        \\
    \end{tabularx}\vspace{-0.2cm}
    \caption{Visual comparison among frame interpolation methods. We used our own implementation of PWC-Net here.}\vspace{-0.3cm}
    \label{fig:examples}
\end{figure*}

\vspace{0.05in}
\noindent\textbf{Loss functions.} We consider three different loss functions to train our frame synthesis neural network, as detailed in Section~\ref{sec:network}. Whereas the $\ell_1$ color loss and the Laplacian loss aim to minimize the color difference, the feature loss focuses on perceptual difference. For simplicity, we refer to the model that has been trained with the color loss as ``$\mathcal{L}_1$" and the model trained with the Laplacian loss as ``$\mathcal{L}_{\textit{Lap}}$". Following Niklaus~\etal~\cite{Niklaus_ICCV_2017}, we do not directly train the model with the feature loss. Instead, we first train it with the Laplacian loss and then refine it with the feature loss. We refer to this model as ``$\mathcal{L}_F$".

As reported in Table~\ref{tbl:public}, the Laplacian loss $\mathcal{L}_{\textit{Lap}}$ and the color loss $\mathcal{L}_1$, especially the former, outperform the feature loss $\mathcal{L}_F$ as well as the state-of-the-art methods quantitatively. As expected, the color loss functions yield better results since they directly optimize the evaluation metric. On the other hand, we find that the feature loss tends to produce visually more pleasant results, as shown in our user study in Section~\ref{subsec:visual} and Figure~\ref{fig:loss}. This finding is consistent with what was reported in recent work~\cite{Niklaus_ICCV_2017}.

\vspace{0.05in}
\noindent\textbf{Frame synthesis network vs pixel-wise blending.} To examine the effectiveness of our frame synthesis neural network, we compare it to a blending baseline that employs the off-the-shelf warping algorithm from the Middlebury benchmark~\cite{Baker_OTHER_2011} to warp the input frames and then blends them together. The same bidirectional optical flow used in our synthesis network approach is used in this baseline. As a reference, we also compare to another baseline that also employs the algorithm from the Middlebury benchmark but only uses the forward flow for warping. As shown in Table~\ref{tbl:bidirectional}, our synthesis network approach shows a clear advantage over the blending approach. This can be attributed to the capability of our synthesis network in tolerating inaccuracies in optical flow estimation. Note that the demonstrated quantitative advantage also translates to improved visuals as shown in Figure \ref{fig:ablation}.

\vspace{0.05in}
\noindent\textbf{Contextual information.} To understand the effect of the contextual information, we trained a synthesis network that only receives two pre-warped frames as input. As shown in Table~\ref{tbl:context}, the contextual information significantly helps our frame synthesis network to produce high-quality results.

Our method uses the \verb|conv1| layer of ResNet-18~\cite{He_CVPR_2016} to extract per-pixel contextual information. We also tested other options, such as using layers \verb|conv1_1| and \verb|conv1_2| of VGG-19~\cite{Simonyan_CORR_2014}. As reported in Table~\ref{tbl:context}, while the \verb|conv1| layer of ResNet-18 overall works better, the difference between various ways to extract contextual information is minuscule. They all significantly outperform the baseline network without contextual information.

\vspace{0.05in}
\noindent\textbf{Optical flow.} To evaluate the effect of the utilized optical flow algorithm on our method, we use SPyNet~\cite{Ranjan_CVPR_2017} as an alternative to our implementation of PWC-Net~\cite{Sun_CORR_2017}. In optical flow benchmarks, SPyNet performs less well than PWC-Net and as shown in Table~\ref{tbl:flow}, this decreased accuracy also affects the synthesis results. Furthermore, training a synthesis network to directly operate on the input frames, without any pre-warping, significantly worsens the synthesis quality. This shows that it is helpful to use motion compensation to provide a good initialization for frame interpolation. In fact, even using H.264 motion vectors is already beneficial. This is remarkable, considering that these motion vectors are only available per block.

\subsection{Quantitative evaluation}

We perform a quantitative comparison with state of the art frame interpolation algorithms on the interpolation section of the Middlebury benchmark for optical flow~\cite{Baker_OTHER_2011}. As reported in Table~\ref{tbl:hidden}, our method establishes a new state-of-the-art and improves the previously best performing method by a notable margin. Furthermore, according to the feedback from the Middlebury benchmark organizer, our interpolation results are ranked 1\textsuperscript{st} among all the over 100 algorithms listed on the benchmark website.

\begin{figure*}\centering
    \vspace*{0.2cm}\hspace*{-0.2cm}\includegraphics[]{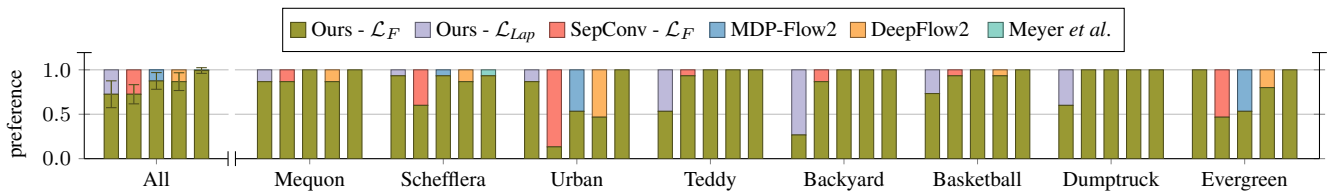}\vspace{-0.3cm}
	\caption{Results from the user study. The error bars denote the standard deviation.}\vspace{-0.35cm}
	\label{fig:study}
\end{figure*}

\begin{figure}\centering
    \setlength{\tabcolsep}{0.0cm}
    \setlength{\itemwidth}{4.15cm}

    \begin{tabularx}{\textwidth}{c @{\hspace{0.05cm}} c}
            \includegraphics[width=\itemwidth]{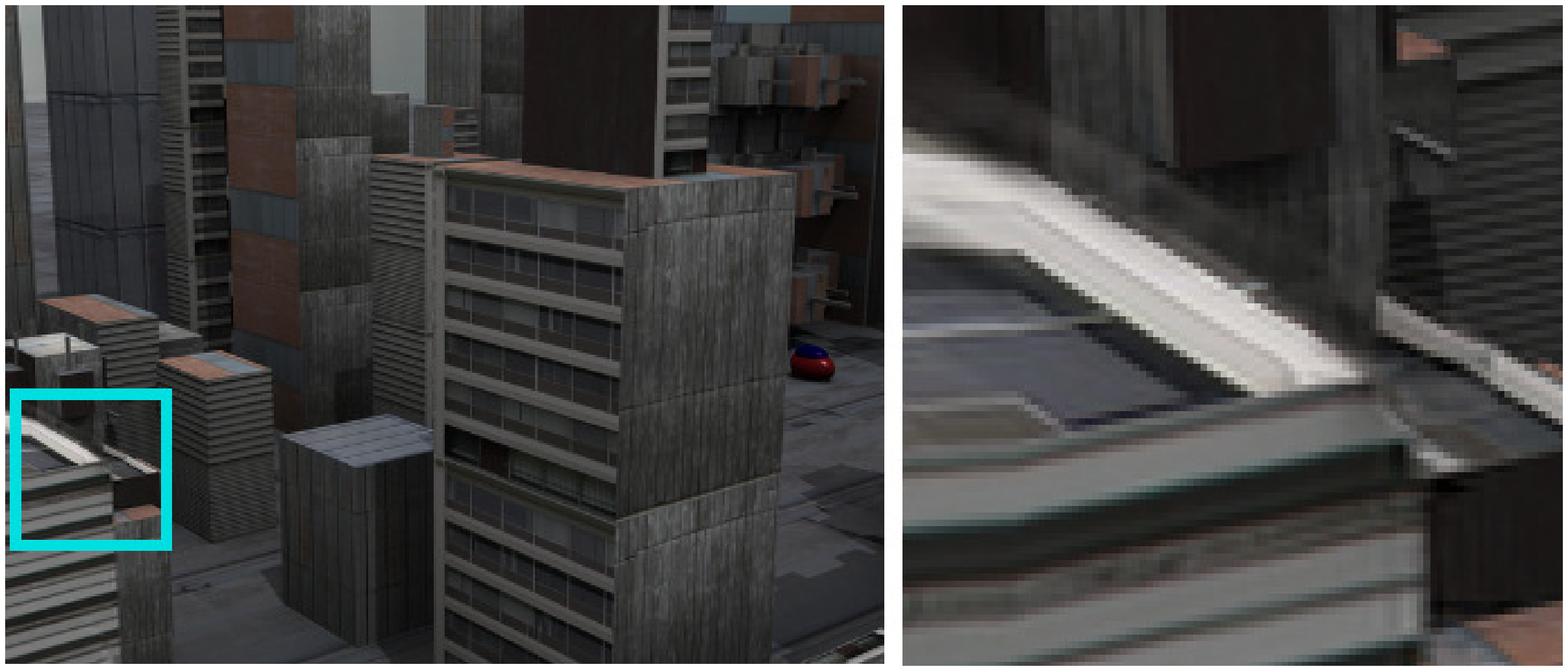}
        &
            \includegraphics[width=\itemwidth]{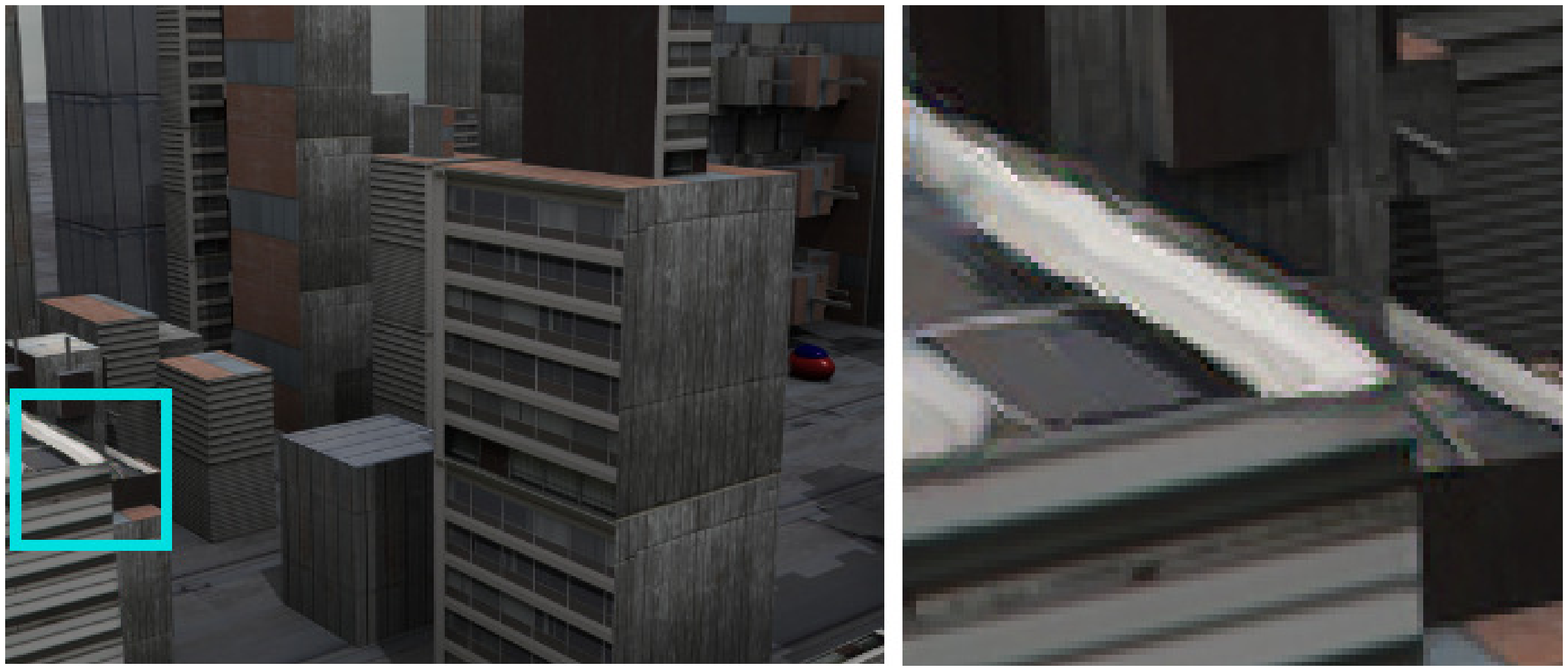}
        \vspace{-0.1cm} \\
            \footnotesize SepConv - $\mathcal{L}_F$
        &
            \footnotesize Ours - $\mathcal{L}_F$
        \vspace{0.1cm} \\
            \includegraphics[width=\itemwidth]{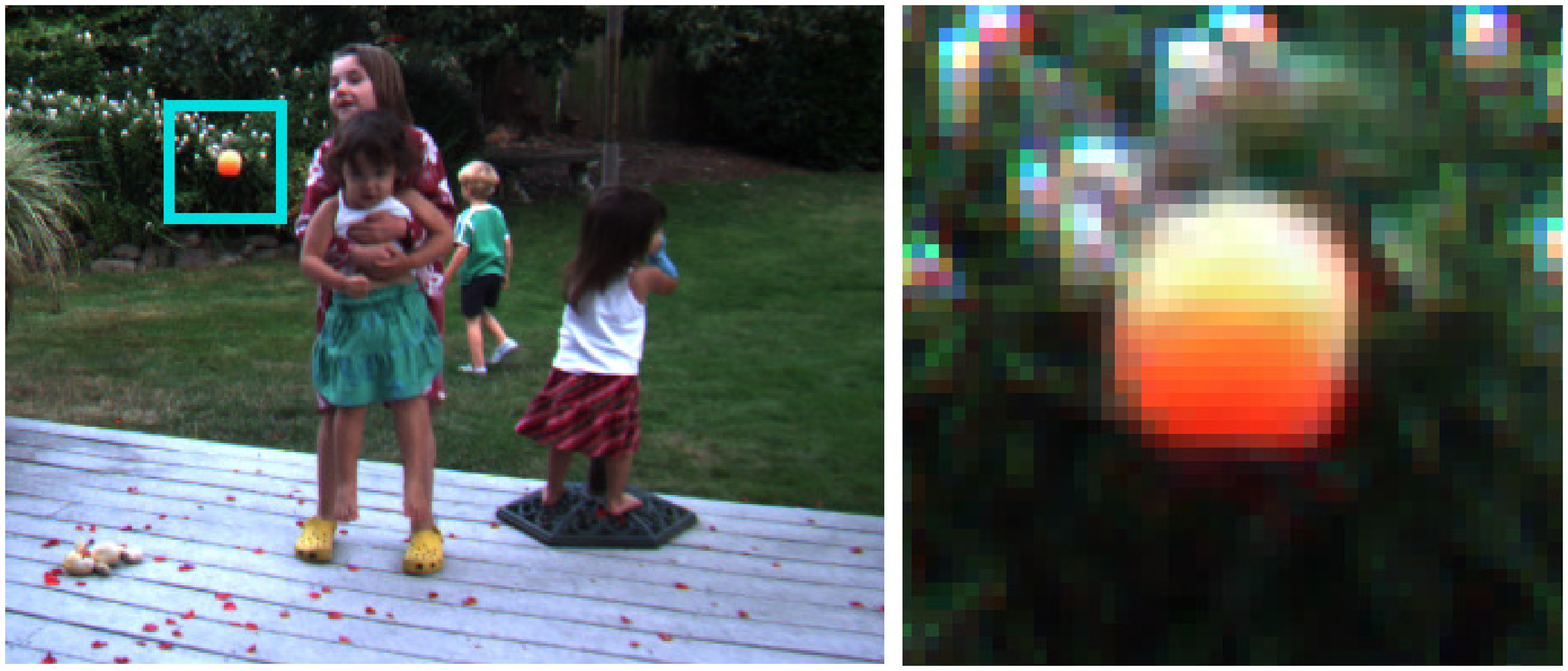}
        &
            \includegraphics[width=\itemwidth]{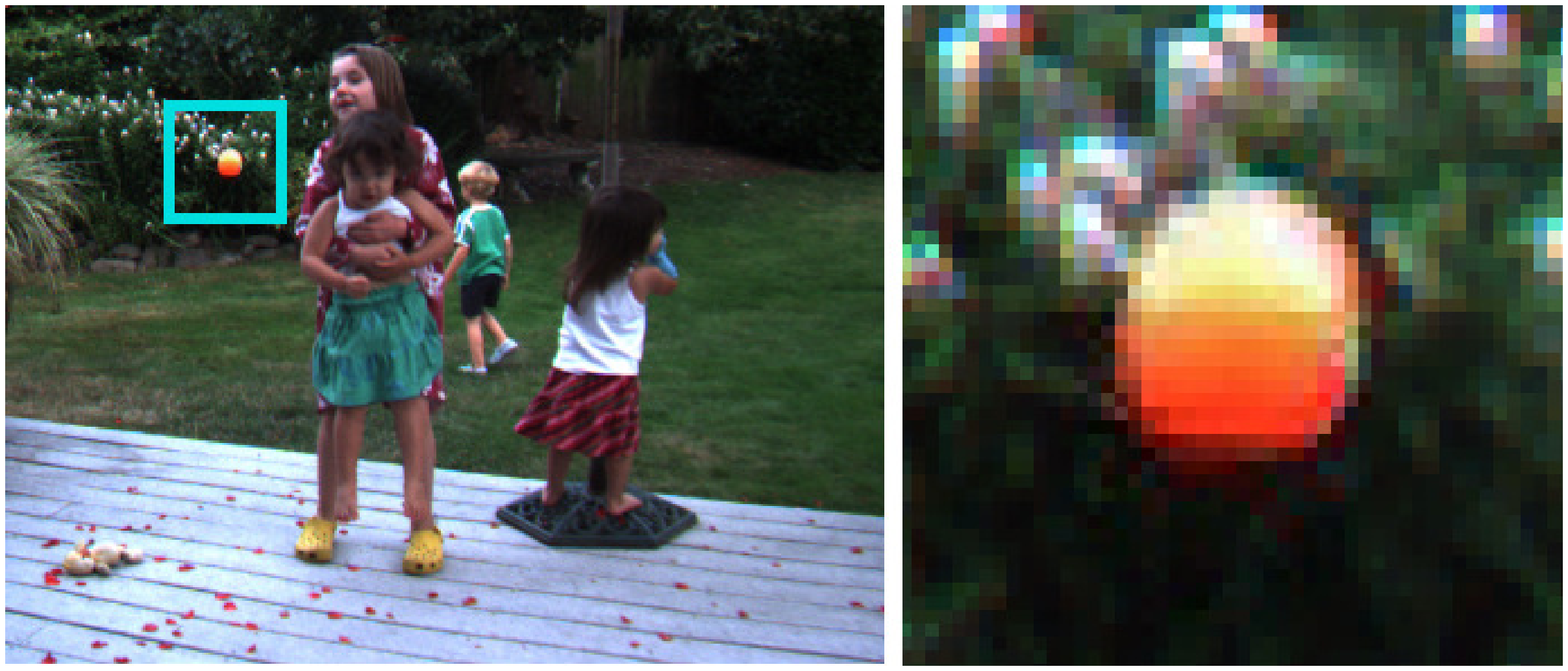}
        \vspace{-0.1cm} \\
            \footnotesize Ours - $\mathcal{L}_{\textit{Lap}}$
        &
            \footnotesize Ours - $\mathcal{L}_F$
        \\
    \end{tabularx}\vspace{-0.2cm}
    \caption{Examples for which our $\mathcal{L}_F$ results were not preferred in the user study.}\vspace{-0.3cm}
    \label{fig:failure}
\end{figure}

\subsection{Visual comparison}\label{subsec:visual}

We show a comparison of our proposed approach with state-of-the-art video frame interpolation methods in Figure~\ref{fig:examples}. These examples are subject to significant motion and occlusion. In general, our approach is capable of handling these challenging scenarios, with the $\mathcal{L}_F$ loss trained synthesis network retaining more high-frequency details. In comparison, SepConv-$\mathcal{L}_{F}$ fails to compensate for the large motion and is limited by the size of its adaptive kernels. The optical flow based methods MDP-Flow2, DeepFlow2, and PWC-Net handle the large motion better than SepConv-$\mathcal{L}_{F}$ but introduce artifacts due to their limited synthesis capabilities. Lastly, the approach from Meyer~\etal fails to handle large motion due to phase ambiguities. Please refer to the supplemental video to see these examples in motion.

\begin{figure}\centering
    \setlength{\tabcolsep}{0.0cm}
    \setlength{\itemwidth}{2.75cm}

    \begin{tabularx}{\textwidth}{c @{\hspace{0.05cm}} c @{\hspace{0.05cm}} c}
            \includegraphics[width=\itemwidth]{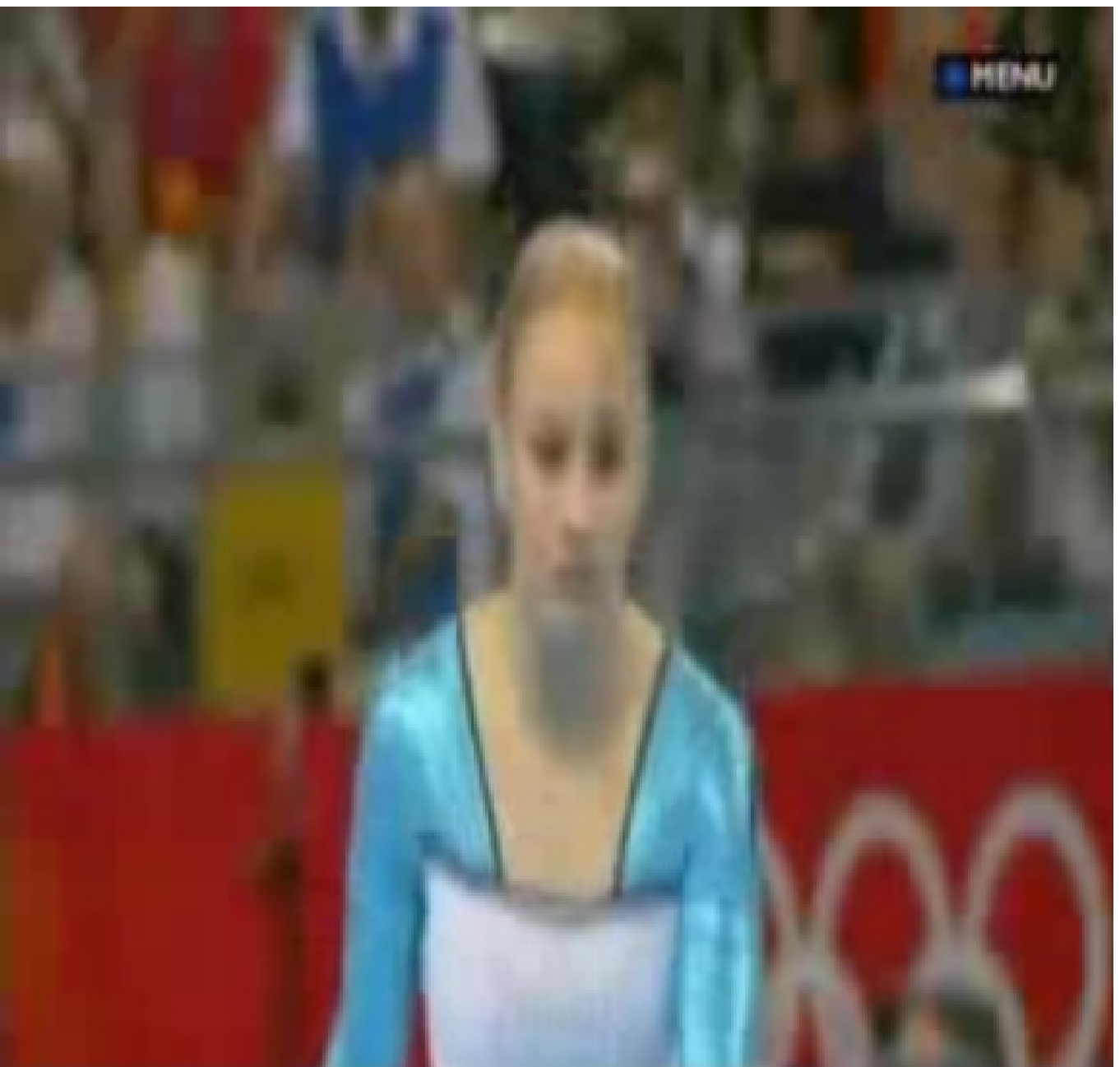}
        &
            \includegraphics[width=\itemwidth]{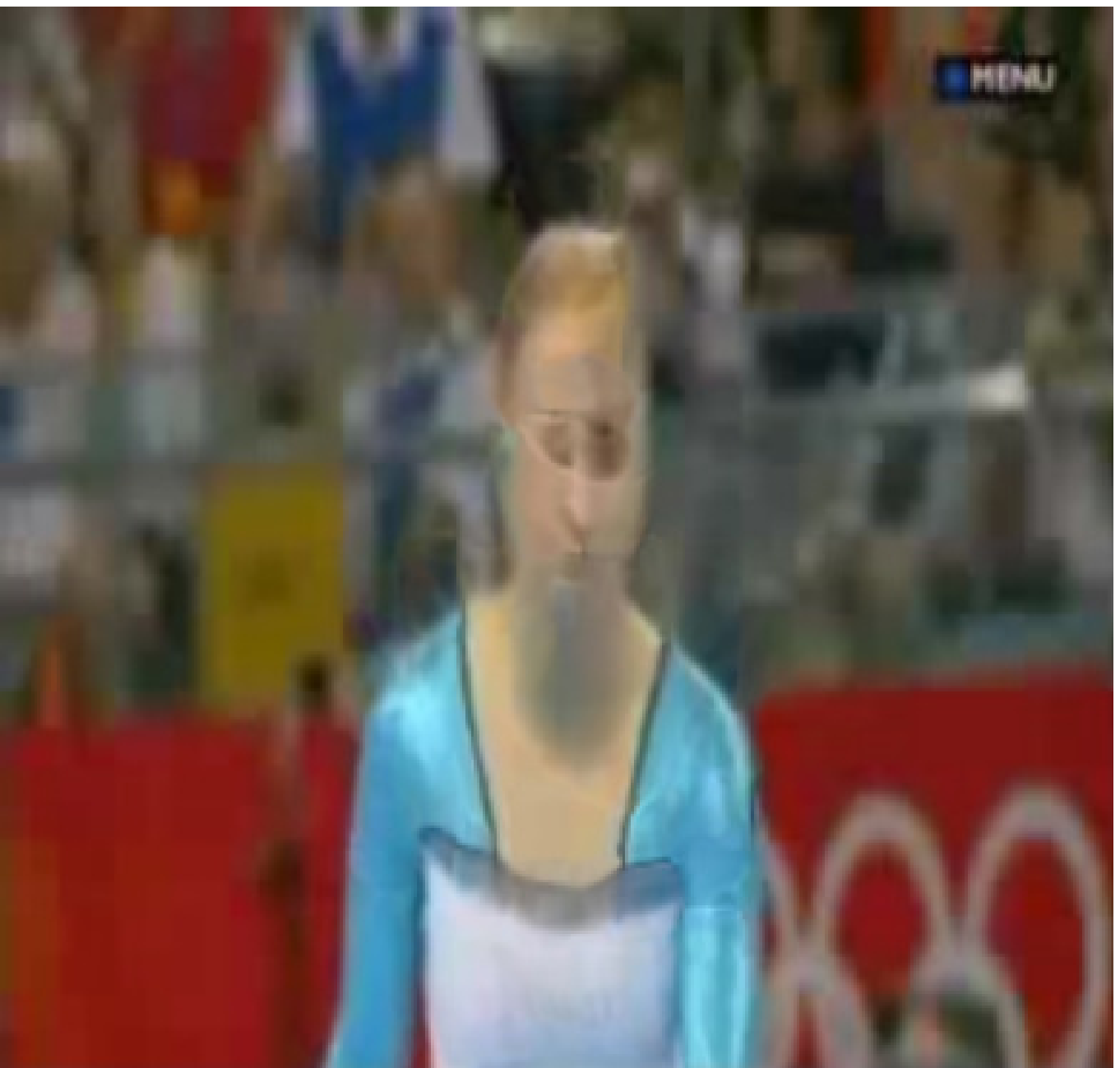}
        &
            \includegraphics[width=\itemwidth]{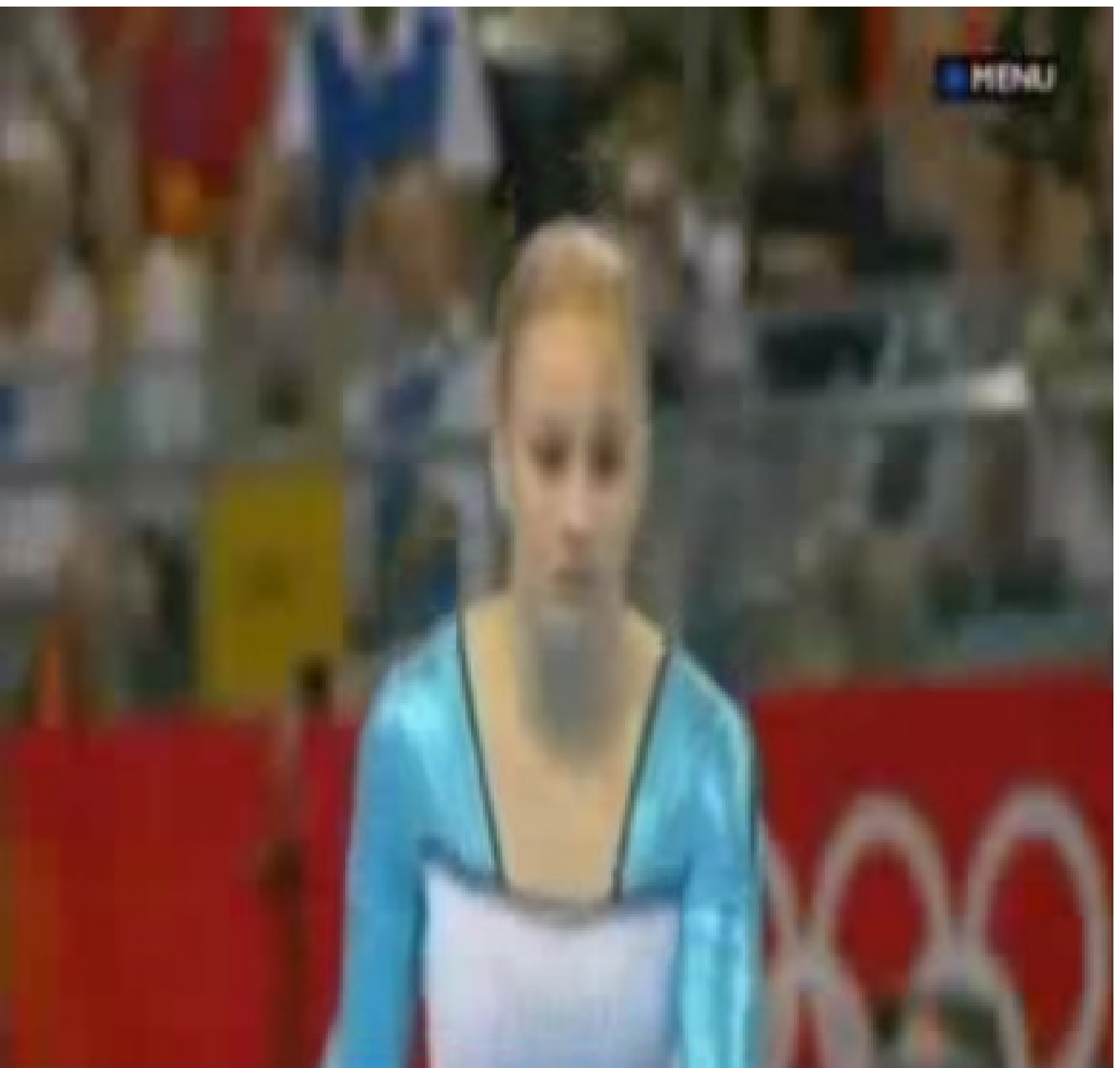}
        \vspace{-0.1cm} \\
    \end{tabularx}
    \begin{tabularx}{\textwidth}{c @{\hspace{0.05cm}} c @{\hspace{0.05cm}} c}
            \includegraphics[width=\itemwidth]{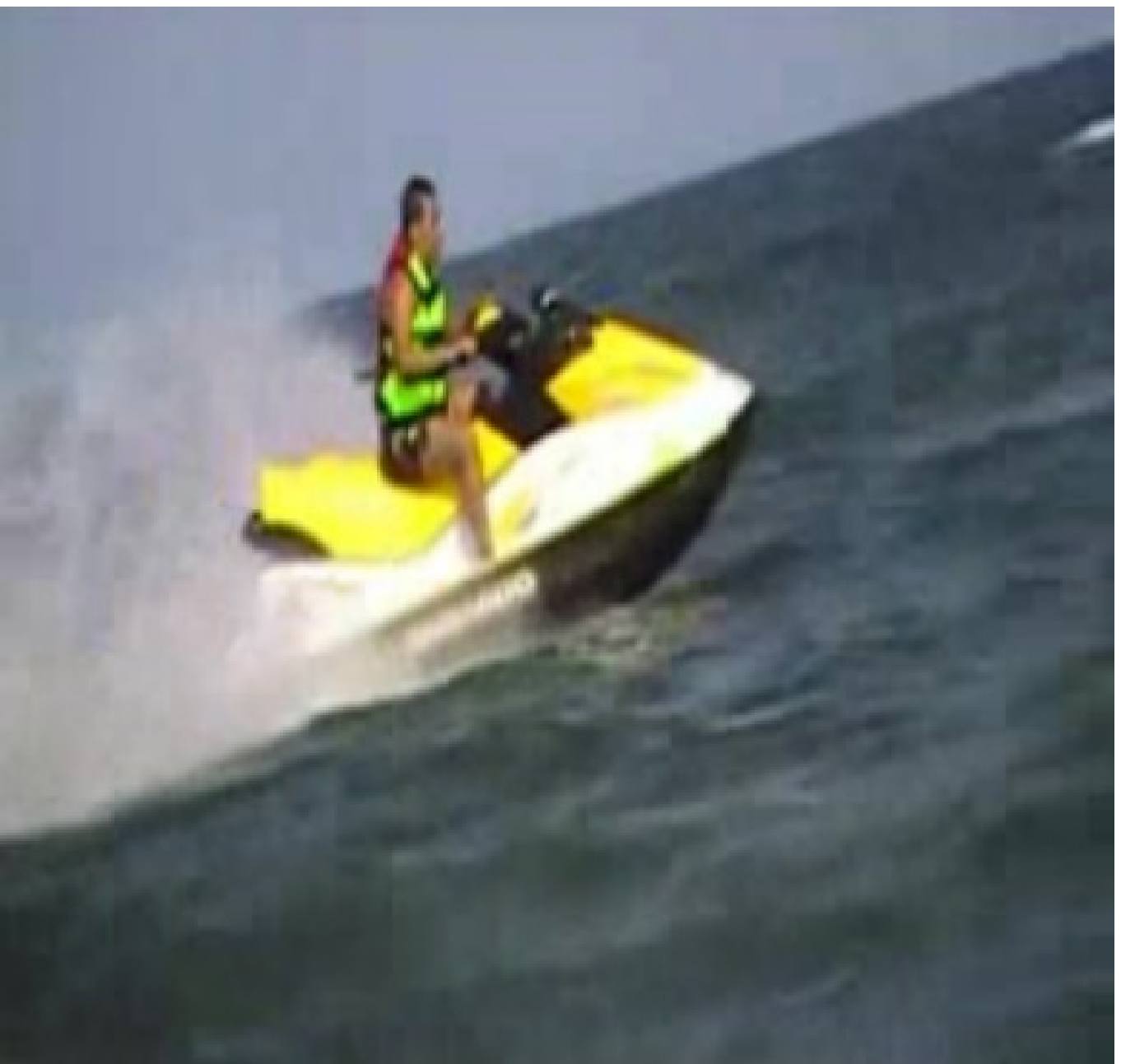}
        &
            \includegraphics[width=\itemwidth]{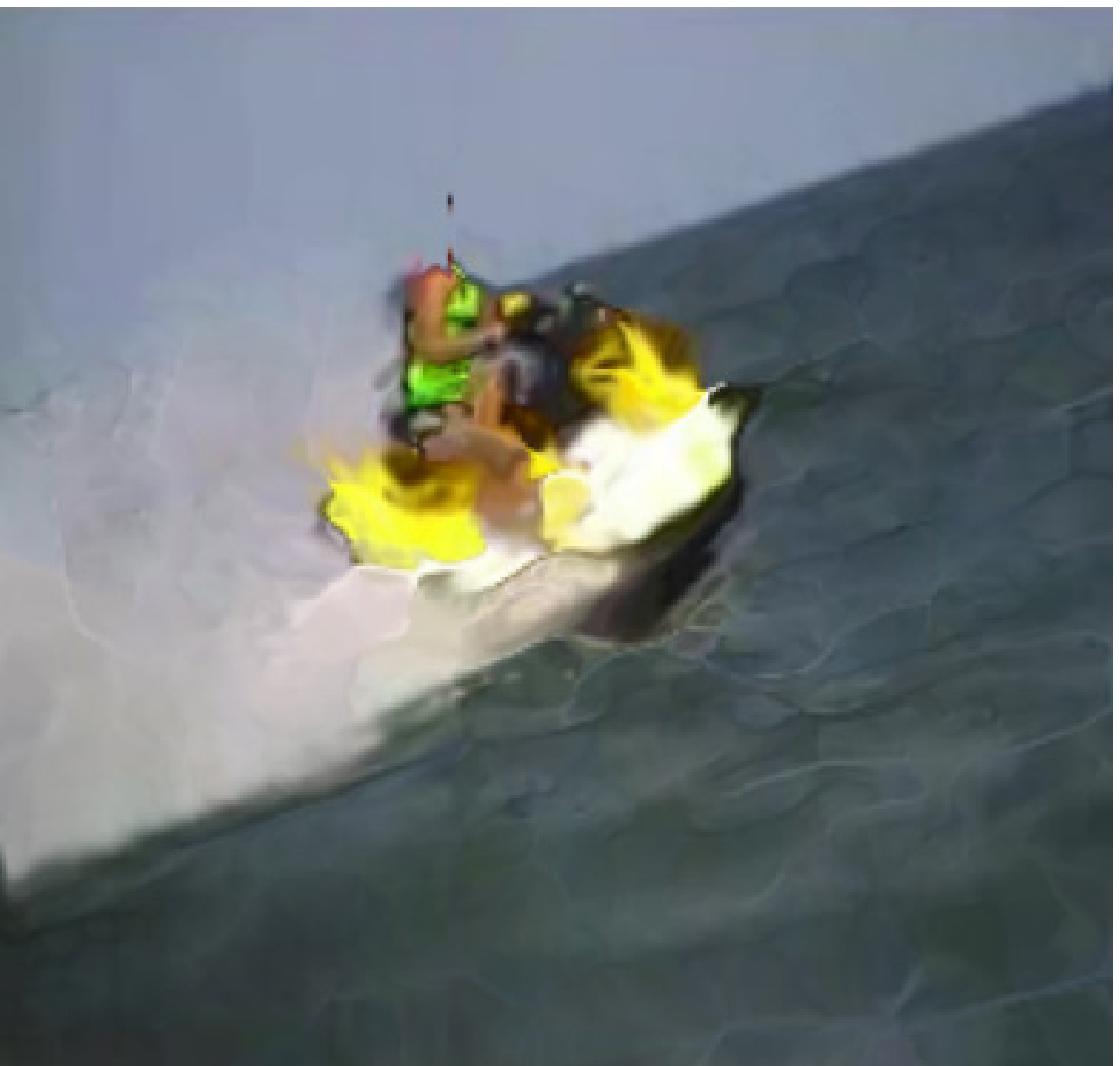}
        &
            \includegraphics[width=\itemwidth]{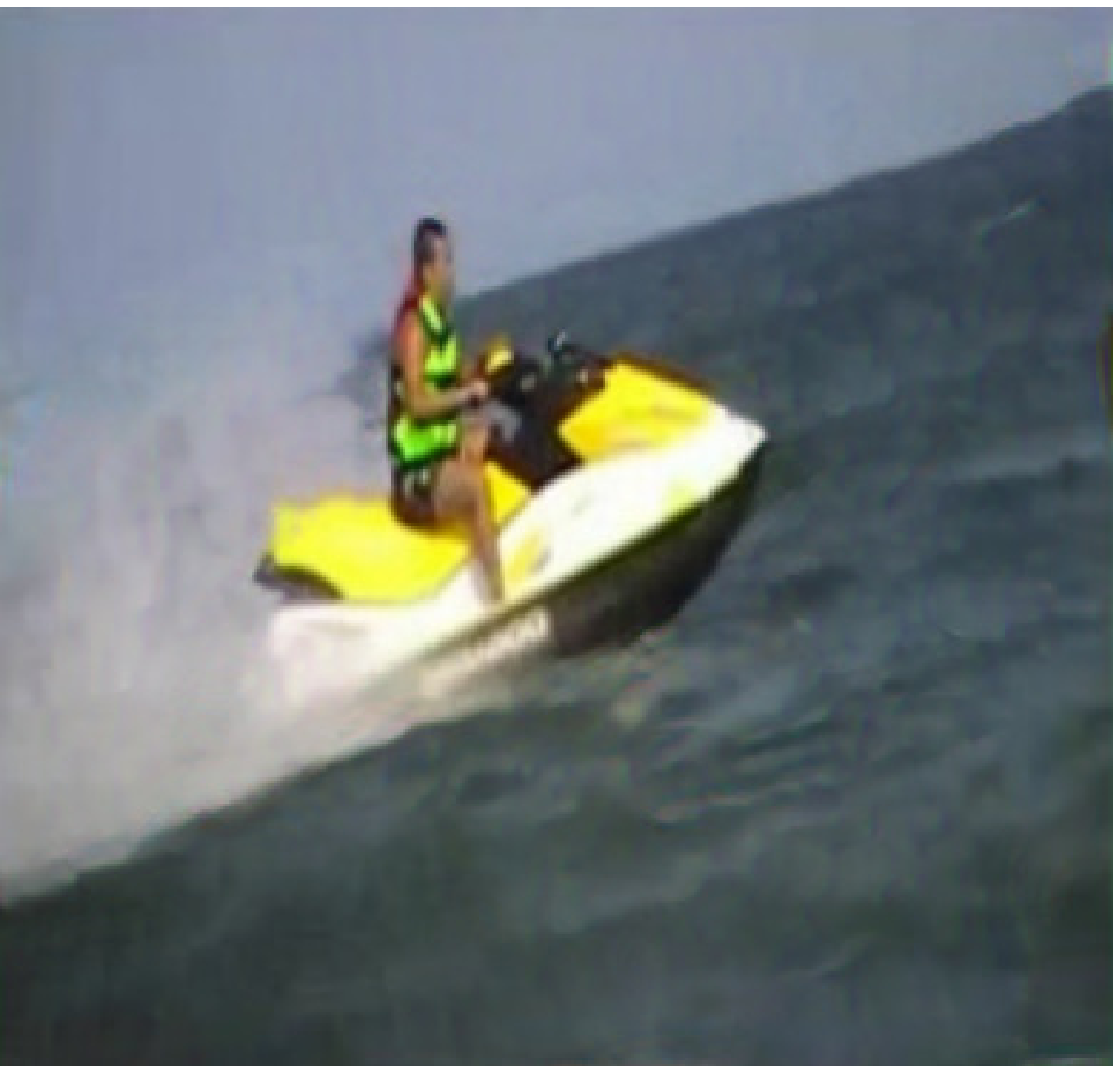}
        \vspace{-0.1cm} \\
            \footnotesize Ground truth
        &
            \footnotesize Voxel Flow
        &
            \footnotesize Ours - $\mathcal{L}_F$
        \\
    \end{tabularx}\vspace{-0.2cm}
    \caption{Comparison with the recent voxel flow method.}\vspace{-0.5cm}
    \label{fig:voxel}
\end{figure}

\begin{figure*}\centering
    \setlength{\tabcolsep}{0.0cm}
    \setlength{\itemwidth}{2.787cm}

    \begin{tabularx}{\textwidth}{c @{\hspace{0.15cm}} c @{\hspace{0.15cm}} c @{\hspace{0.15cm}} c @{\hspace{0.15cm}} c @{\hspace{0.15cm}} c @{\hspace{0.15cm}} c}
            \includegraphics[width=\itemwidth,trim={2.7cm 1.0cm 7.8cm 1.3cm},clip]{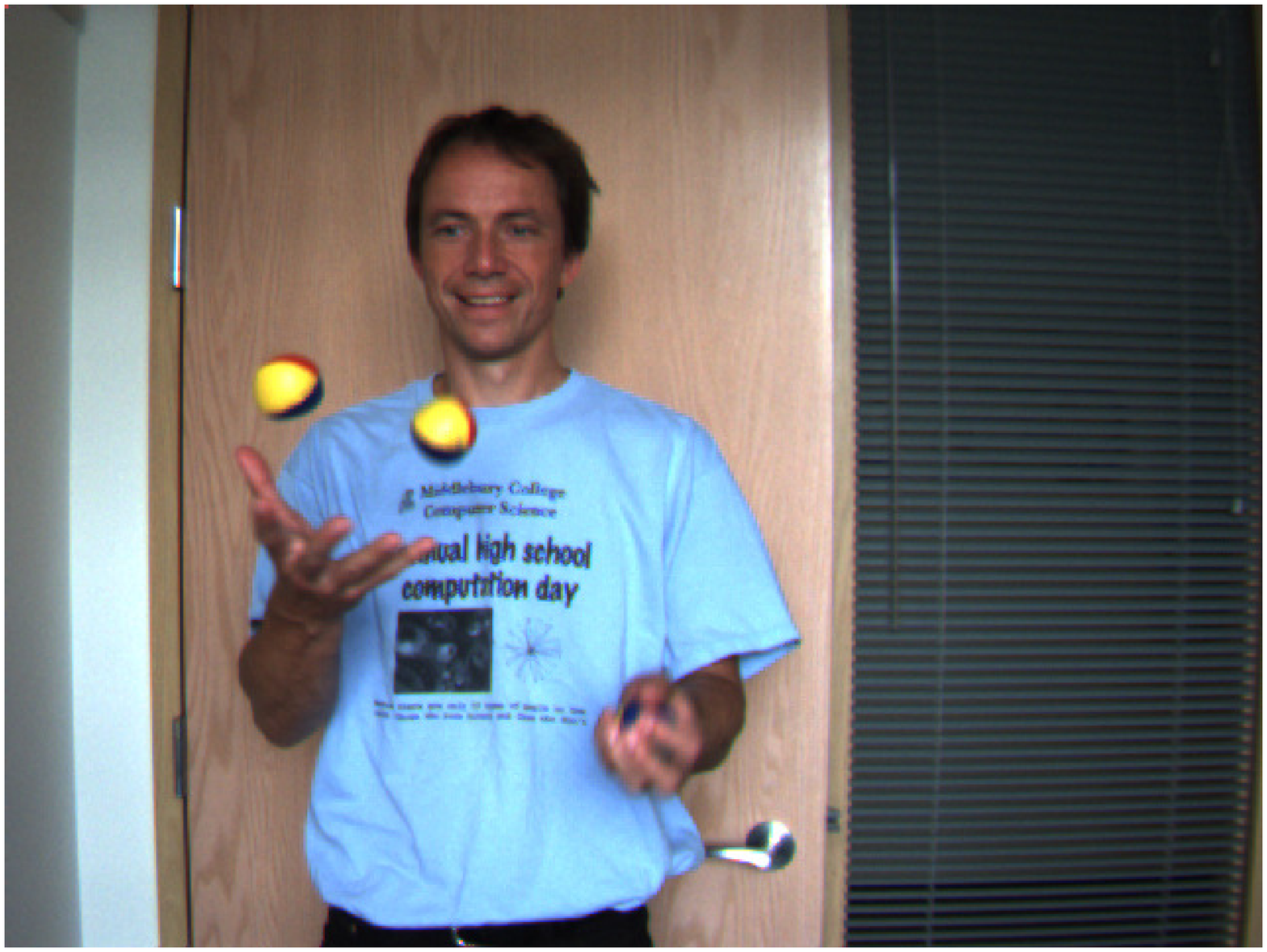}
        &
            \includegraphics[width=\itemwidth,trim={2.7cm 1.0cm 7.8cm 1.3cm},clip]{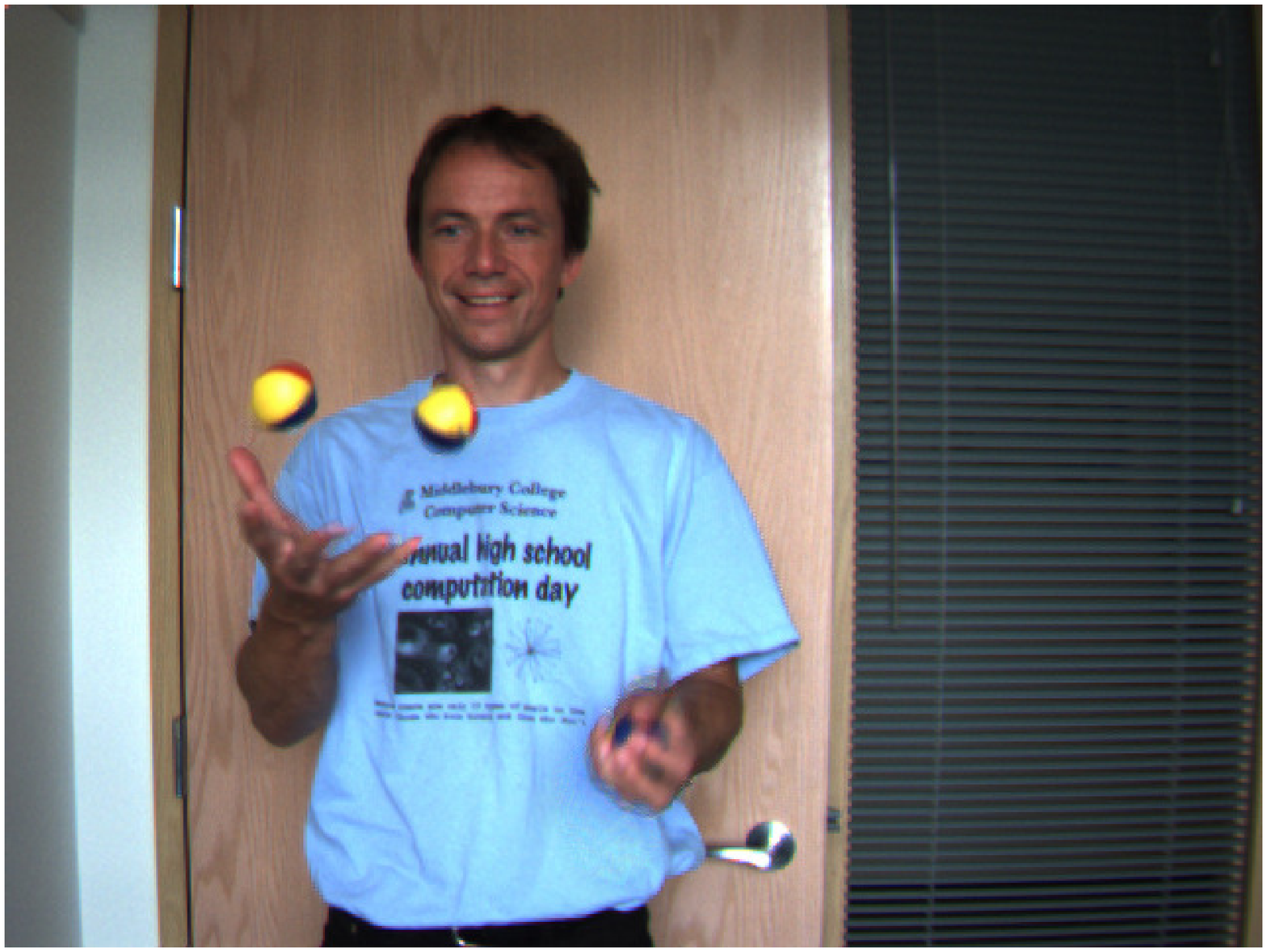}
        &
            \includegraphics[width=\itemwidth,trim={2.7cm 1.0cm 7.8cm 1.3cm},clip]{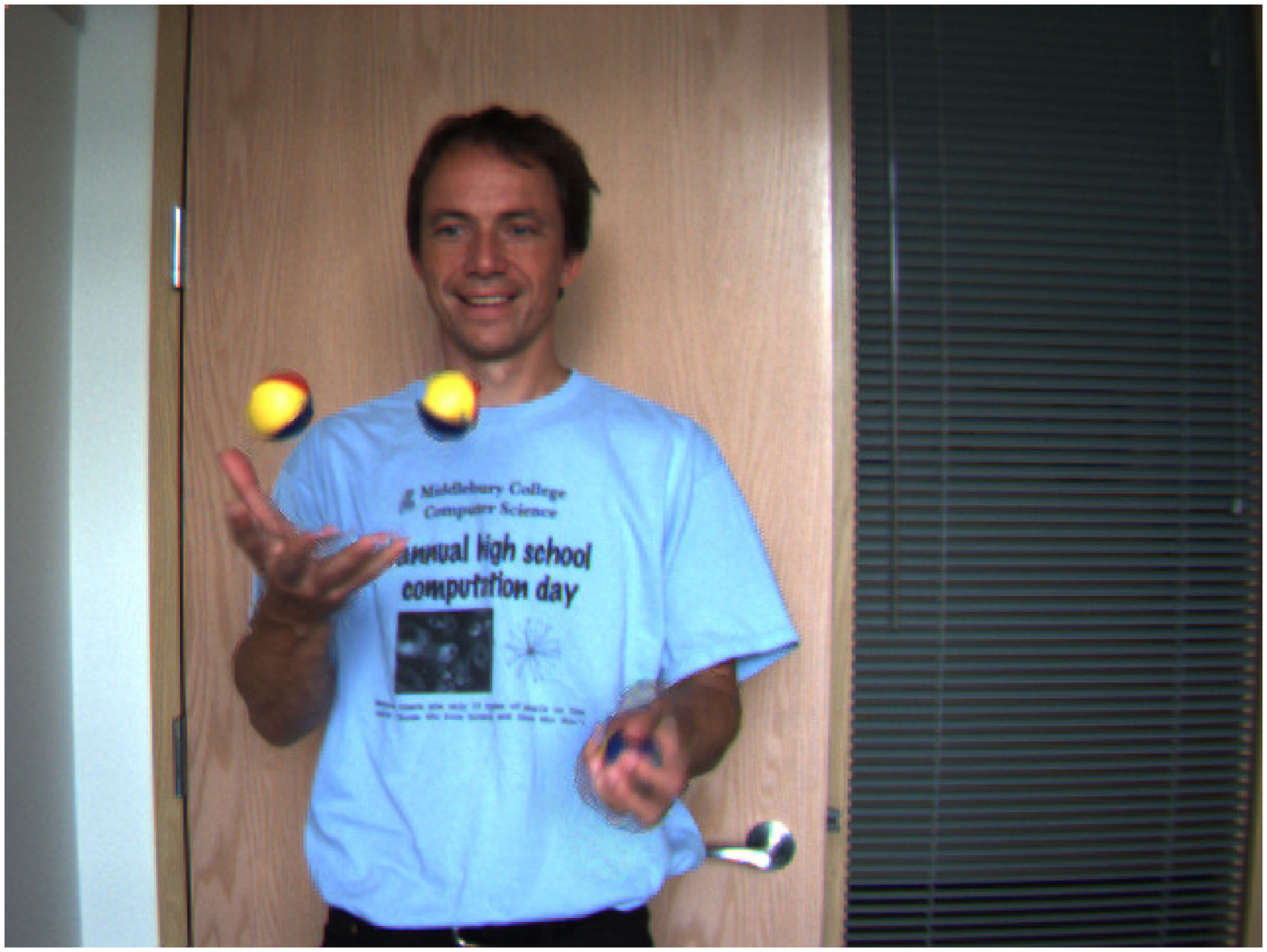}
        &
            \includegraphics[width=\itemwidth,trim={2.7cm 1.0cm 7.8cm 1.3cm},clip]{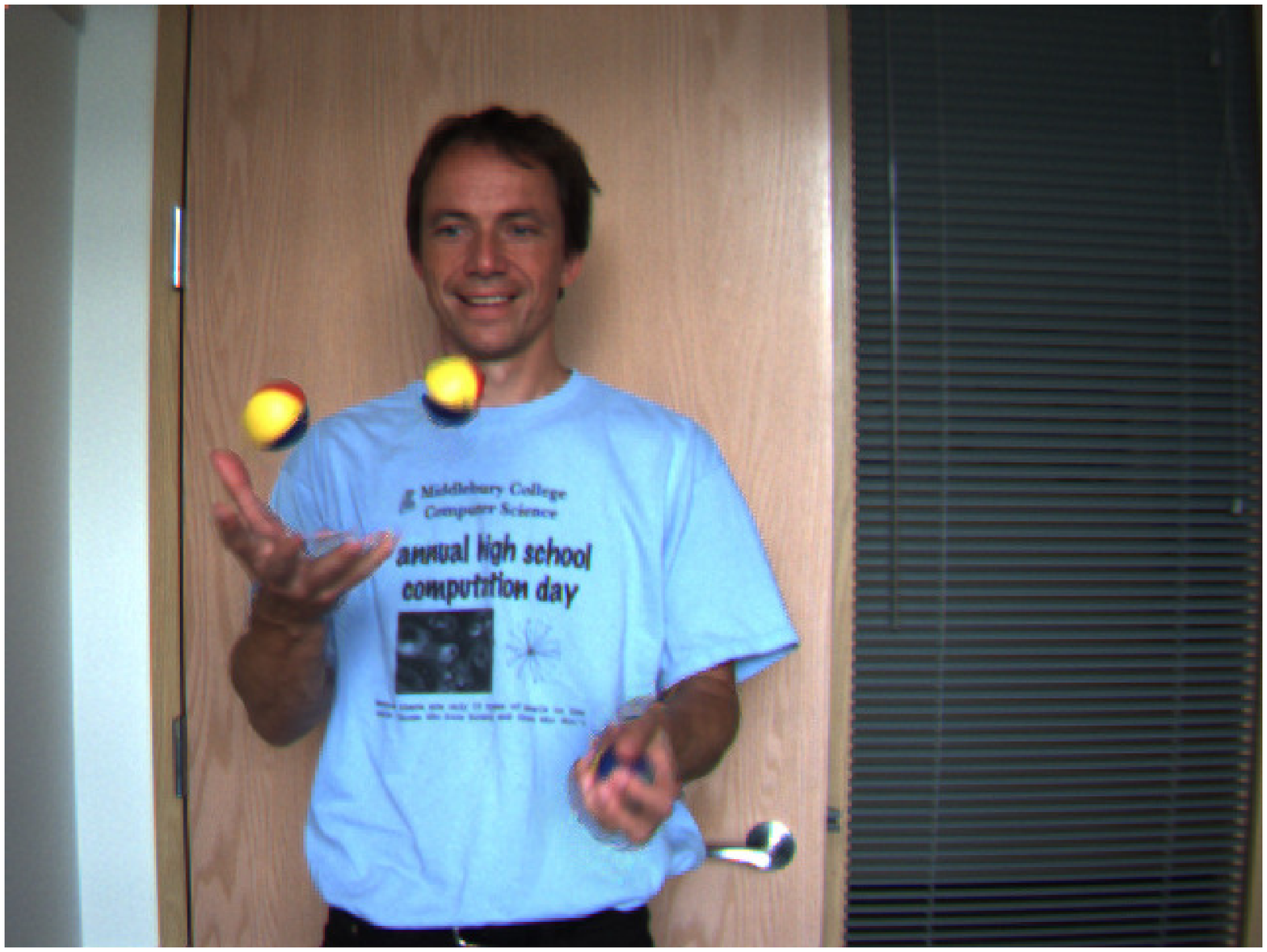}
        &
            \includegraphics[width=\itemwidth,trim={2.7cm 1.0cm 7.8cm 1.3cm},clip]{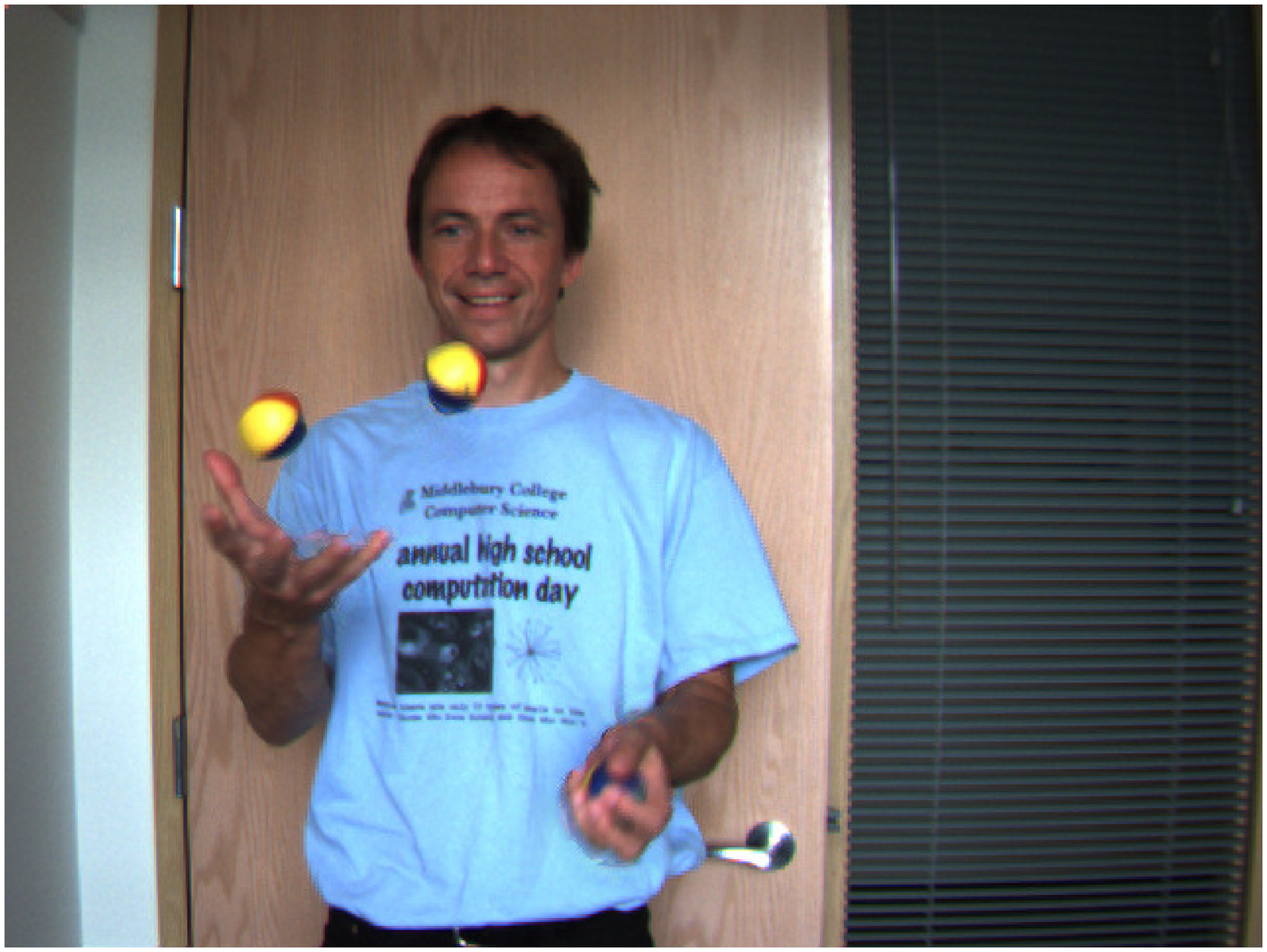}
        &
            \includegraphics[width=\itemwidth,trim={2.7cm 1.0cm 7.8cm 1.3cm},clip]{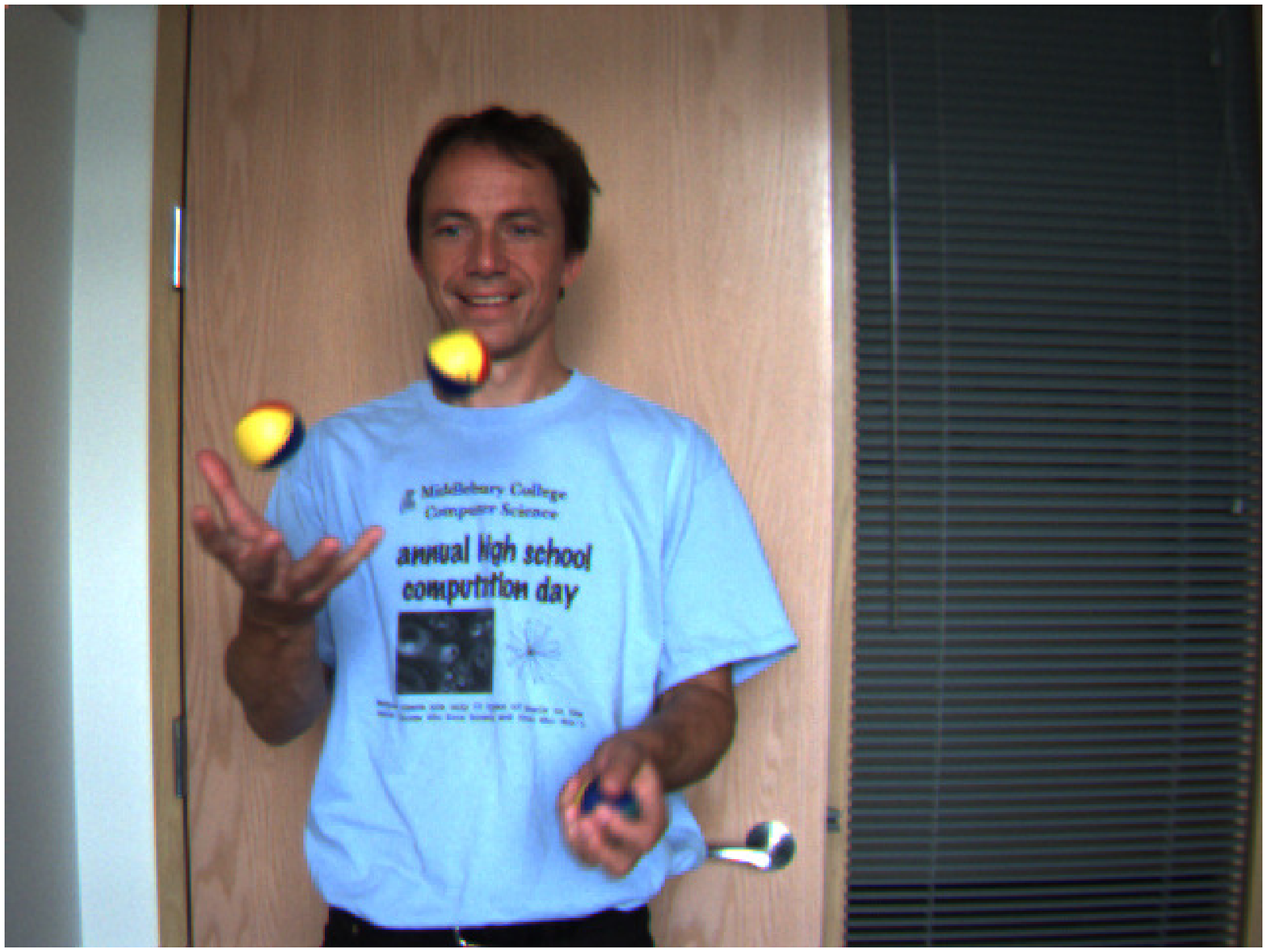}
        \vspace{-0.1cm} \\
            \footnotesize Input at $t = 0$
        &
            \footnotesize Ours - $\mathcal{L}_F$ at $t = 0.2$
        &
            \footnotesize Ours - $\mathcal{L}_F$ at $t = 0.4$
        &
            \footnotesize Ours - $\mathcal{L}_F$ at $t = 0.6$
        &
            \footnotesize Ours - $\mathcal{L}_F$ at $t = 0.8$
        &
            \footnotesize Input at $t = 1$
        \\
    \end{tabularx}\vspace{-0.2cm}
	\captionof{figure}{A sequence of frames demonstrating that our method can interpolate at an arbitrary temporal position.}\vspace{-0.5cm}
	\label{fig:temporal}
\end{figure*}

\begin{figure}\centering
    \setlength{\tabcolsep}{0.0cm}
    \setlength{\itemwidth}{4.15cm}

    \begin{tabularx}{\textwidth}{c @{\hspace{0.05cm}} c}
            \includegraphics[width=\itemwidth]{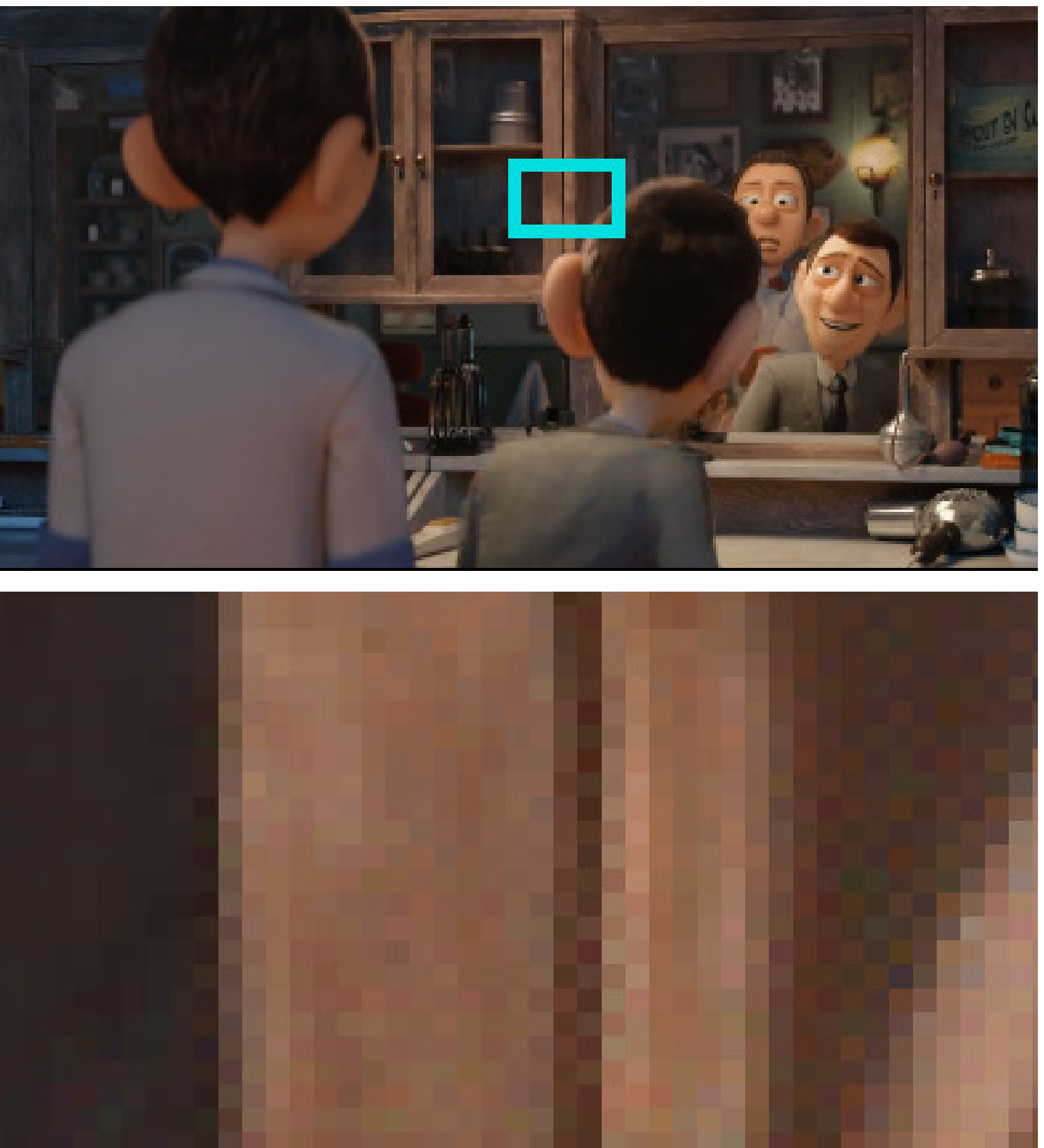}
        &
            \includegraphics[width=\itemwidth]{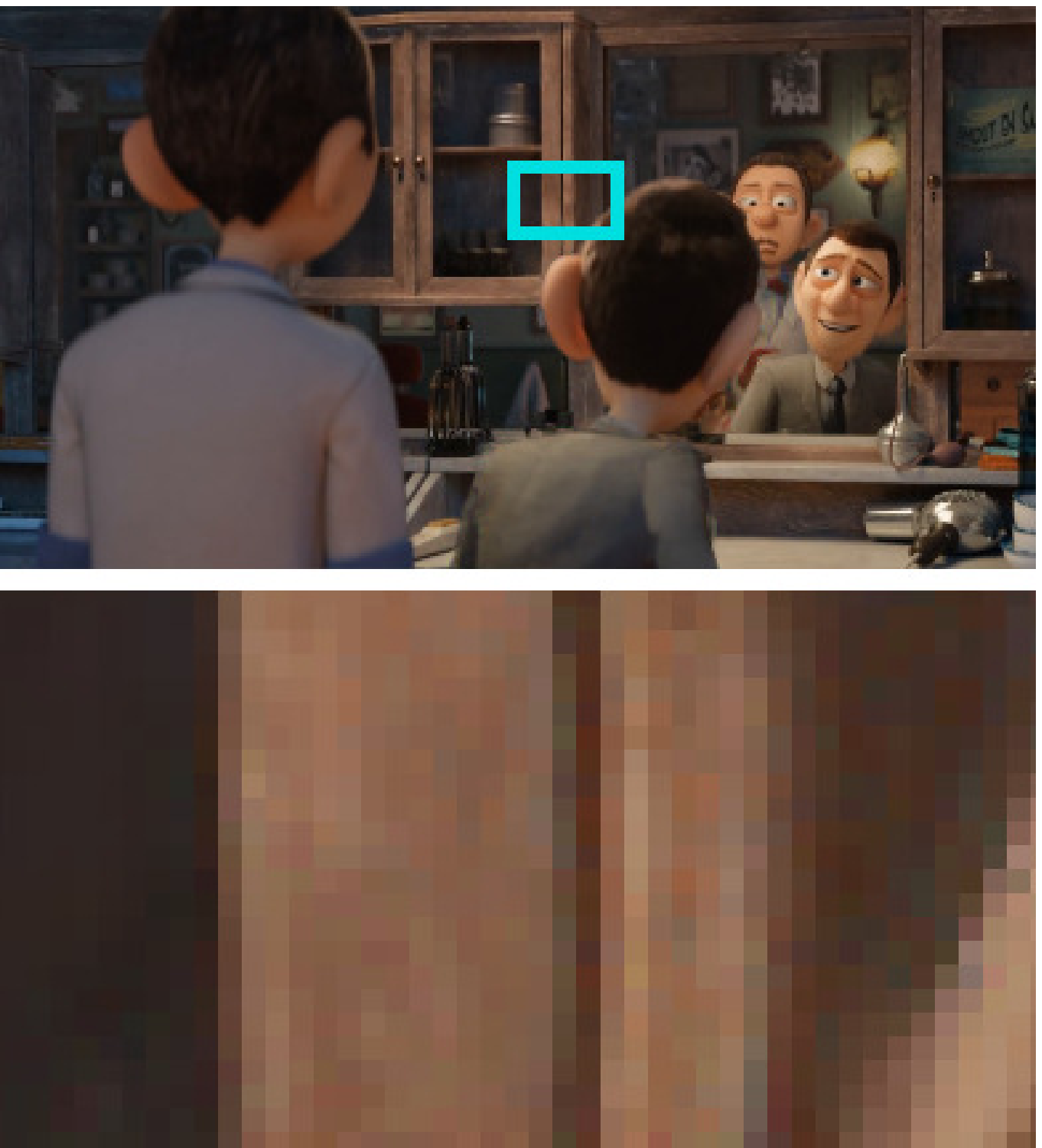}
        \vspace{-0.1cm} \\
            \footnotesize Default GridNet - $\mathcal{L}_F$
        &
            \footnotesize Modified GridNet - $\mathcal{L}_F$
        \\
    \end{tabularx}\vspace{-0.2cm}
    \caption{Checkerboard artifacts prevention.}\vspace{-0.5cm}
    \label{fig:checkerboard}
\end{figure}

\vspace{0.05in}
\noindent\textbf{User study.} We also conducted a user study to further compare our method to other frame interpolation methods on all eight examples from the Middlebury benchmark. Specifically, we compare the results of our approach when using $\mathcal{L}_F$ loss with five state-of-the-art video frame interpolation methods. We recruited 15 students in computer science as participants and supported them with an interface that allowed them to easily switch between two images. For each pair, we asked them to select the better one and let each participant perform 40 such comparisons, thus comparing our result to each of the five competing methods for all 8 examples. The results from this study are summarized in Figure~\ref{fig:study}. Overall, the participants preferred our approach that utilizes $\mathcal{L}_F$ loss. The scenarios where the preference has not been in our favor, Urban with SepConv-$\mathcal{L}_{F}$ and Backward with our $\mathcal{L}_{\textit{Lap}}$ loss, are shown in Figure~\ref{fig:failure}. In the Urban example, SepConv-$\mathcal{L}_{F}$ has fewer artifacts at the boundary. In the Backyard example, $\mathcal{L}_F$ loss introduced chromatic artifacts around the orange ball that is subject to large motion. Nevertheless, this study shows that overall our method with $\mathcal{L}_{F}$ loss achieves better perceptual quality although $\mathcal{L}_{\textit{Lap}}$ loss performs better quantitatively.

\subsection{Discussion}\label{sec:disc}

The recent voxel flow-based video frame interpolation method estimates voxel flow to sample a $2^3$ spatio-temporal neighborhood to generate the interpolation results~\cite{Liu_ICCV_2017}. Since it only samples a $2^3$ space-time volume, it is still fairly limited in handling inaccuracies in motion estimation. As shown in Figure~\ref{fig:voxel}, our approach is able to produce better results due to its flexibility. Quantitatively, our method has a PSNR of $34.62$ while voxel flow has a PSNR of $34.12$ on their DVF dataset.

Since we perform motion compensation before synthesizing the output frame, we are able to interpolate a frame at an arbitrary temporal position $t \in \left[ 0, 1 \right]$, as shown in Figure~\ref{fig:temporal}. Other efforts that use convolutional neural networks for video frame interpolation either had to retrain their synthesis network for a specific $t$ or continue the interpolation recursively in order to achieve this~\cite{Niklaus_CVPR_2017, Niklaus_ICCV_2017}. Furthermore, our proposed approach is not limited to video frame interpolation. In fact, it can also be utilized to synthesize between stereo frames as shown in our supplemental video.

When using $\mathcal{L}_F$ loss, checkerboard artifacts can occur if the architecture of the utilized neural network is not chosen well due to uneven overlaps~\cite{Odena_OTHER_2016}. To avoid such artifacts, we modified the GridNet architecture of the frame synthesis network and utilize bilinear upsampling instead of transposed convolutions. As shown in Figure~\ref{fig:checkerboard}, this decision successfully prevents checkerboard artifacts.

As discussed in Section~\ref{subsec:ablation}, pre-warping input frames and context maps using optical flow is important for our method to produce high-quality frame interpolation results. While PWC-Net provides a good initialization for our method, future advances in optical flow research will benefit our frame interpolation method. 

It has been shown in recent research on image synthesis that a proper adversarial loss can help to produce high quality visual results~\cite{Sajjadi_CORR_2016}. It will be interesting to explore its use to further improve the quality of frame interpolation. Furthermore, as shown in recent papers~\cite{Mayer_CORR_2018, Nah_CVPR_2017}, the composition of the training dataset can be important for low-level computer vision tasks. A comprehensive study of the effect of the training dataset in the context of video frame interpolation could potentially provide important domain-specific insights and further improve the interpolation quality.

\section{Conclusion}
\label{sec:concl}

This paper presented a context-aware synthesis approach for video frame interpolation. This method is developed upon three ideas. First, using bidirectional flow in combination with a flexible frame synthesis neural network can handle challenging cases like occlusions and accommodate inaccuracies in motion estimation. Second, contextual information enables our frame synthesis neural network to perform informative interpolation. Third, using optical flow to provide a good initialization for interpolation is helpful. As demonstrated in our experiments, these ideas enable our method to produce high-quality video frame interpolation results and outperform state-of-the-art methods.

\vspace{0.05in}
\noindent\textbf{Acknowledgments.} Figures~\ref{fig:teaser},~\ref{fig:loss} (top), and ~\ref{fig:examples} (bottom) originate from the DAVIS challenge. Figures~\ref{fig:architecture},~\ref{fig:examples} (top), and~\ref{fig:checkerboard} are used under a Creative Commons license from the Blender Foundation. Figure~\ref{fig:loss} (bottom) originates from the KITTI benchmark. Figures~\ref{fig:ablation} (top),~\ref{fig:failure}, and~\ref{fig:temporal} originate from the Middlebury benchmark. Figure~\ref{fig:ablation} (bottom) is used under a Creative Commons license from GaijinPot. Figure~\ref{fig:voxel} originates from the DVF (from UCF101) dataset. We thank Nvidia for their GPU donation.

{\small
\bibliographystyle{ieee}
\bibliography{egbib}
}

\end{document}